\documentclass[10pt,twocolumn,letterpaper]{article}

\usepackage[pagenumbers]{iccv} 

\usepackage{makecell}
\usepackage{multirow}
\usepackage{adjustbox}
\usepackage{graphicx}
\usepackage{placeins}
\usepackage{float}
\usepackage{afterpage}
\newcommand \blfootnote[1]{
    \begingroup
        \renewcommand
        \thefootnote{}\footnote{#1}
        \addtocounter{footnote}{-1}
        \vspace{-1ex}
    \endgroup
}
\definecolor{iccvblue}{rgb}{0.21,0.49,0.74}
\usepackage[pagebackref,breaklinks,colorlinks,allcolors=iccvblue]{hyperref}

\def\paperID{****} 
\def\confName{ICCV}
\def\confYear{2025}

\title{ConsisLoRA: Enhancing Content and Style Consistency for \\LoRA-based Style Transfer}

\author{Bolin Chen$^{1}$, Baoquan Zhao$^{1}$, Haoran Xie$^{2}$, Yi Cai$^{3}$, Qing Li$^{4}$, Xudong Mao$^{1}$\footnotemark[1]\>\,\\
{$^1$Sun Yat-sen University\ \ $^2$Lingnan University\ \ $^3$South China University of Technology}\\
{$^4$The Hong Kong Polytechnic University }\\
{\href{https://consislora.github.io/}{https://consislora.github.io \vspace{-15pt}}}
}

\begin{document}


\twocolumn[{%
\vspace{-1em}
\maketitle
\renewcommand\twocolumn[1][]{#1}%
\vspace{-0.1in}
\begin{center}
    \centering
    \includegraphics[width=0.98\textwidth]{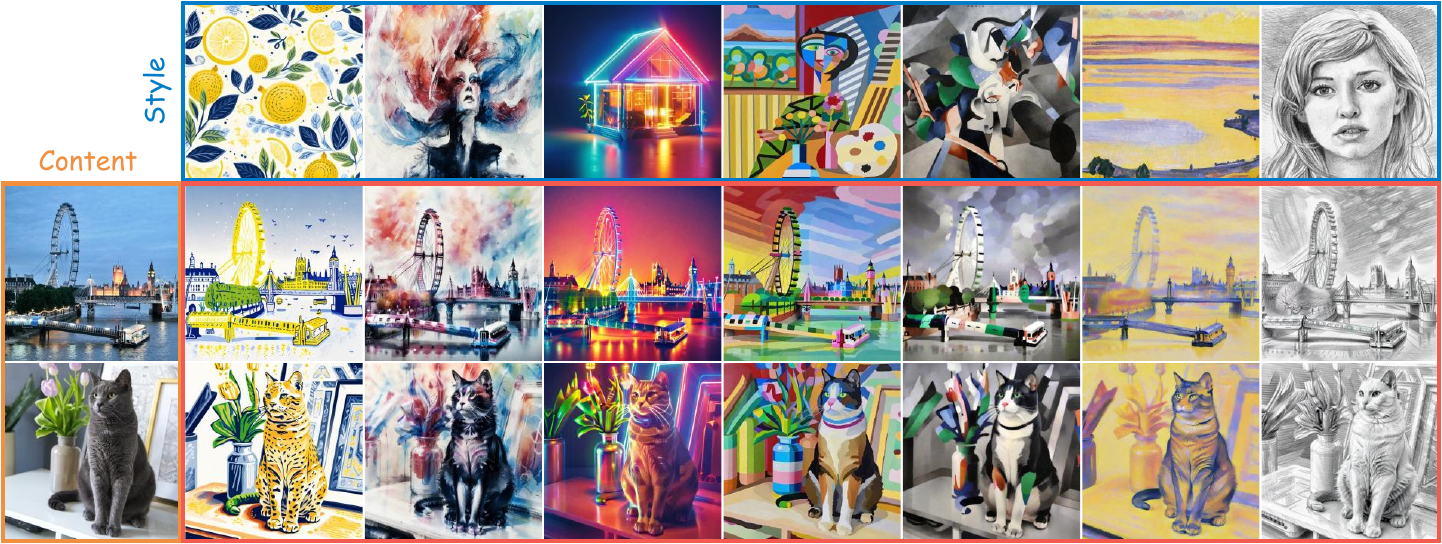}
    \vspace{-0.2cm}
    \captionof{figure}{
    Style transfer results of our method. Given a content image and a style reference image, ConsisLoRA enables high-fidelity stylized generations that excel in both content preservation and style alignment.
    }
    \label{fig:teaser}
\end{center}%
}]

\blfootnote{$^*$Corresponding author (xudong.xdmao@gmail.com).}

\begin{abstract}
Style transfer involves transferring the style from a reference image to the content of a target image. Recent advancements in LoRA-based (Low-Rank Adaptation) methods have shown promise in effectively capturing the style of a single image. However, these approaches still face significant challenges such as content inconsistency, style misalignment, and content leakage. In this paper, we comprehensively analyze the limitations of the standard diffusion parameterization, which learns to predict noise, in the context of style transfer. To address these issues, we introduce ConsisLoRA, a LoRA-based method that enhances both content and style consistency by optimizing the LoRA weights to predict the original image rather than noise. We also propose a two-step training strategy that decouples the learning of content and style from the reference image. To effectively capture both the global structure and local details of the content image, we introduce a stepwise loss transition strategy. Additionally, we present an inference guidance method that enables continuous control over content and style strengths during inference. Through both qualitative and quantitative evaluations, our method demonstrates significant improvements in content and style consistency while effectively reducing content leakage.

\end{abstract}

\section{Introduction}
Diffusion models have emerged as a powerful paradigm for text-to-image synthesis, achieving significant breakthroughs in controllable generation tasks, including personalized generation~\cite{dreambooth, textualInversion}, image editing~\cite{p2p, Instructpix2pix}, and image stylization~\cite{inST, styleID}. 
Despite these advancements, style transfer remains challenging due to the inherently complex and underdetermined nature of style. The goal of style transfer is to transform a content image to match a desired style from a style reference image.

\begin{figure*}[t]
 \centering
 \includegraphics[width=0.95\linewidth]{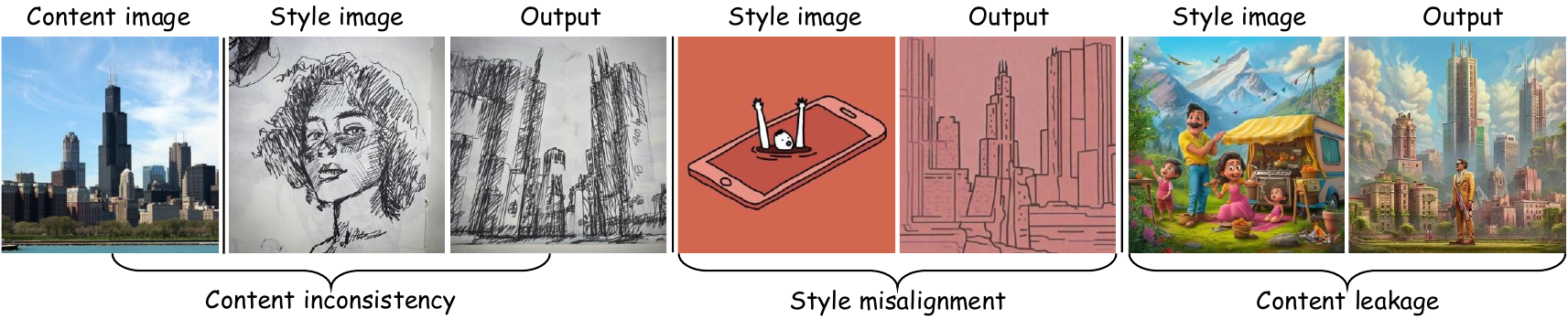}
\caption{Examples of three significant challenges encountered by existing LoRA-based methods: 1) Content inconsistency: the structure of the generated image is inconsistent with that of the content image; 2) Style misalignment: the style of the generated image does not align with that of the style image; 3) Content leakage: content from the style image undesirably leaks into the generated image.}
\label{fig:analysis}
\vspace{-0.2cm}
\end{figure*}

Diffusion models have been extensively applied to style transfer, utilizing methods such as fine-tuning-based approaches~\cite{inST, styleDrop} and tuning-free approaches~\cite{instantStyle, styleID, styleTokenizer}. Recently, LoRA-based techniques~\cite{ziplora, unziplora, PairCustomization} have shown remarkable efficacy in capturing style from a single image. Notably, B-LoRA~\cite{B-LoRA} separates content and style within an image by jointly learning two distinct LoRAs: one for content and another for style. However, as illustrated in \cref{fig:analysis}, current LoRA-based methods still encounter significant challenges. First, accurately capturing high-level structural and stylistic features remains difficult, often resulting in outputs that are inconsistent with the original content or suffer from style misalignment. Second, the precise separation of style and content continues to be challenging, sometimes leading to content leakage~\cite{instantStyle}.

Previous studies on text-to-image personalization~\cite{dreambooth, dreambooth-lora} have revealed that DreamBooth-LoRA~\cite{dreambooth-lora} tends to capture major concepts of the input image (often a part of the image) rather than its entire global structure. This limitation is particularly problematic for style transfer, which requires 1) learning global style information from the entire style image, and 2) capturing the global structure of the content image to ensure content-consistent generations. We attribute these issues to the inappropriate noise prediction loss used in existing LoRA-based methods~\cite{B-LoRA,ziplora}, which fails to adequately focus on global and high-level features.

To overcome these challenges, we introduce ConsisLoRA, a novel approach that optimizes LoRA weights by predicting the original image, where the predicted image is reconstructed from the predicted noise. This reformulated loss function significantly enhances both content and style consistency for LoRA-based style transfer. To further decouple the learning of style and content, we employ a two-step training strategy: initially learning a content-consistent LoRA, followed by learning a style LoRA while keeping the content LoRA fixed. Moreover, we propose a stepwise loss transition approach to capture both the overall structure and the fine details of the content image. We also introduce an inference guidance method that allows for continuous control of content and style strengths during inference.

To demonstrate the effectiveness of ConsisLoRA, we conduct a comprehensive evaluation comparing it against four state-of-the-art baseline methods through both qualitative and quantitative assessments. The results show that ConsisLoRA outperforms the baselines in terms of content preservation and style alignment, while effectively reducing the content leakage.

\section{Related Work}

\paragraph{Fine-tuning Diffusion Models.}
Recent advancements in text-to-image models~\cite{ldm,imagen} have leveraged fine-tuning techniques for personalization, enabling diffusion models to generate images of new concepts from several provided images. Textual Inversion~\cite{textualInversion} optimizes text embeddings to learn new concepts, while DreamBooth~\cite{dreambooth} fine-tunes the entire U-Net of the diffusion model. To enhance fine-tuning efficiency, several parameter-efficient approaches have been proposed~\cite{custom,lora,oft,svdiff}. Notably, LoRA~\cite{lora}, originally developed for fine-tuning large language models, has gained popularity in fine-tuning diffusion models due to its effectiveness and parameter efficiency. In this context, the parameterization of $\epsilon$-prediction~\cite{DDPM} is commonly used for fine-tuning because of its ability to produce high-quality and diverse visual outputs. In this paper, we propose replacing $\epsilon$-prediction with $x_0$-prediction to improve content and style consistency in style transfer. Recently, a concurrent work~\cite{lotus} also employs $x_0$-prediction for high-quality dense prediction by directly altering the output of U-Net to $x_0$. In contrast, our approach derives the predicted image from the predicted noise, without modifying the output of U-Net.

\paragraph{Style Transfer.}
Style transfer, which involves transferring the visual style from a reference image to a target content image, remains a significant challenge in computer vision~\cite{quilting,analogies}. Recent advancements in diffusion models have revolutionized the field of style transfer. These diffusion-based methods can be primarily classified into two main categories. The first approach~\cite{pasd,SigStyle,DreamStyler,Style-Friendly} involves learning the style representation by fine-tuning diffusion models, such as InST~\cite{inST} and StyleDrop~\cite{styleDrop}. The second approach~\cite{SwappingSelfAttention,zero_shot,ArtAdapter,styleAligned,instantStyle,portrait_diffusion,SGDM} explores tuning-free methods to accelerate the stylization process. In particular, IP-Adapter~\cite{Ip-adapter} and Style-Adapter~\cite{styleAdapter} train lightweight adapters to inject style features into cross-attention layers of U-Nets. Some methods achieve this by utilizing large-scale datasets for training~\cite{ControlStyle,CSGO,DEADiff,styleTokenizer}. Additionally, there is a growing trend of research attempting to improve content preservation ~\cite{instantStyle-plus,DiffArtist,styleID,FreeTuner,ctrlx,ArtBank,one_shot}.

\paragraph{LoRA-based Style Transfer.}
Recently, LoRA-based methods~\cite{ziplora, PairCustomization, B-LoRA, unziplora, lora_rar, finestyle,k_lora} have demonstrated the effectiveness in style transfer. These methods often involve training two separate LoRAs to capture content and style, respectively. ZipLoRA~\cite{ziplora} introduces a method that effectively merges independently trained style and content LoRAs, enabling the generation of any subject in any style. Pair Customization~\cite{PairCustomization} jointly learn content and style LoRAs by capturing the stylistic differences between a pair of content and style images. B-LoRA~\cite{B-LoRA} reveals that jointly training the LoRA weights of two specific blocks within the SDXL architecture can effectively separate content and style within a single image. Despite these advancements, challenges remain in generating stylized images that preserve content structure and align the desired style.

\section{Preliminaries}

\textbf{Latent Diffusion Models.}
The Latent Diffusion Model (LDM)~\cite{ldm} utilizes an autoencoder to provide a low-dimensional latent space. The encoder $\mathcal{E}$ maps an image $x$ to a latent representation $z=\mathcal{E}(x)$, and the decoder $\mathcal{D}$ reconstructs the image from this latent representation, i.e., $\mathcal{D}(\mathcal{E}(x))\approx x$. The Denoising Diffusion Probabilistic Model (DDPM)~\cite{DDPM} is employed to train the model within the latent space of the autoencoder.

\paragraph{Parameterizations of Diffusion Models.}
DDPM~\cite{DDPM} introduces two parameterizations of the objective function for model training: $\epsilon$-prediction and $x_0$-prediction. In the context of LDM, the objective functions are defined as:
\begin{align}
    \mathcal{L}_{\epsilon} &= \mathbb{E}_{z_0,\epsilon, t}\left[ \|\epsilon - \epsilon_{\theta}(z_t, t)\|_2^2 \right], \label{eq:epsilon_loss} \\
    \mathcal{L}_{x_0} &= \mathbb{E}_{z_0,\epsilon, t}\left[ \|z_0 - z_{\theta}(z_t, t)\|_2^2 \right], \label{eq:x0_loss}
\end{align}
where the denoising networks $\epsilon_{\theta}$ and $z_{\theta}$ are tasked with predicting the added noise and the original latent, respectively, from the noised latent $z_t$, given a specific timestep $t$. $\epsilon$-prediction is typically used as the training objective, as it empirically yields high-quality and diverse visual outputs.

\paragraph{B-LoRA.}
After examining the SDXL architecture~\cite{SDXL} for LoRA optimization, B-LoRA~\cite{B-LoRA} finds that jointly optimizing the LoRA weights of two specific transformer blocks ($W_0^4$ and $W_0^5$) effectively separates content and style within a single image. Following DreamBooth-LoRA~\cite{dreambooth-lora}, the model is fine-tuned using the diffusion loss (Eq.~\ref{eq:epsilon_loss}) to reconstruct the input image. One trained, the two learned LoRAs can then be used independently or together for various stylization tasks, such as style transfer and text-based stylization. 

\begin{figure}[t]
    \centering
    \includegraphics[width=\columnwidth]{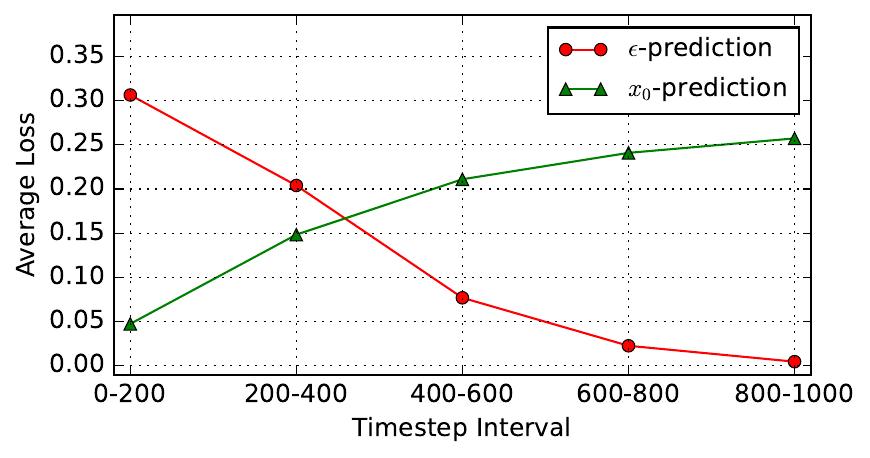} 
    \vspace{-0.6cm}
    \caption{Comparison of the average loss across various timestep intervals for different parameterizations of diffusion models.}
    \vspace{-0.5cm}
    \label{fig:vis_loss}
\end{figure}

\section{Method}

\begin{figure*}[t]
 \centering
 \includegraphics[width=0.95\linewidth]{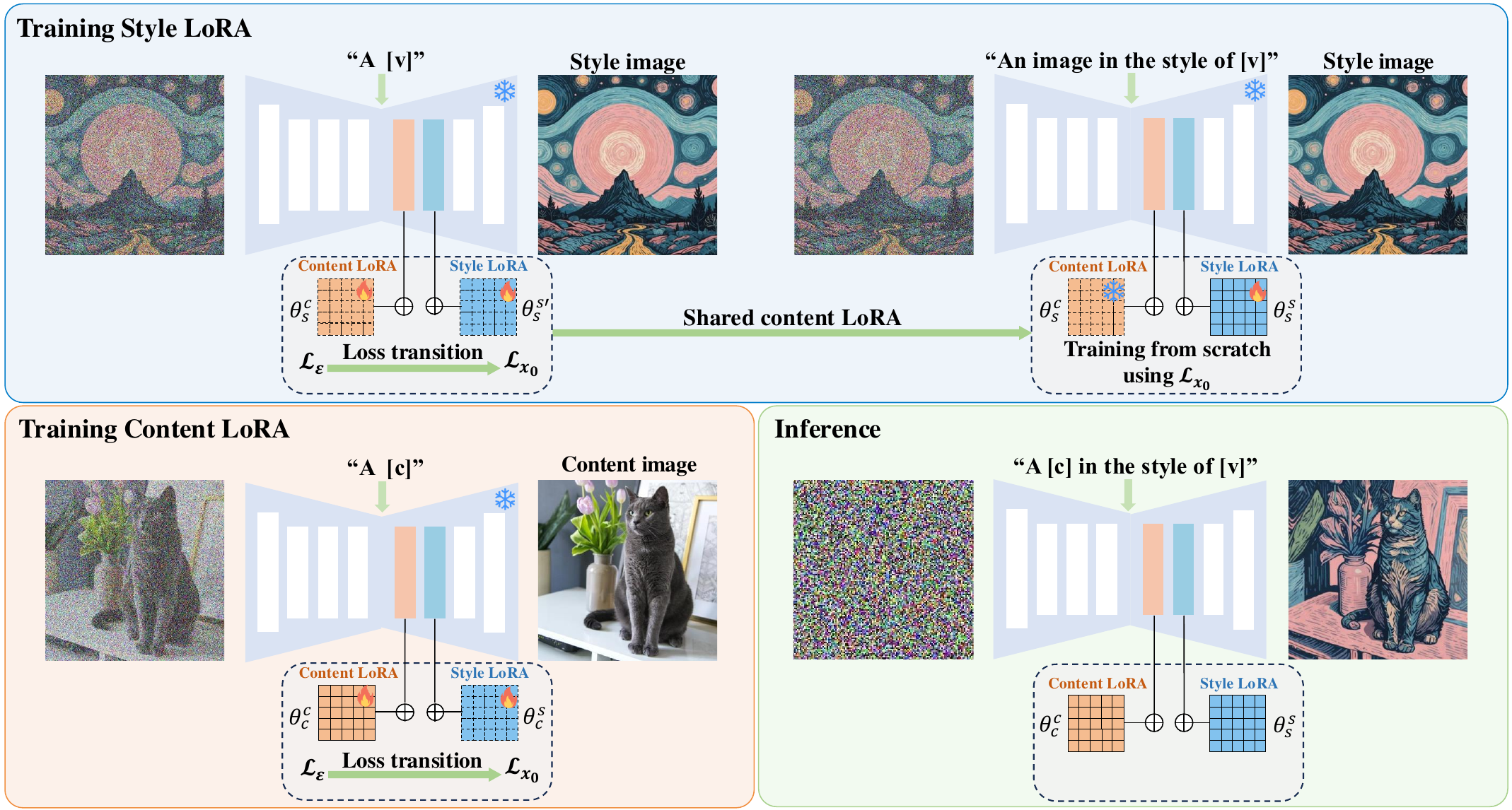}
\caption{\textbf{Method Overview.} We replace the standard $\epsilon$-prediction with $x_0$-prediction for training both style and content LoRAs. (Bottom-left) For training the content LoRA, we propose a loss transition strategy to capture both the global structure and local details of the content image. (Top) To disentangle the learning of style and content from the style image, we introduce a two-step training strategy: first, learn a content-consistent LoRA using the proposed loss transition, and then, train a style LoRA while keeping the content LoRA fixed.}
\label{fig:framework}
\vspace{-0.2cm}
\end{figure*}

\subsection{Analysis of $\epsilon$-prediction for Style Transfer} \label{sec:analysis}
The $\epsilon$-prediction loss, as defined in Eq.\ref{eq:epsilon_loss}, is commonly employed as the objective function for training~\cite{DDPM,ldm} or fine-tuning~\cite{dreambooth,B-LoRA} diffusion models. However, as illustrated in \cref{fig:analysis}, when $\epsilon$-prediction is applied to style transfer, it leads to three significant issues: 1) the structure of the generated image is inconsistent with that of the content image, 2) the style of the generated image does not align with that of the style image, and 3) content from the style image leaks into the generated image.

These problems can be attributed to the inherent focus of $\epsilon$-prediction on low-level local details rather than on high-level structure and style. In \cref{fig:vis_loss}, we present the average loss values of $\epsilon$-prediction at various timestep stages, demonstrating that the loss is high at small $t$ and diminishes as $t$ increases. This pattern occurs because at large $t$, the noised image approaches pure noise, simplifying the task for the model to predict the noise. Conversely, at small $t$, where the noised image closely resembles the original, the model must discern fine details to effectively predict the noise. Consequently, $\epsilon$-prediction emphasizes low-level features at early timesteps and neglects high-level features at later timesteps. Given that style transfer requires capturing the global structure of the content image and the overall style of the style image, $\epsilon$-prediction is suboptimal for this application.

\subsection{ConsisLoRA}
\label{sec:consisLoRA}
Our approach builds upon B-LoRA~\cite{B-LoRA}, which jointly learns content and style LoRAs corresponding to two specific blocks within SDXL from a single image. We introduce ConsisLoRA, a LoRA-based method designed to enhance content and style consistency in style transfer. ConsisLoRA is based on three main ideas. First, we replace the standard $\epsilon$-prediction loss with $x_0$-prediction loss to address the challenges detailed in \cref{sec:analysis}. Second, we introduce a two-step training strategy that more effectively separates the content and style representations within the style image. Third, we propose a stepwise loss transition strategy to simultaneously capture the overall structure and fine details of the content image. An overview of the proposed ConsisLoRA is depicted in \cref{fig:framework}.

\paragraph{Content- and Style-Consistent LoRA.}
As analyzed in \cref{sec:analysis}, the $\epsilon$-prediction loss tends to focus on low-level local details rather than high-level structure and style, making it unsuitable for style transfer. To address this, we propose replacing the traditional $\epsilon$-prediction (Eq.~\ref{eq:epsilon_loss}) with the $x_0$-prediction (Eq.~\ref{eq:x0_loss}) for optimizing both content and style LoRAs. It is important to note that we do not directly alter the output of U-Net from the predicted noise $\epsilon_\theta$ to the predicted latent $z_\theta$. Instead, we derive the predicted latent from the predicted noise through
\begin{equation} \label{eq:epsilon_to_x0}
    \hat{z}_0 = \frac{z_t-\sqrt{1-\bar{\alpha}_t} \epsilon_\theta}{\sqrt{\bar{\alpha}_t}},
\end{equation}
where $\bar{\alpha}_t=\prod_{i=1}^t(1-\beta_i)$ and $\{\beta_i\}$ represents the variance schedule. Then, we minimize the difference between the predicted latent $\hat{z}_0$ and the original latent $z_0$ as:
\begin{equation}\label{eq:our_objective}
    \mathcal{L}_{\hat{z}_0} = \mathbb{E}_{z_0,\epsilon, t}\left[ \|z_0 - \hat{z}_0\|_2^2 \right].    
\end{equation}
As shown in \cref{fig:vis_loss}, in contrast to $\epsilon$-prediction, the proposed loss exhibits a large value for large $t$ and a small value for small $t$. This behavior arises because the $x_0$-prediction loss is scaled by a factor of $\frac{\sqrt{1-\bar{\alpha}_t}}{\sqrt{\bar{\alpha}_t}}$, which becomes substantial at large timesteps. This indicates that $x_0$-prediction more effectively emphasizes high-level features compared to $\epsilon$-prediction, as these features are primarily determined at large timesteps~\cite{p2p}.

\paragraph{Stepwise Loss Transition for Content LoRA.}
In \cref{fig:epsilon_x0} (see Appendix~\ref{sec:appendix_epsilon_x0}), we compare the outputs of using $\epsilon$-prediction and $x_0$-prediction. As shown, while $x_0$-prediction more accurately captures the global structure of the content image, it occasionally fails to retain some local details. To address this, we propose a stepwise loss transition strategy for the content LoRA. Initially, we optimize the LoRA weights using $\epsilon$-prediction for a subset of the training steps and subsequently switch to $x_0$-prediction for the remaining steps. As demonstrated in \cref{fig:epsilon_x0}, this approach effectively preserves both the global structure and local details. We also experimented a gradual transition from $\epsilon$-prediction to $x_0$-prediction (e.g., a linear change over timesteps), but observed no performance gains. Importantly, this stepwise loss transition is not applied to the style LoRA, as our empirical findings suggest that $\epsilon$-prediction in style LoRA optimization leads to the inadvertent capture of local content details, resulting in content leakage issues (see \cref{sec:ablation}).

\paragraph{Disentangling Style and Content for Style LoRA.}
To effectively separate the learning of style and content from the reference image, our strategy begins by accurately learning a content LoRA, followed by learning a style LoRA while keeping the learned content LoRA fixed. We utilize our proposed loss transition training strategy to learn a content-consistent LoRA from the reference image. As demonstrated in \cref{fig:ablation}, the jointly learned style LoRA tends to exhibit content leakage, likely due to two primary reasons: 1) simultaneous optimization of style and content LoRAs can lead them to learn shared features that are relevant to both style and content, and 2) the use of $\epsilon$-prediction in the loss transition strategy causes the style LoRA to inadvertently capture local content details. To overcome these issues, we propose training the style LoRA separately from scratch using $x_0$-prediction, while keeping the learned content LoRA fixed. Moreover, this approach of separate training allows for more focused style learning by using a style-specific prompt, such as ``An image in the style of [v]'', instead of a generic prompt like ``A [v]'' used in \cite{B-LoRA}, guiding the LoRA to exclusively capture style attributes.

\begin{figure*}[t]
    \centering
    \setlength{\tabcolsep}{0.85pt}
    \renewcommand{\arraystretch}{0.5}
    {
    \begin{tabular}{c c@{\hspace{0.07cm}} | @{\hspace{0.07cm}}c c c c c}
        
        Content & \multicolumn{1}{c@{}}{Style} & \multicolumn{1}{c}{Ours} & B-LoRA & ZipLoRA & StyleID & StyleAligned \\
          
        \includegraphics[width=0.125\textwidth]{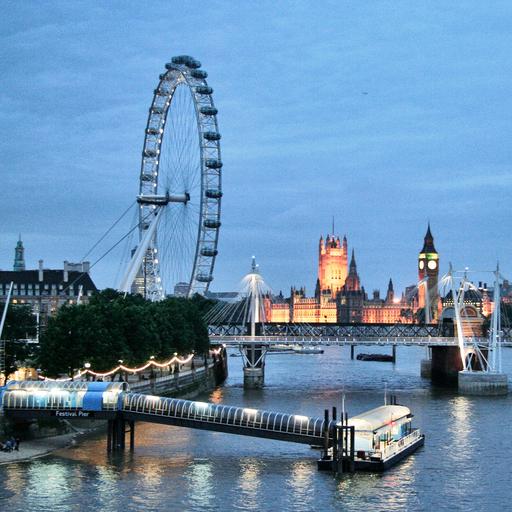} &
        \includegraphics[width=0.125\textwidth]{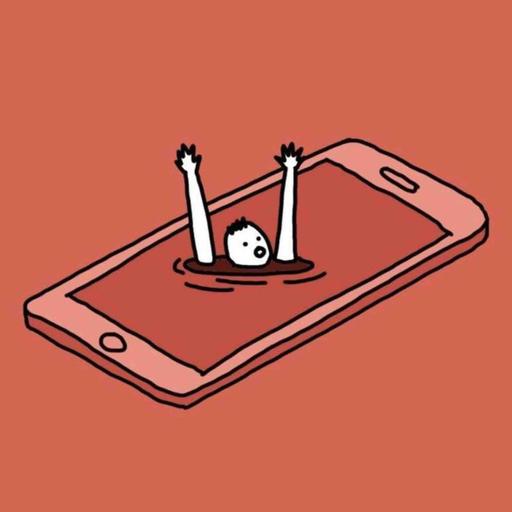} &
        \includegraphics[width=0.125\textwidth]{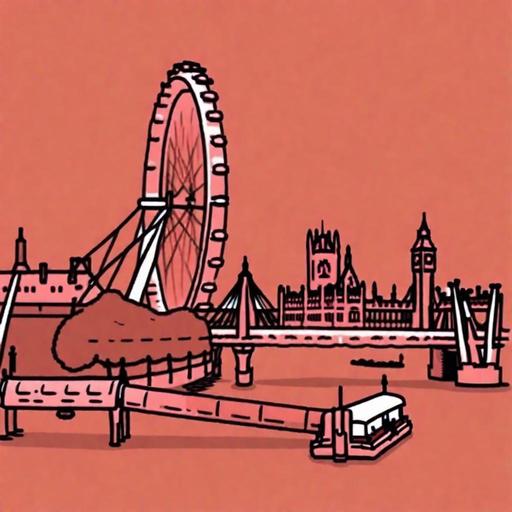} &
        \includegraphics[width=0.125\textwidth]{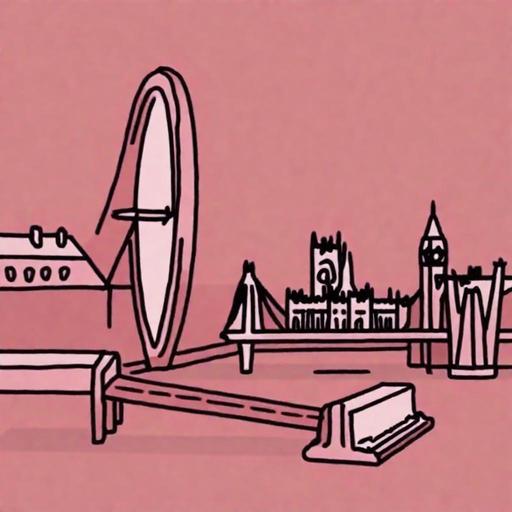} &
        \includegraphics[width=0.125\textwidth]{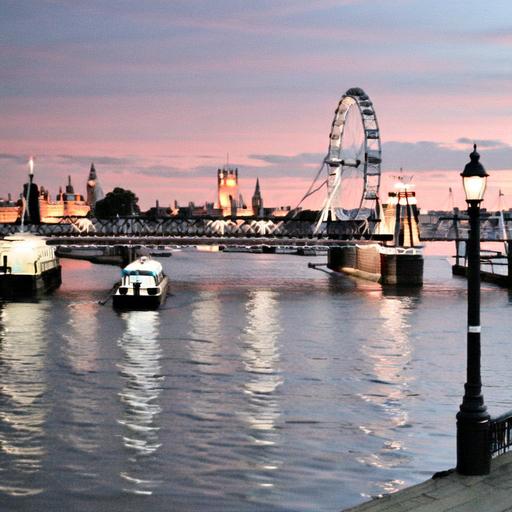} &
        \includegraphics[width=0.125\textwidth]{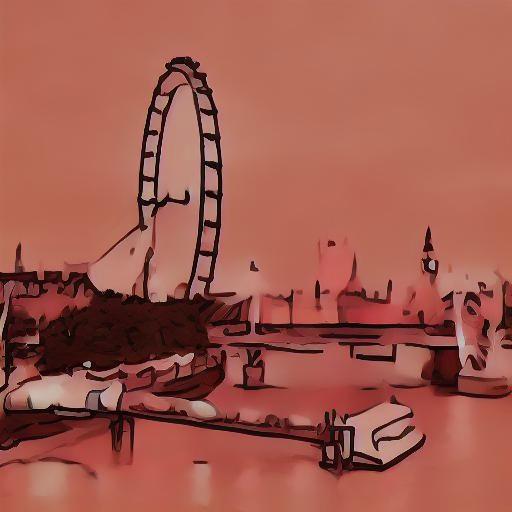} &
        \includegraphics[width=0.125\textwidth]{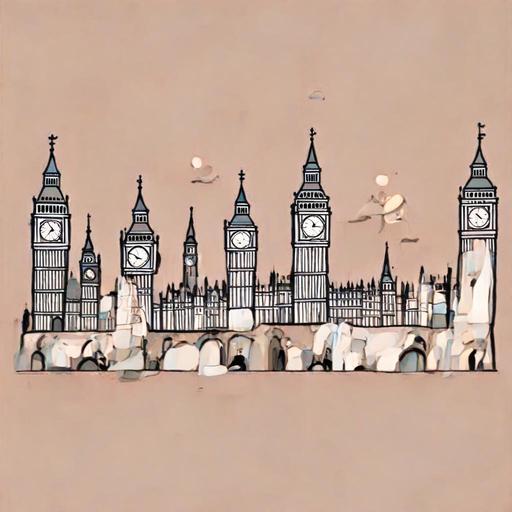} \\

        \includegraphics[width=0.125\textwidth]{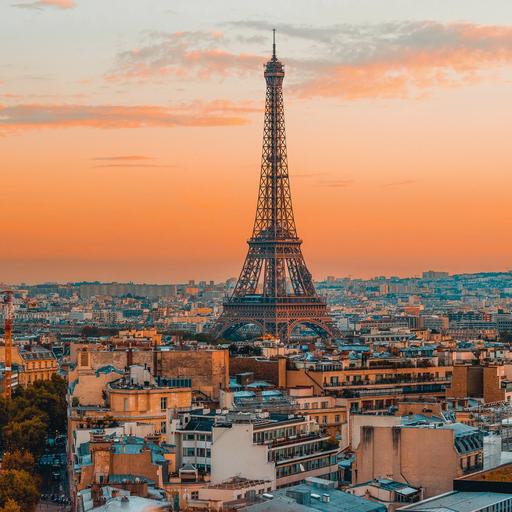} &
        \includegraphics[width=0.125\textwidth]{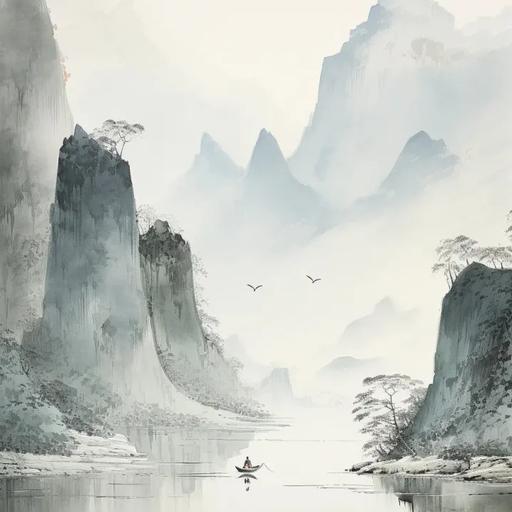} &
        \includegraphics[width=0.125\textwidth]{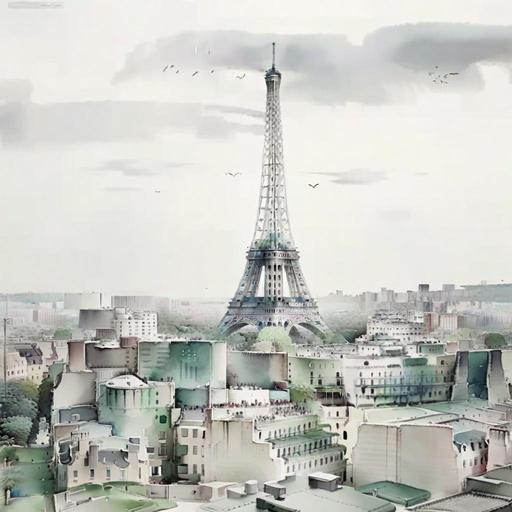} &
        \includegraphics[width=0.125\textwidth]{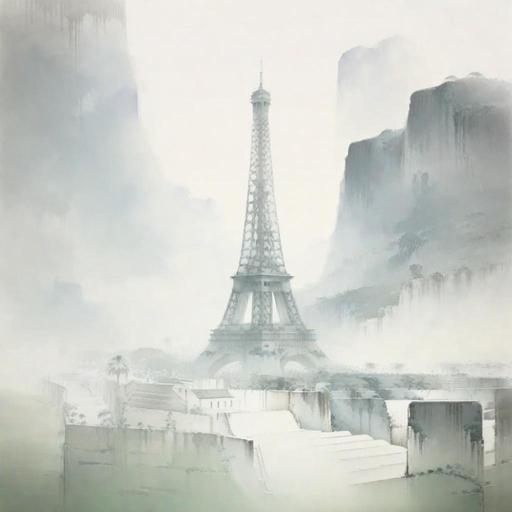} &
        \includegraphics[width=0.125\textwidth]{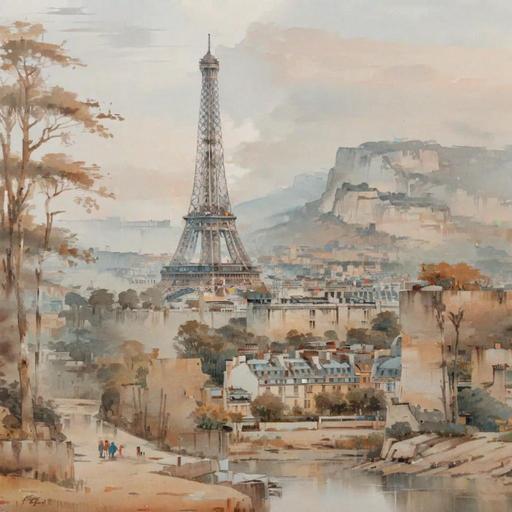} &
        \includegraphics[width=0.125\textwidth]{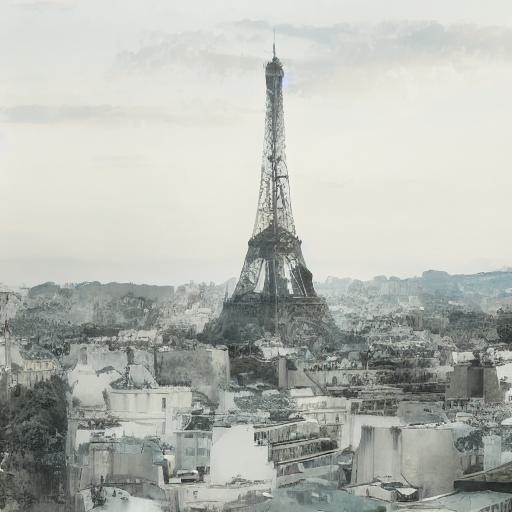} &
        \includegraphics[width=0.125\textwidth]{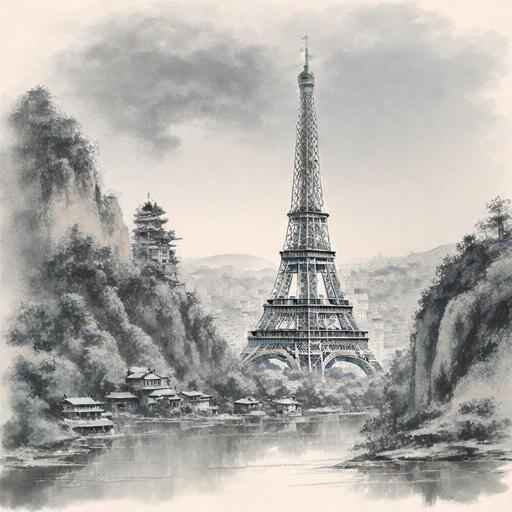} \\

        \includegraphics[width=0.125\textwidth]{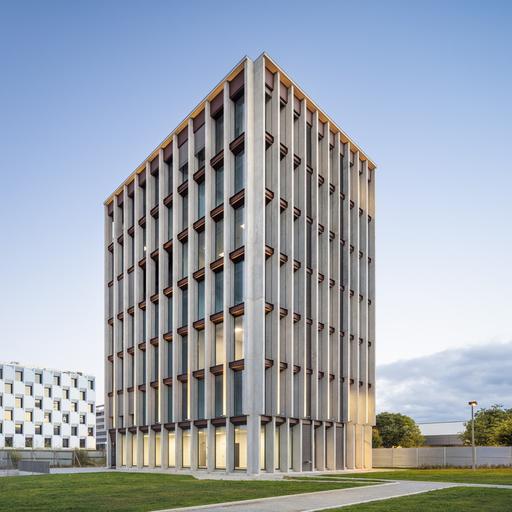} &
        \includegraphics[width=0.125\textwidth]{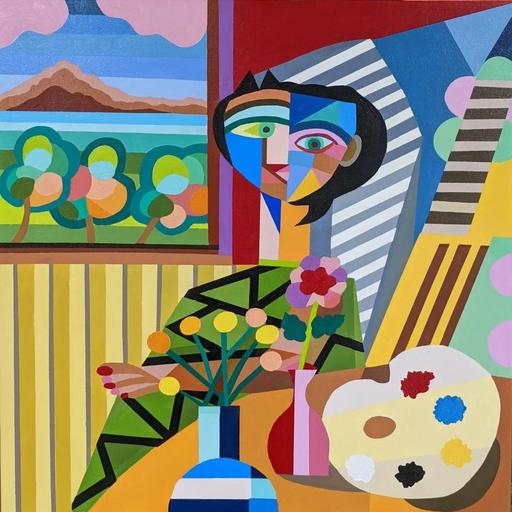} &
        \includegraphics[width=0.125\textwidth]{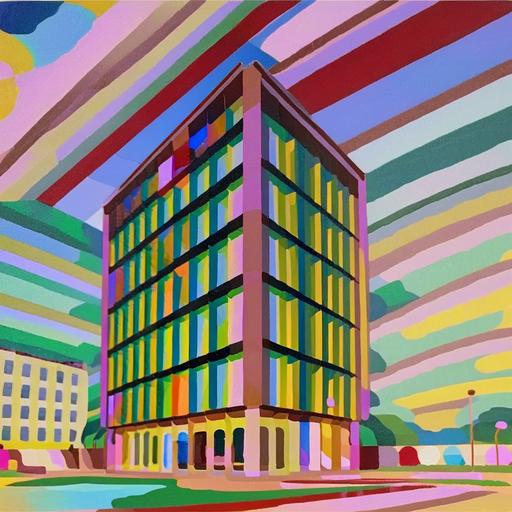} &
        \includegraphics[width=0.125\textwidth]{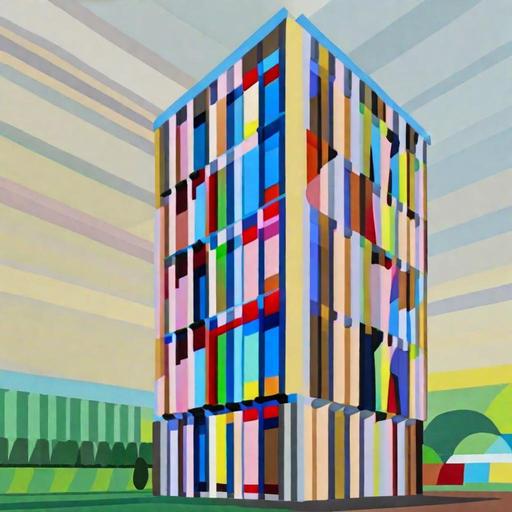} &
        \includegraphics[width=0.125\textwidth]{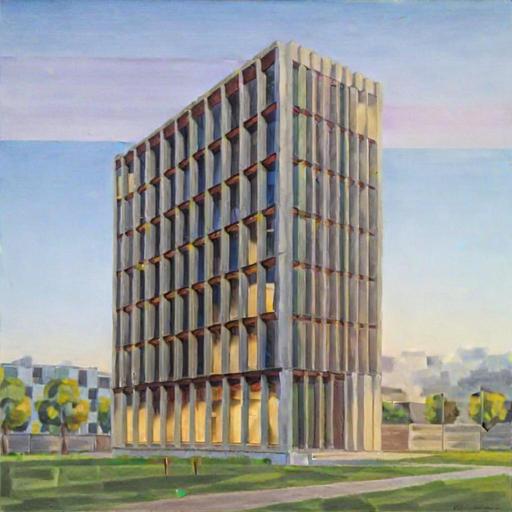} &
        \includegraphics[width=0.125\textwidth]{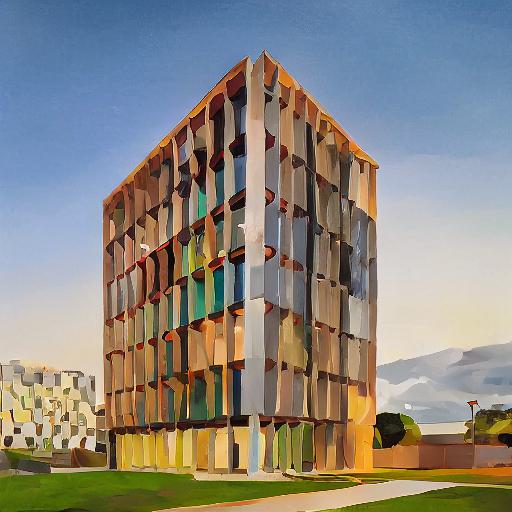} &
        \includegraphics[width=0.125\textwidth]{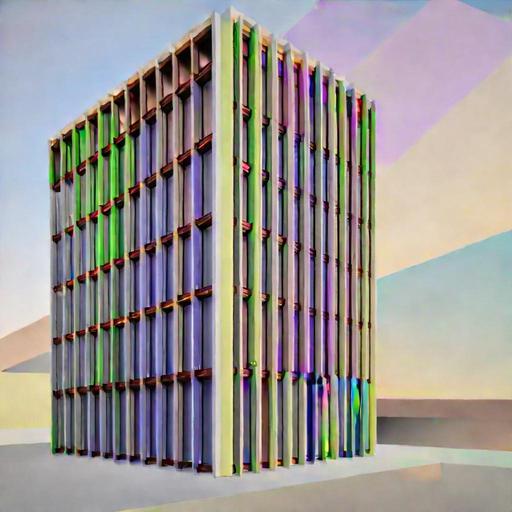} \\

        \includegraphics[width=0.125\textwidth]{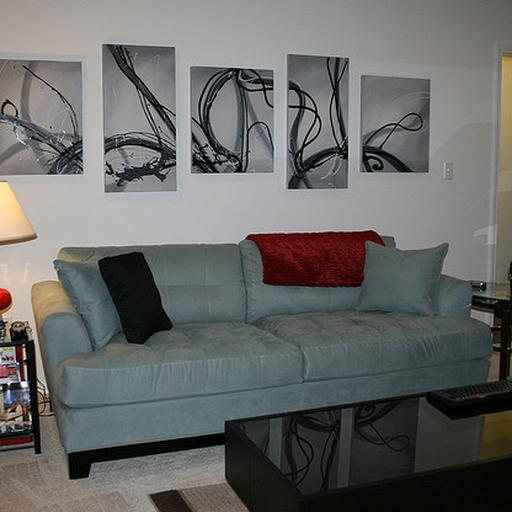} &
        \includegraphics[width=0.125\textwidth]{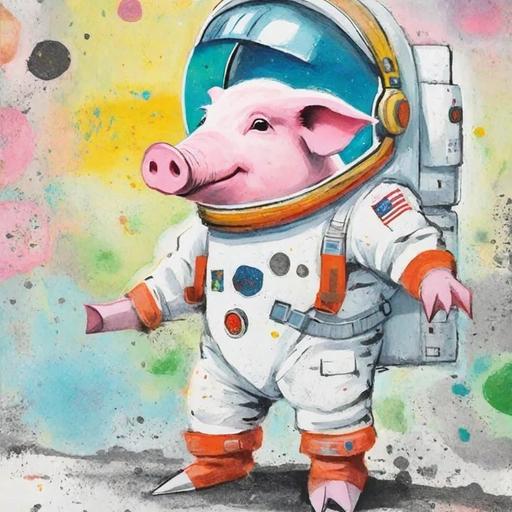} &
        \includegraphics[width=0.125\textwidth]{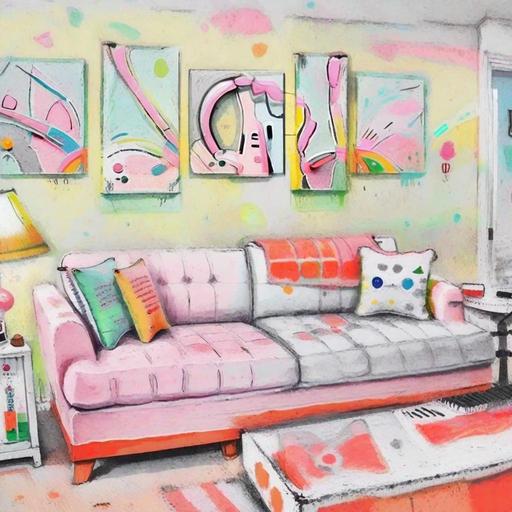} &
        \includegraphics[width=0.125\textwidth]{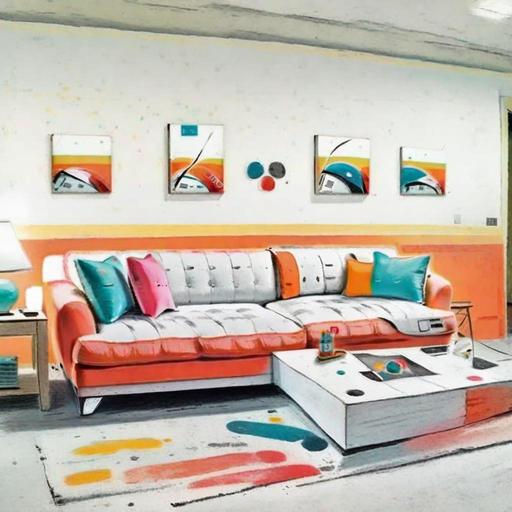} &
        \includegraphics[width=0.125\textwidth]{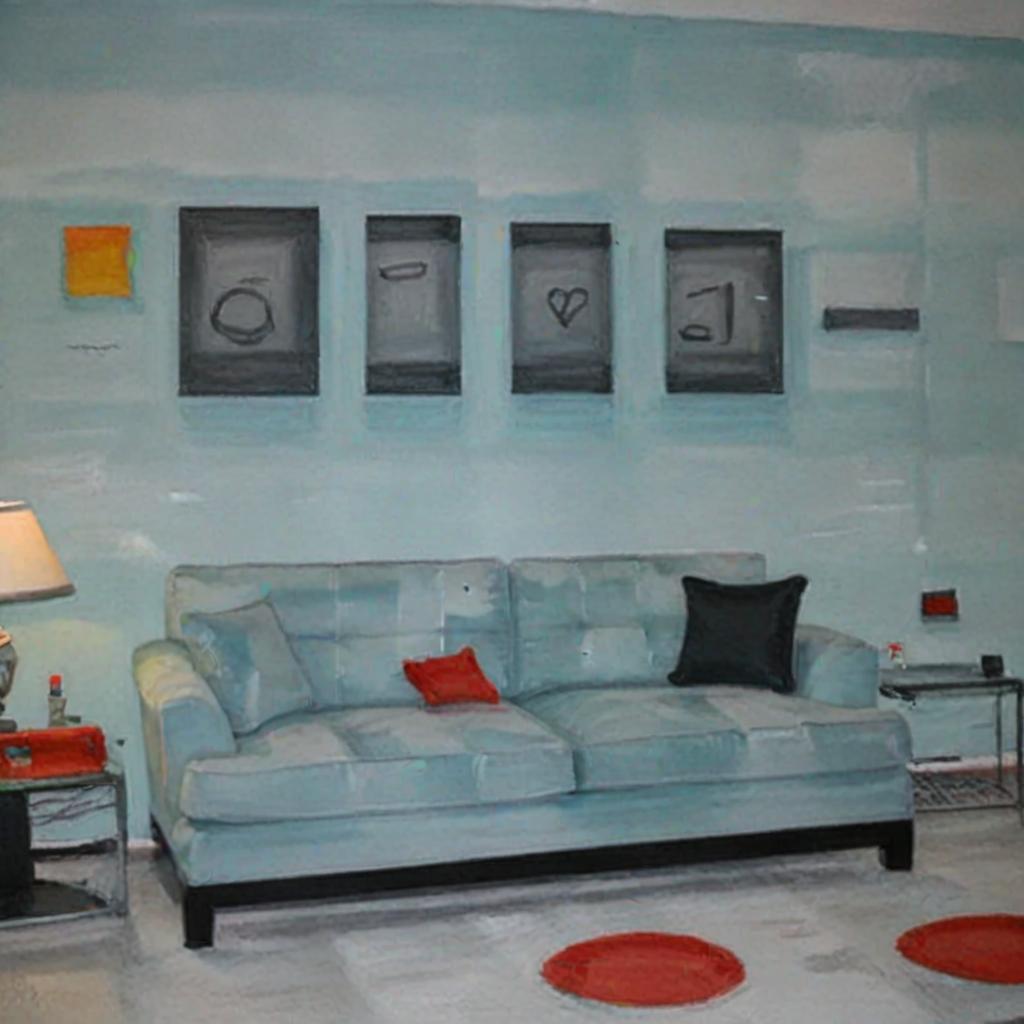} &
        \includegraphics[width=0.125\textwidth]{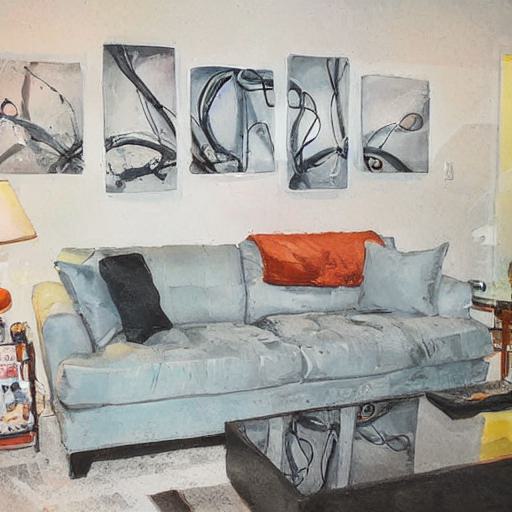} &
        \includegraphics[width=0.125\textwidth]{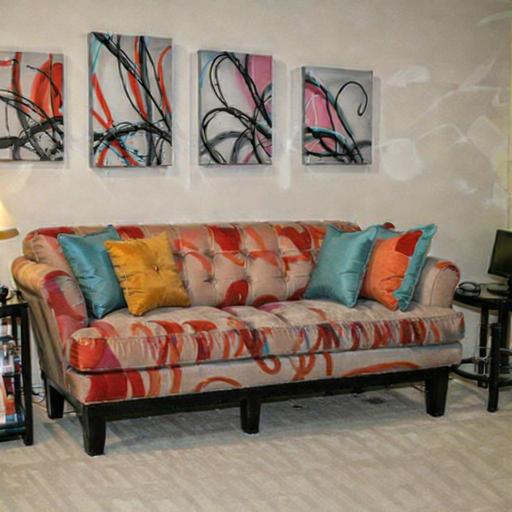} \\

        \includegraphics[width=0.125\textwidth]{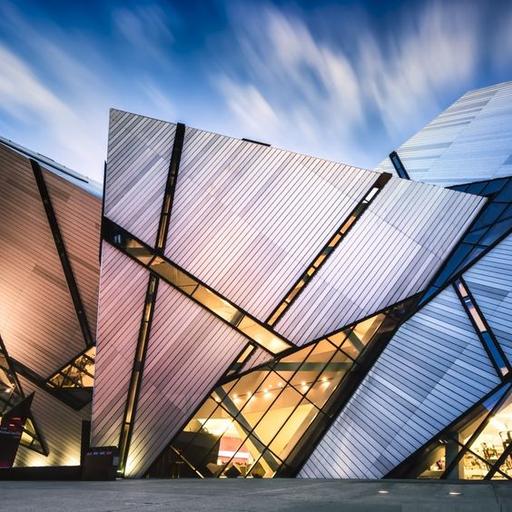} &
        \includegraphics[width=0.125\textwidth]{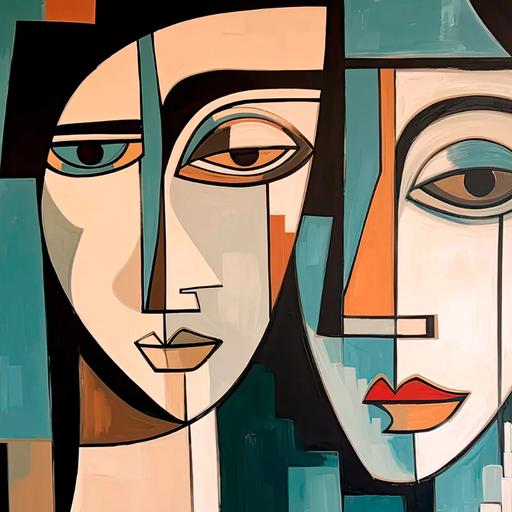} &
        \includegraphics[width=0.125\textwidth]{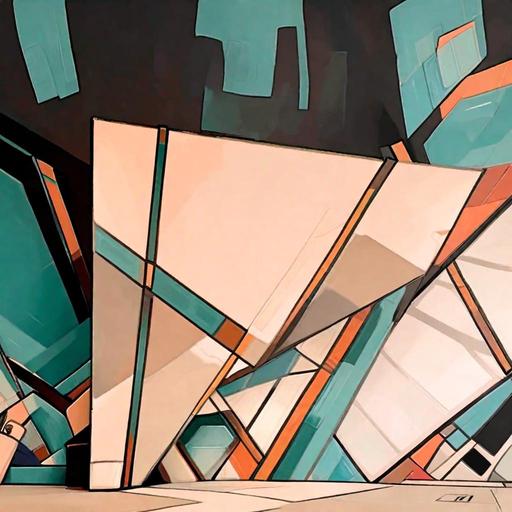} &
        \includegraphics[width=0.125\textwidth]{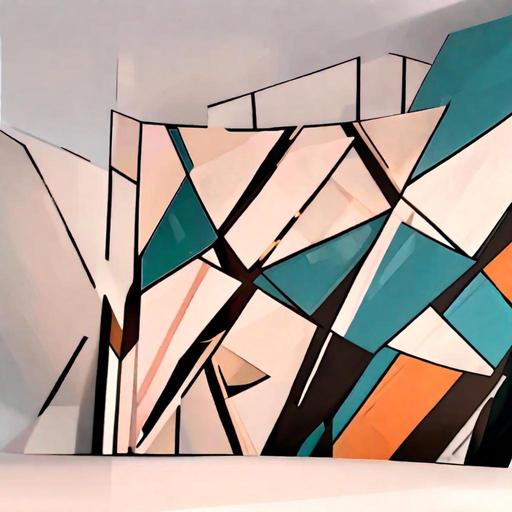} &
        \includegraphics[width=0.125\textwidth]{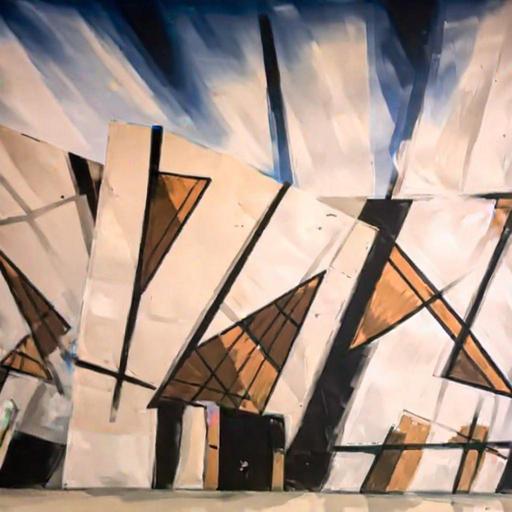} &
        \includegraphics[width=0.125\textwidth]{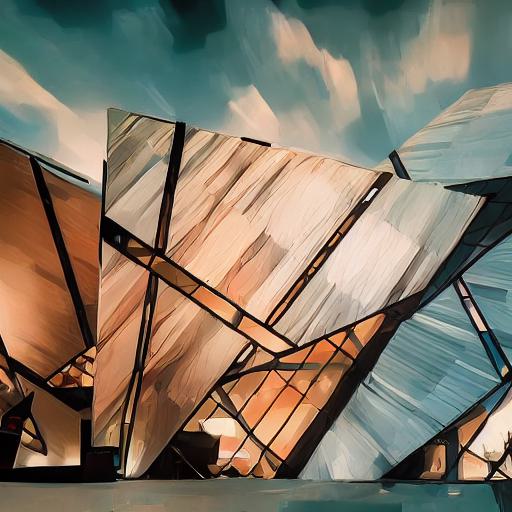} &
        \includegraphics[width=0.125\textwidth]{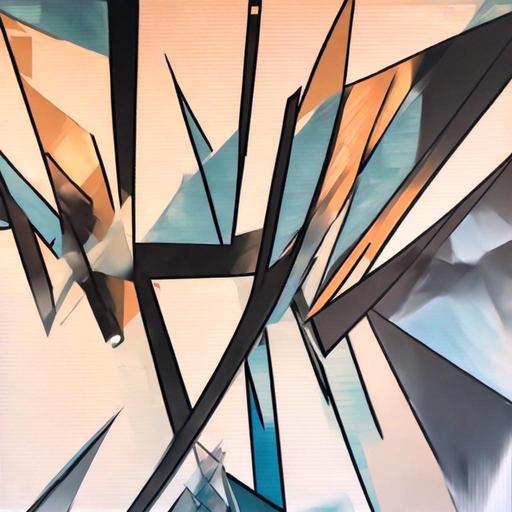} \\

        \includegraphics[width=0.125\textwidth]{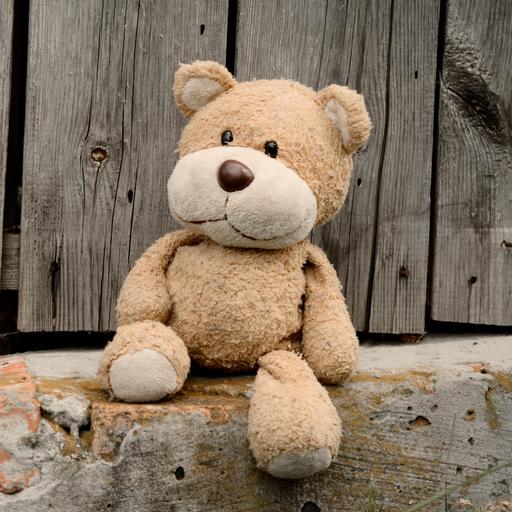} &
        \includegraphics[width=0.125\textwidth]{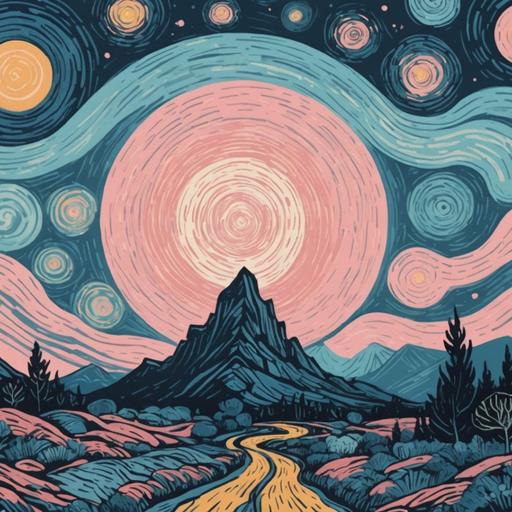} &
        \includegraphics[width=0.125\textwidth]{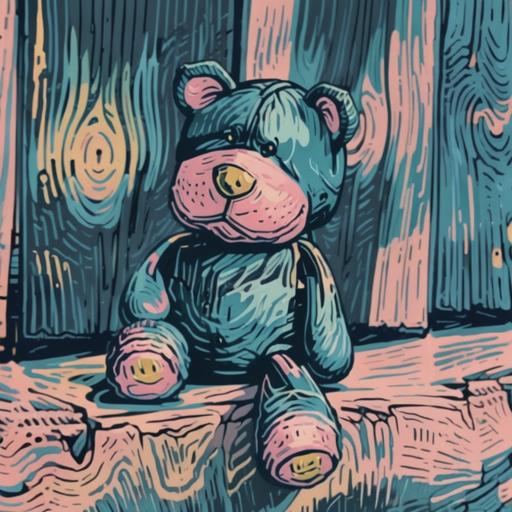} &
        \includegraphics[width=0.125\textwidth]{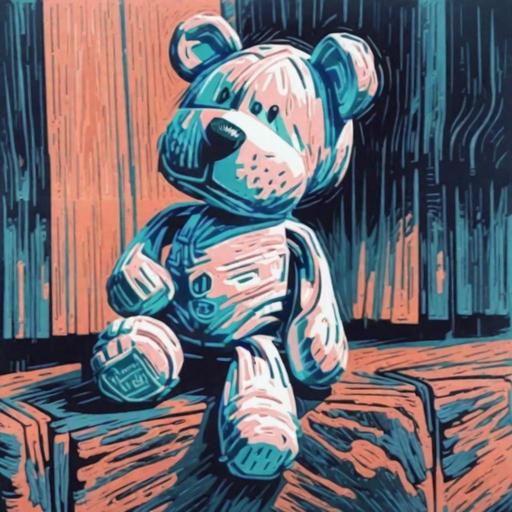} &
        \includegraphics[width=0.125\textwidth]{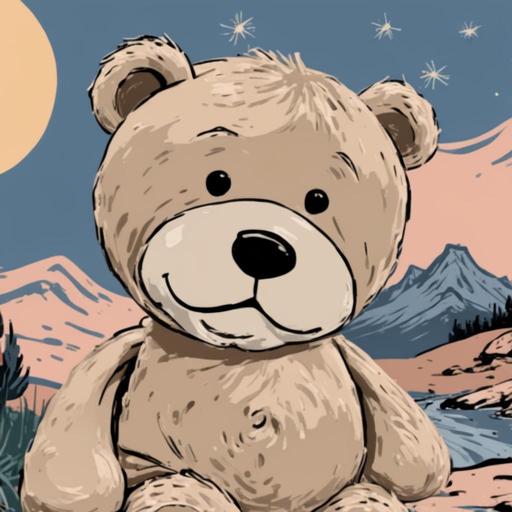} &
        \includegraphics[width=0.125\textwidth]{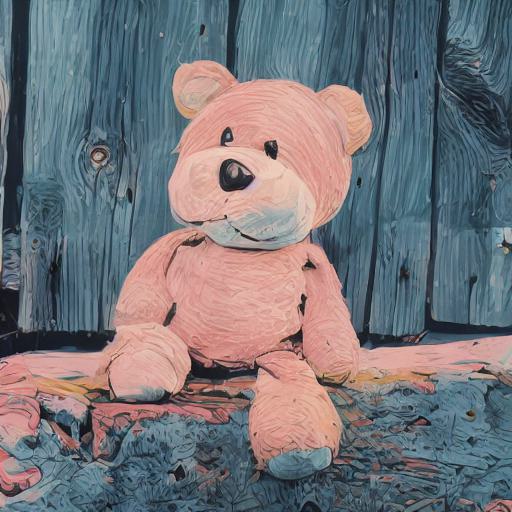} &
        \includegraphics[width=0.125\textwidth]{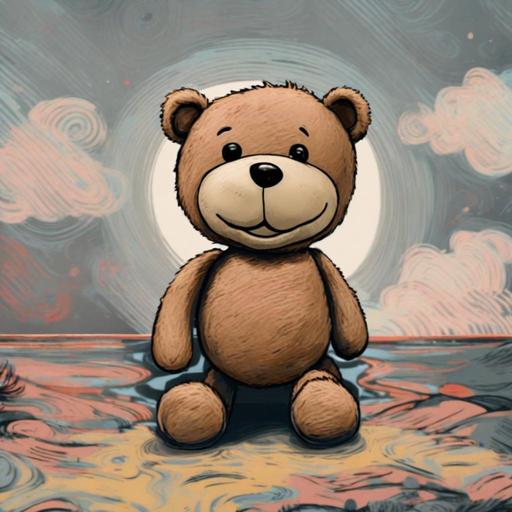} \\
        
    \end{tabular}
    }   
    \caption{\textbf{Qualitative comparison.} We present style transfer results of our method and four baseline methods, including B-LoRA~\cite{B-LoRA}, ZipLoRA~\cite{ziplora}, StyleID~\cite{styleID}, and StyleAligned~\cite{styleAligned}. Our method demonstrates superior performance in preserving the structure of the content image while accurately applying the style from the reference style image.}
    \label{fig:qualitative_comparison}
    \vspace{-0.2cm}
\end{figure*}

\begin{figure*}[t]
    \centering
    \setlength{\tabcolsep}{0.85pt}
    \renewcommand{\arraystretch}{0.5}
    {\small
    \begin{tabular}{c@{\hspace{0.07cm}} | @{\hspace{0.07cm}}c c c c c c}
        
        \quad \ \ \ \ Content \ \ \raisebox{0.33in}{\rotatebox[origin=t]{90}{Style}}&
        \includegraphics[width=0.125\textwidth]{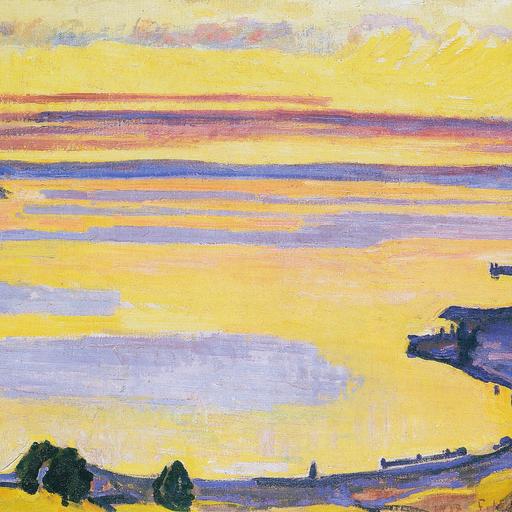} &
        \includegraphics[width=0.125\textwidth]{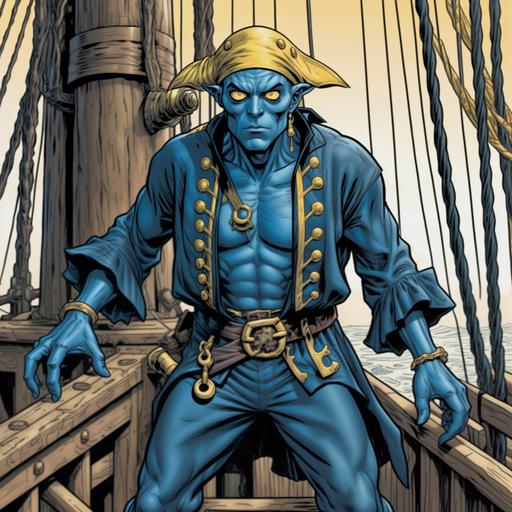} &
        \includegraphics[width=0.125\textwidth]{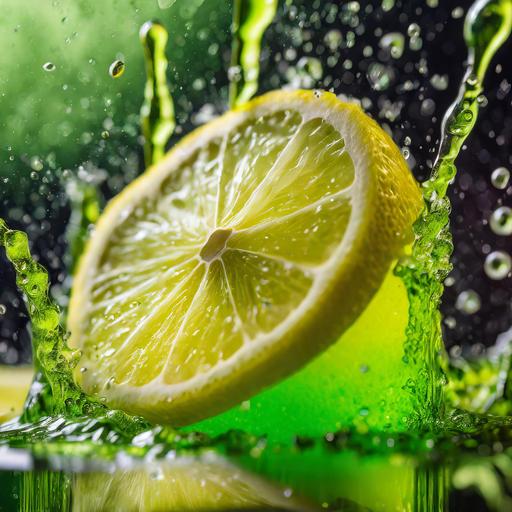} &
        \includegraphics[width=0.125\textwidth]{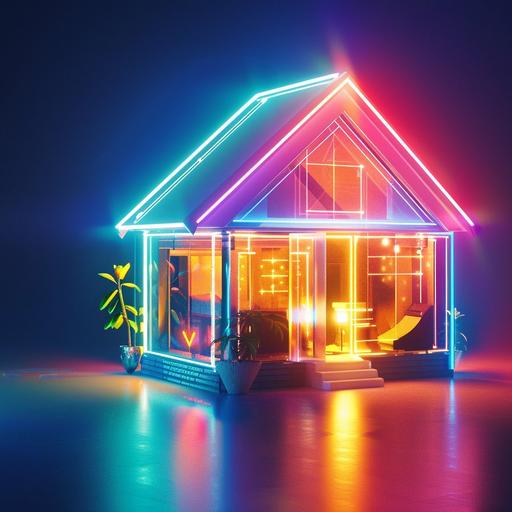} &
        \includegraphics[width=0.125\textwidth]{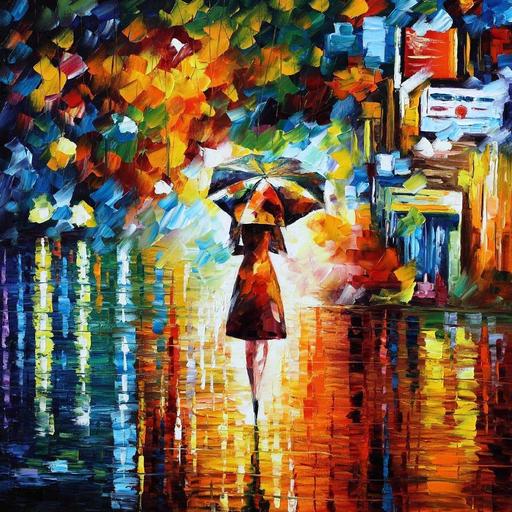} &
        \includegraphics[width=0.125\textwidth]{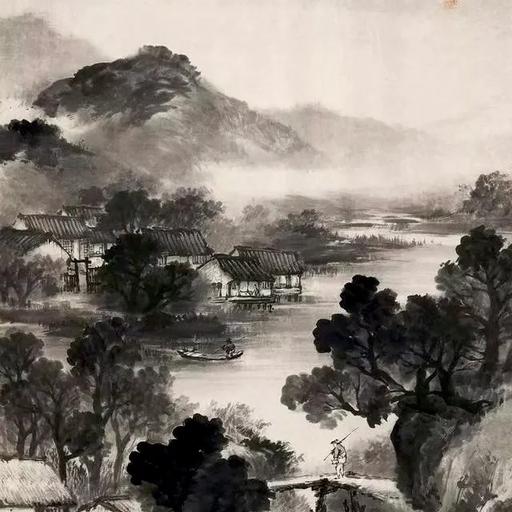} \\   
        
        \noalign{\vskip 0.02cm}\hline\noalign{\vskip 0.07cm}
        
        \includegraphics[width=0.125\textwidth]{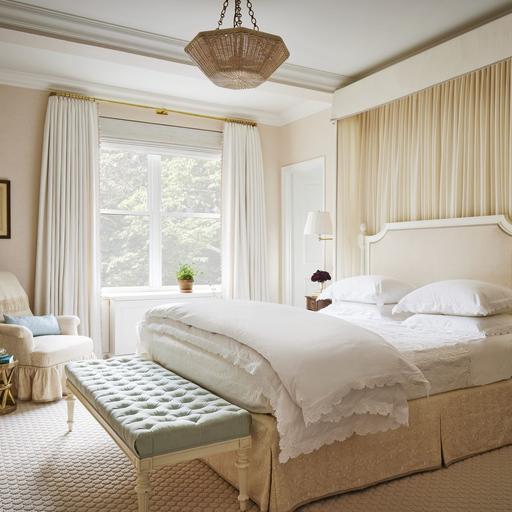} &
        \includegraphics[width=0.125\textwidth]{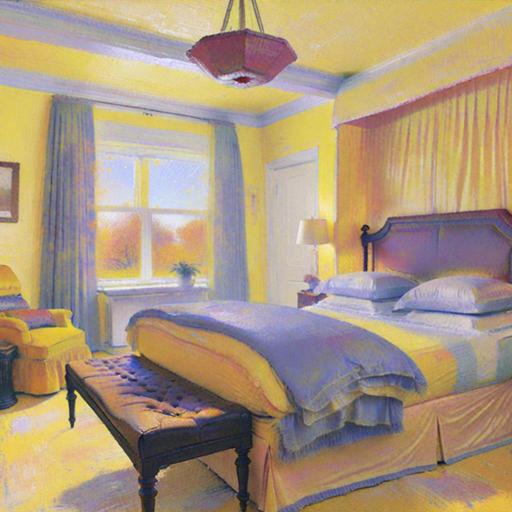} &
        \includegraphics[width=0.125\textwidth]{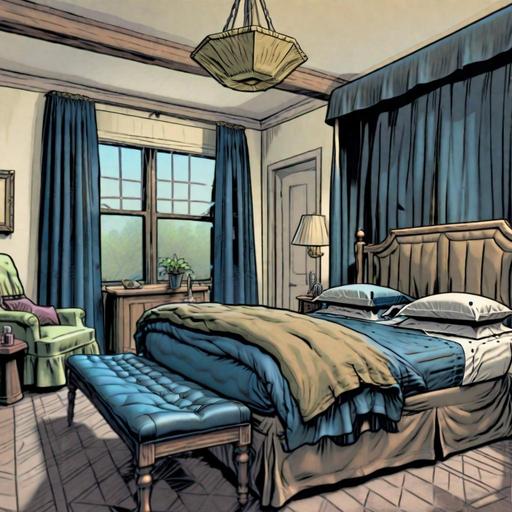} &
        \includegraphics[width=0.125\textwidth]{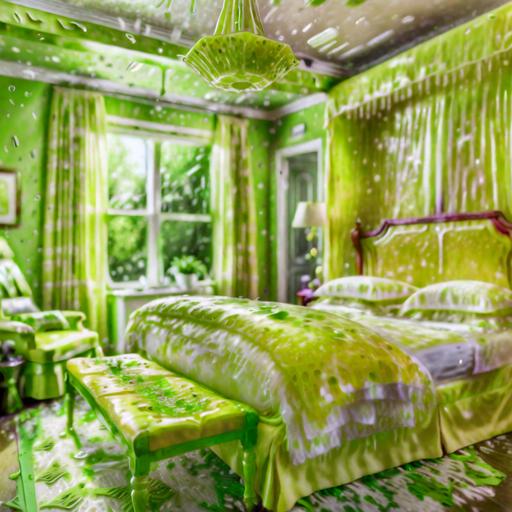} &
        \includegraphics[width=0.125\textwidth]{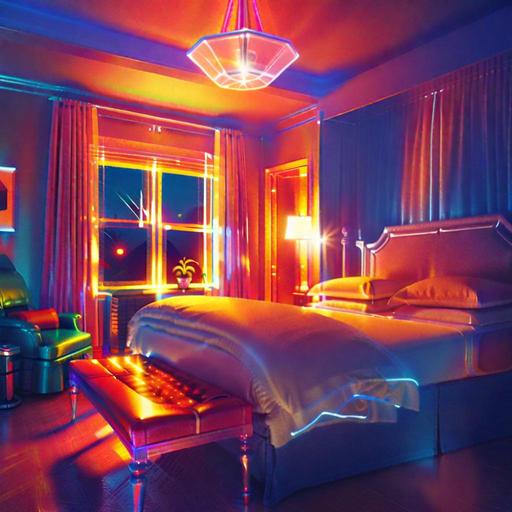} &
        \includegraphics[width=0.125\textwidth]{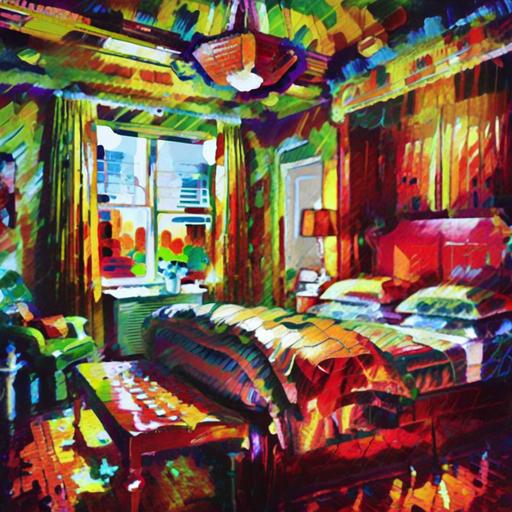} &
        \includegraphics[width=0.125\textwidth]{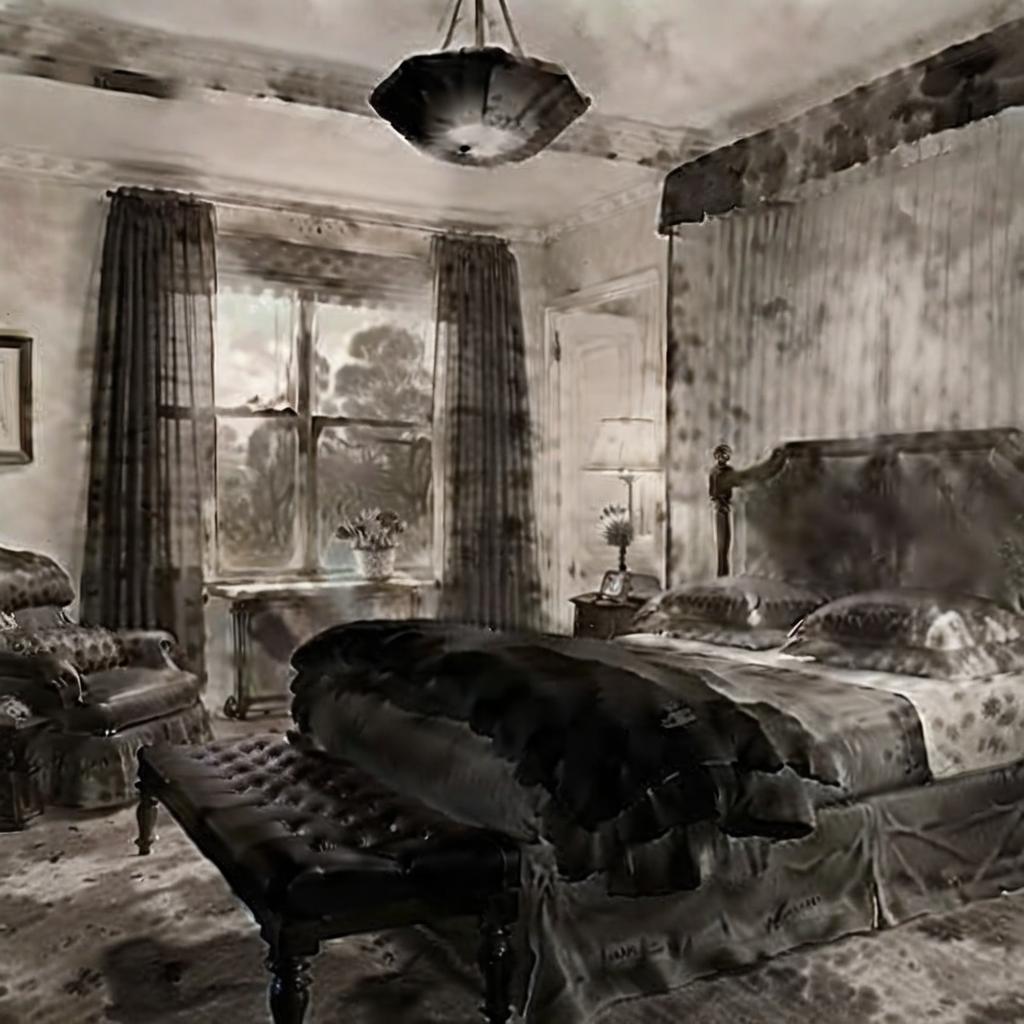} \\
        
        \includegraphics[width=0.125\textwidth] {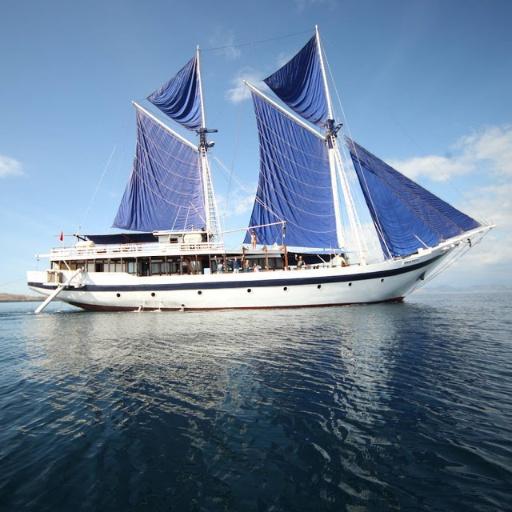} &
        \includegraphics[width=0.125\textwidth] {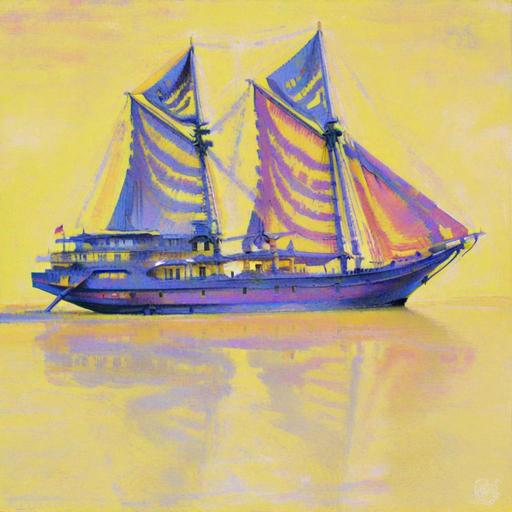} &
        \includegraphics[width=0.125\textwidth]{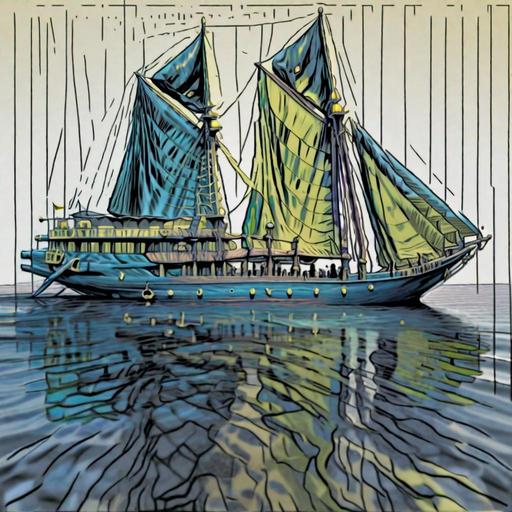} &
        \includegraphics[width=0.125\textwidth]{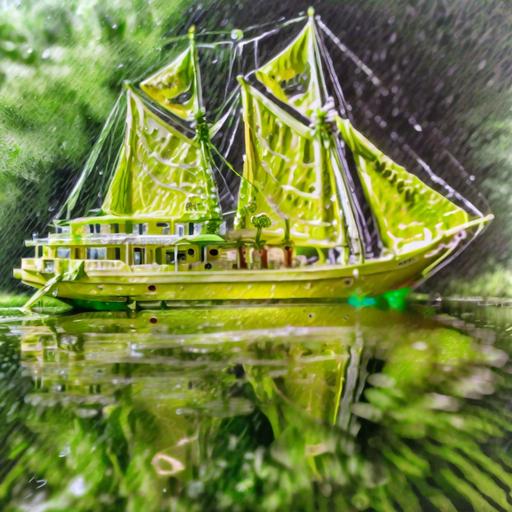} &
        \includegraphics[width=0.125\textwidth]{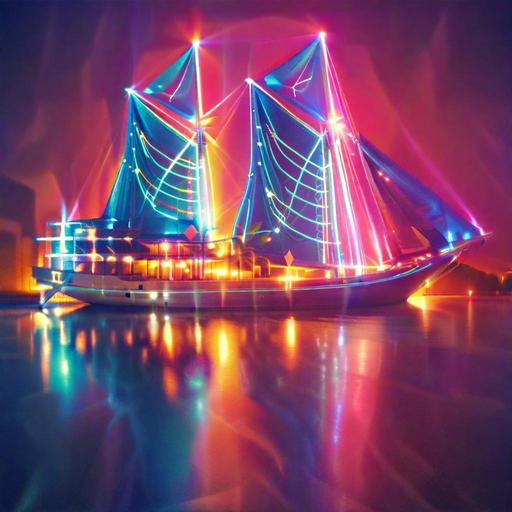} &
        \includegraphics[width=0.125\textwidth]{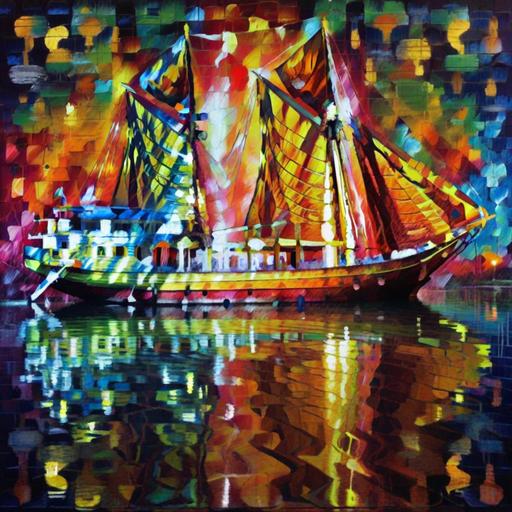} &
        \includegraphics[width=0.125\textwidth]{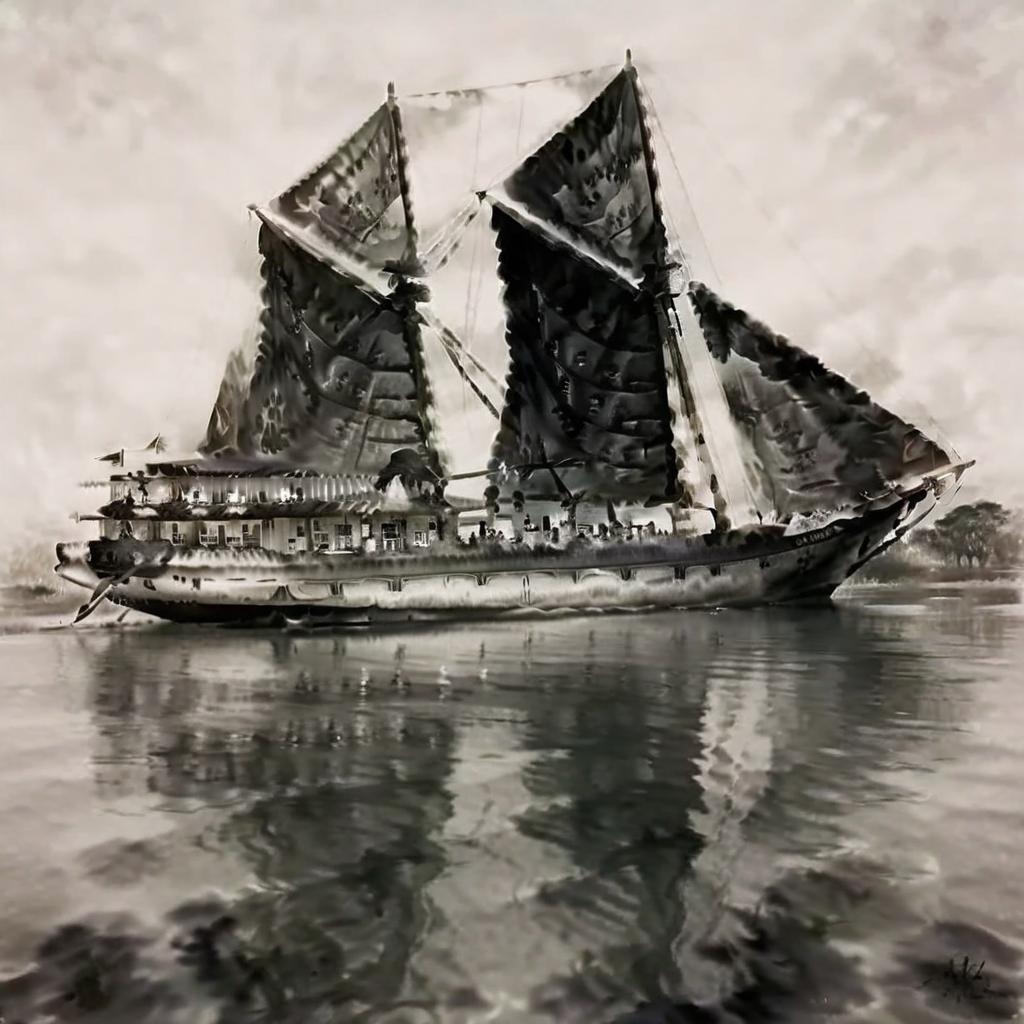}\\   
        
        \noalign{\vskip 0.02cm}\hline\noalign{\vskip 0.07cm}
    
        \multicolumn{1}{c}{Content input} & \multicolumn{1}{c}{``Pixel art''}  & ``Ice'' & ``Vintage''  & ``Cyberpunk'' &  ``Sketch cartoon'' &  ``Gothic dark'' \\

        \includegraphics[width=0.125\textwidth]{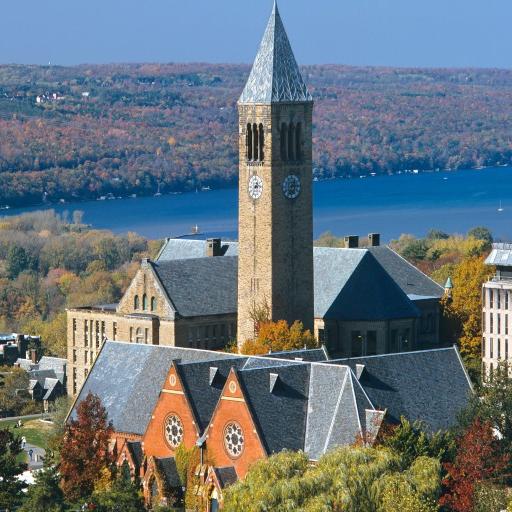} &
        \includegraphics[width=0.125\textwidth]{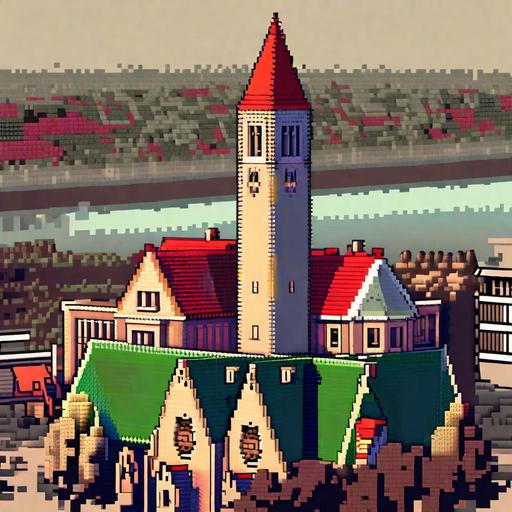} &
        \includegraphics[width=0.125\textwidth]{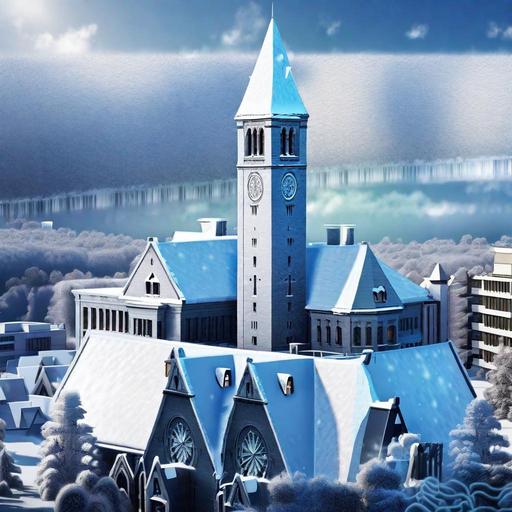} &
        \includegraphics[width=0.125\textwidth]{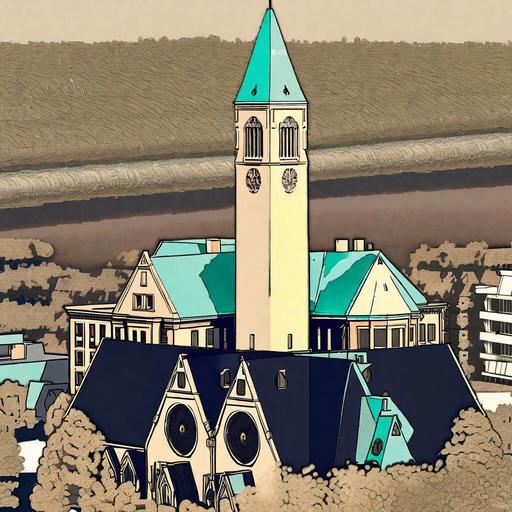} &
        \includegraphics[width=0.125\textwidth]{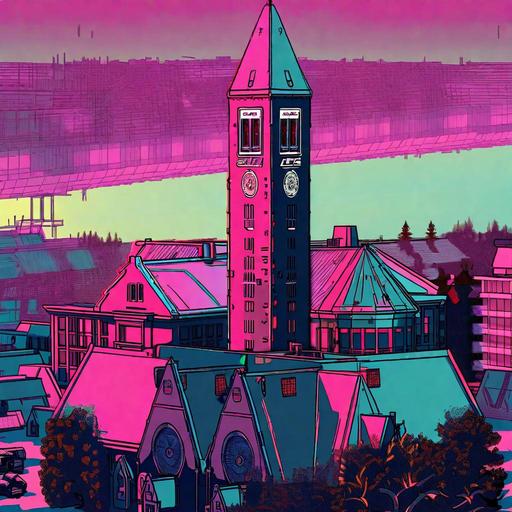} &
        \includegraphics[width=0.125\textwidth]{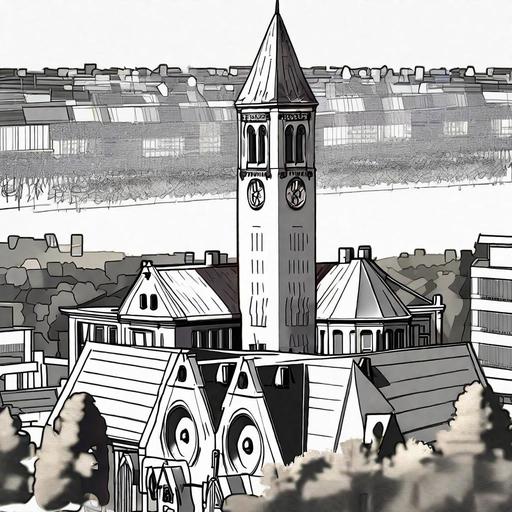} &
        \includegraphics[width=0.125\textwidth]{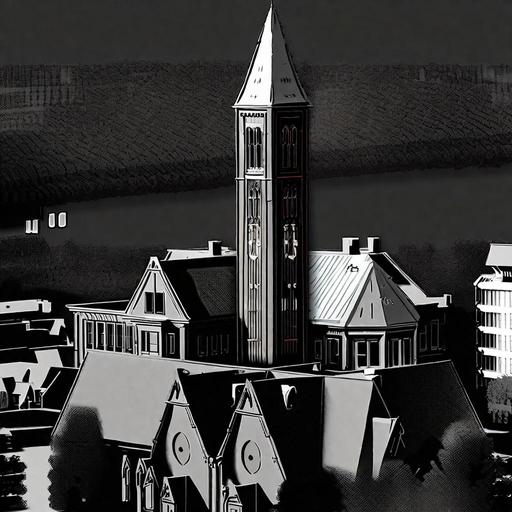} 
        \\ 

        \noalign{\vskip 0.02cm}\hline\noalign{\vskip 0.07cm}

        \multicolumn{1}{c}{Style input} & \multicolumn{1}{c}{``Dog''} & ``Car'' & ``Train'' & ``Bench'' & ``Classroom'' & ``Bedroom'' \\

        \includegraphics[width=0.125\textwidth]{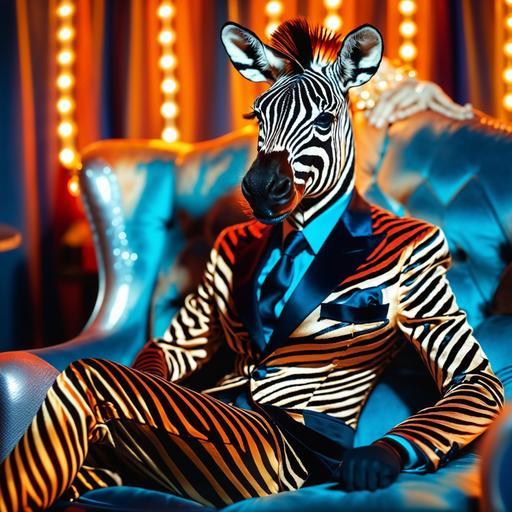} &
        \includegraphics[width=0.125\textwidth]{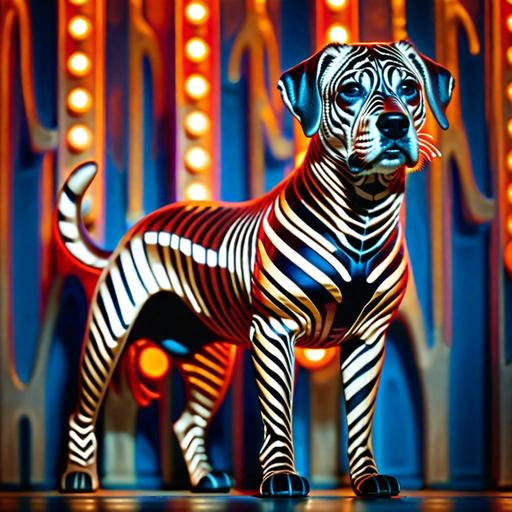} &
        \includegraphics[width=0.125\textwidth]{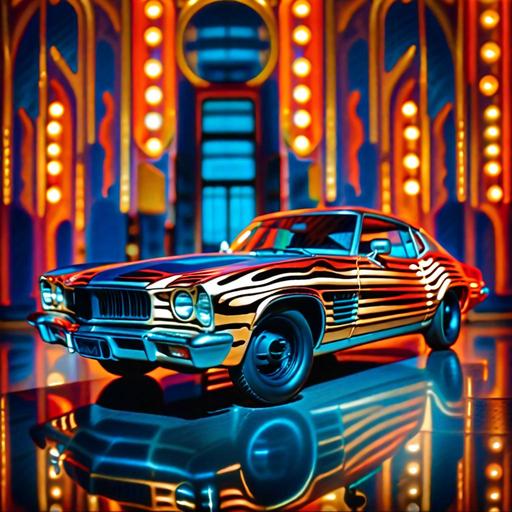} &
        \includegraphics[width=0.125\textwidth]{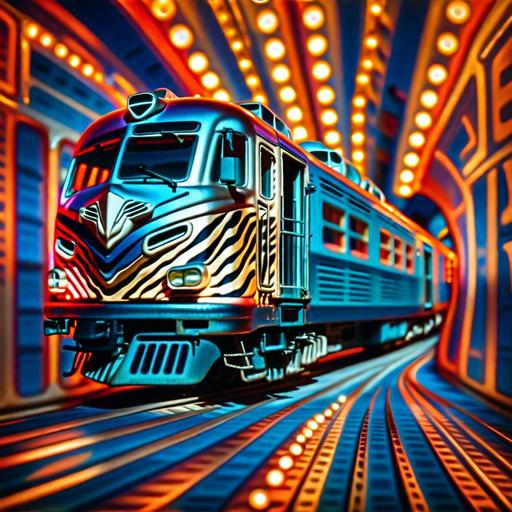} &
        \includegraphics[width=0.125\textwidth]{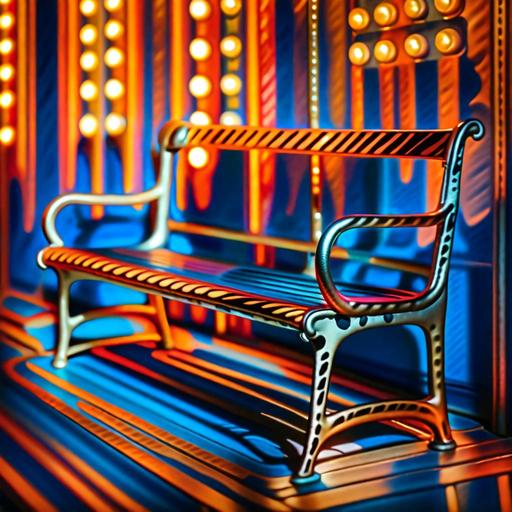} &
        \includegraphics[width=0.125\textwidth]{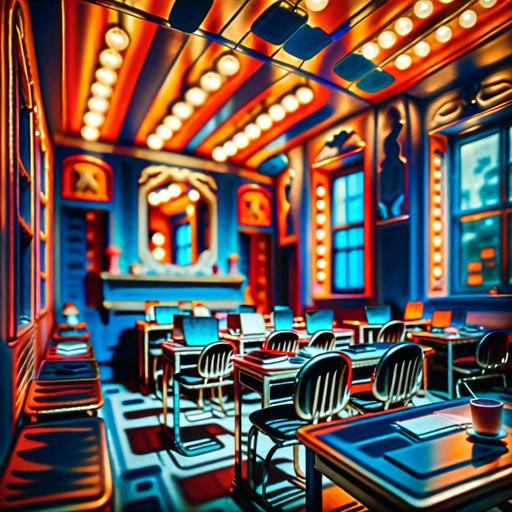} &
        \includegraphics[width=0.125\textwidth]{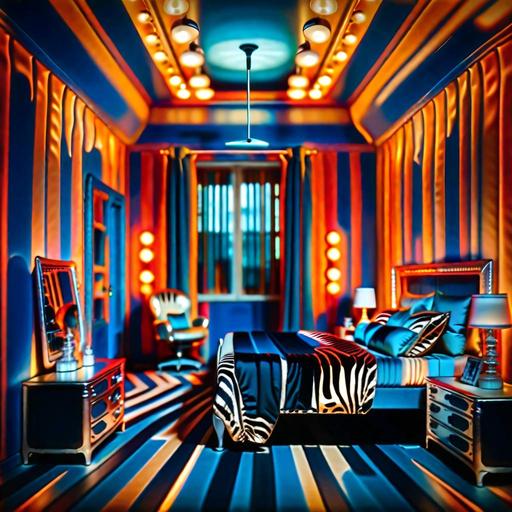} 
        \\ 
         
    \end{tabular}
    }   
    \caption{Results generated by ConsisLoRA for three image stylization tasks: (Top) Transferring the style from a reference image to the content of a target image; (Middle) Applying the style described by prompts to a content image; (Bottom) Generating objects described by prompts with the style extracted from a reference image.}
    \label{fig:qualitative_results}
    \vspace{-0.2cm}
\end{figure*}

\paragraph{Image Stylization Applications.} 
Similar to B-LoRA \cite{B-LoRA}, our method supports a variety of image stylization applications, as illustrated in \cref{fig:qualitative_results}. Style transfer is enabled through the integration of both content and style LoRAs, facilitating the creation of images that accurately reflect the desired content and style. Utilizing only the content LoRA enables text-based image stylization, which is controlled by a style prompt. Conversely, employing only the style LoRA allows for the generation of style-consistent images with any text-described content.

\subsection{Controlling with Inference Guidance}
\label{sec:guidance}
Drawing on the classifier-free guidance~\cite{cfg}, previous research has explored various inference guidance methods for tasks such as stylization~\cite{PairCustomization}, image editing~\cite{sine}, and compositional generation~\cite{compositional}. Inspired by these approaches, we introduce two guidance terms that allow for continuous control over content and style strengths during inference. Formally, after optimizing LoRAs over the content and style images, we obtain four distinct sets of LoRA weights: content and style LoRA weights from the content image (denoted as $\theta_\text{c}^\text{c}$ and $\theta_\text{c}^\text{s}$), and content and style LoRA weights from the style image (denoted as $\theta_\text{s}^\text{c}$ and $\theta_\text{s}^\text{s}$). Our inference algorithm is defined as follows:
\begin{equation} 
\begin{split}
\tilde{\epsilon} = \epsilon_{\theta_\text{c}^\text{c},\theta_\text{s}^\text{s}}(z_t, c) 
&+ \lambda_\text{cfg} \left( \epsilon_{\theta_\text{c}^\text{c},\theta_\text{s}^\text{s}}(z_t, c) - \epsilon_{\theta_\text{c}^\text{c},\theta_\text{s}^\text{s}}(z_t, \varnothing) \right) \\
&+ \lambda_\text{cont} ( \epsilon_{\theta_\text{c}^\text{c}}(z_t, c_\text{c}^\text{c}) - \epsilon_{\theta_\text{s}^\text{c}}(z_t, c_\text{s}^\text{c}) ) \\
&+ \lambda_\text{sty} ( \epsilon_{\theta_\text{s}^\text{s}}(z_t, c_\text{s}^\text{s}) - \epsilon_{\theta_\text{c}^\text{s}}(z_t, c_\text{c}^\text{s}) ),
\end{split}
\end{equation}
where $\lambda_\text{cfg} \left( \epsilon_{\theta_\text{c}^\text{c},\theta_\text{s}^\text{s}}(z_t, c) - \epsilon_{\theta_\text{c}^\text{c},\theta_\text{s}^\text{s}}(z_t, \varnothing) \right)$ is the classifier-free guidance term~\cite{cfg} with the LoRA weights $\theta_\text{c}^\text{c}$ and $\theta_\text{s}^\text{s}$, $\{\lambda_\text{cont},\lambda_\text{sty}\}$ control the strengths of the guidance, and $\{c_\text{c}^\text{c},c_\text{s}^\text{c},c_\text{s}^\text{s},c_\text{c}^\text{s}\}$ are the text conditioning vectors for the corresponding LoRAs. The content guidance term is defined as the difference between the noises of $\theta_\text{c}^\text{c}$ and $\theta_\text{s}^\text{c}$, used to enhance the content strength from the content image. Similarly, the style guidance term enhances the style strength from the style image. Note that this inference guidance is not applied in our experiments when comparing with the baselines to ensure a fair comparison.

\section{Experiments}

\subsection{Implementation and Evaluation Setup}

\paragraph{Implementation Details.}
Our implementation is based on SDXL v1.0~\cite{SDXL}, with both the model weights and text encoders frozen. The rank of LoRA weights is set to 64. All LoRAs are trained on a single image. For the content image, we initially train for 500 steps using $\epsilon$-prediction, then switch to $x_0$-prediction for an additional 1000 steps. For the style image, we first obtain its content LoRA using the above training strategy, and then separately train a new style LoRA for 1000 steps using $x_0$-prediction. The entire training process takes approximately 12 minutes on a single 4090 GPU. More implementation details of our method and the baselines are provided in Appendix~\ref{sec:appendix_implementation_details}.

\paragraph{Evaluation Setup.}
We compare our method with four state-of-the-art stylization methods, including StyleID~\cite{styleID}, StyleAligned~\cite{styleAligned}, ZipLoRA~\cite{ziplora}, and B-LoRA~\cite{B-LoRA}. For a fair comparison, we collect 20 content images and 20 style images from different studies~\cite{B-LoRA,styleID,instantStyle,styleDrop,dreambooth}. Using these images, we compose 400 pairs of content and style images for quantitative evaluation.

\subsection{Results}
\paragraph{Qualitative Evaluation.}
In \cref{fig:qualitative_comparison}, we present a visual comparison of style transfer results between our method and the baselines. As shown, the outputs produced by B-LoRA, ZipLoRA, and StyleAligned exhibit structural inconsistencies with the content image, as the $\epsilon$-prediction loss tends to capture broad concepts rather than the precise global structure. Moreover, as observed in the first and second rows, B-LoRA sometimes suffers from style misalignment and content leakage. ZipLoRA struggles to balance the merged content and style LoRAs, sometimes neglecting the style from the reference image. Although StyleID achieves good content preservation through DDIM inversion~\cite{ddim}, it often fails to accurately capture the style of the reference image, thereby diminishing the stylistic impact. The outputs from StyleAligned show significant structural inconsistencies with the content image and occasionally incorporate structural elements from the reference image. In contrast, our method generates content-consistent images with accurate stylization and effectively prevents content leakage. \cref{fig:qualitative_results} shows more results of different stylization applications using our method. Additional qualitative evaluation is provided in Appendix~\ref{sec:appendix_results}.

\begin{table}[t]
    \centering
    \setlength{\tabcolsep}{3pt}
    \caption{\textbf{Quantitative comparison}. We measure style and content alignment using DreamSim (DS) distance and cosine similarities calculated over CLIP and DINO features.}
    \begin{tabular}{>{\raggedright\arraybackslash}m{2.1cm}>{\centering\arraybackslash}m{1cm}>{\centering\arraybackslash}m{1cm}>{\centering\arraybackslash}m{1cm}>{\centering\arraybackslash}m{1cm}>{\centering\arraybackslash}m{1cm}}
        \toprule
        \multirow{2}{*}{\vspace{-1.5mm}Methods} & \multicolumn{2}{c}{Style Align.} & \multicolumn{3}{c}{Content Align.} \\
        \cmidrule(lr){2-3} \cmidrule(lr){4-6}
        & DS$\downarrow$ & CLIP$\uparrow$ & DINO$\uparrow$ & DS$\downarrow$ & CLIP$\uparrow$ \\
        \midrule
        StyleAlign~\cite{styleAligned} & 0.591 & 0.645 & 0.441 & 0.561 & 0.647 \\
        StyleID~\cite{styleID} & 0.653 & 0.638 & \textbf{0.679} & \textbf{0.494} & \textbf{0.693} \\
        ZipLoRA~\cite{ziplora} & 0.646 & 0.643 & 0.488 & 0.543 & 0.668 \\
        B-LoRA~\cite{B-LoRA} & \underline{0.573} & \underline{0.654} & 0.536 & 0.568 & 0.643 \\
        Ours & \textbf{0.567} & \textbf{0.659} & \underline{0.629} & \underline{0.524} & \underline{0.671} \\
        \bottomrule
    \end{tabular}
    \label{tab:quantitative}
    \vspace{-0.25cm}    
\end{table}

\paragraph{Quantitative Evaluation.}
We conduct a quantitative evaluation of each method in terms of style and content alignment. Style alignment between the generated and reference images is measured using the DreamSim distance~\cite{dream_sim} and CLIP score~\cite{clip}. Content alignment between the generated and content images is assessed using the DINO score~\cite{dino}, DreamSim distance, and CLIP score. Each method is evaluated across 400 pairs of style and content images, with the results detailed in \cref{tab:quantitative}. StyleID achieves the best performance in content alignment but ranks lowest in style alignment. This is consistent with the qualitative observations, where StyleID often diminishes the stylistic impact. Besides this extreme case, our method outperforms all baselines in both style and content alignment. In particular, compared to B-LoRA, our method shows significant improvements in content alignment, especially evident in the DINO score. While B-LoRA achieves a CLIP score comparable to ours in style alignment, this score may be inflated due to content leakage from the reference image.

\paragraph{User Study.}
We also conducted a user study to evaluate our method. In this study, participants were presented with a content image, a reference image, and two stylized images: one generated by our method and the other by a baseline method. Participants were tasked with selecting the image that better aligns with the style of the reference image while preserving the content of the content image. We collected a total of 1,500 responses from 50 participants, as detailed in \cref{tab:user_study}. The results demonstrate a strong preference for our method.

\begin{table}
    \centering
    \caption{\textbf{User Study.} Participants were presented with two images: one by our method and another by a baseline method. The results demonstrate a clear preference for our method.}
    \resizebox{1.01\linewidth}{!}{
    \begin{tabular}{l c c }
      \toprule
      Baselines  & Prefer Baseline    & Prefer Ours  \\
      \midrule
      B-LoRA~\cite{B-LoRA}    & 24.3\%  & \textbf{75.7\%}  \\
      ZipLoRA~\cite{ziplora}  & 11.7\%  & \textbf{88.3\%}  \\
      StyleID~\cite{styleID}  & 19.2\%  & \textbf{80.8\%}  \\
      StyleAligned~\cite{styleAligned}  & 10.4\%  & \textbf{89.6\%}  \\
      \bottomrule
    \end{tabular}

    }
    \label{tab:user_study}
    \vspace{-0.25cm}
\end{table}

\begin{figure}[t]
 \centering
 \includegraphics[width=1.01\linewidth]{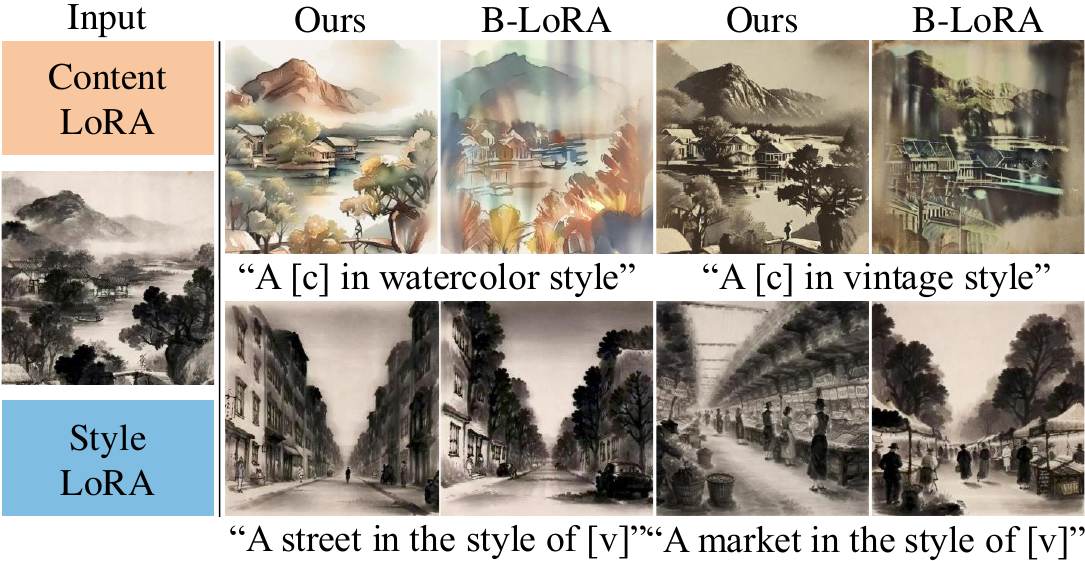}
\caption{\textbf{Content and style decomposition.} Our method achieves a more accurate and disentangled decomposition of content and style compared to the baseline method.}
\label{fig:decomposition}
\end{figure}

\paragraph{Content and Style Decomposition.} \label{sec:decomposition}
Given a single input image, we compare our method with B-LoRA for content and style decomposition, applying the content and style LoRAs separately, as illustrated in \cref{fig:decomposition}. When B-LoRA employs content LoRA with new styles described by text prompts, it struggles to preserve the global structure of the input image and fails to align the generated images with the specified styles in the prompts. Additionally, B-LoRA is unable to learn a disentangled style LoRA from the input image, resulting in severe content leakage issues in the generated images. In contrast, our method effectively disentangles the content and style of the input image, demonstrating clear advantages in content and style decomposition. More decomposition results are provided in Appendix~\ref{sec:appendix_decomposition}.

\begin{figure}[t]
    \centering
    \setlength{\tabcolsep}{0.3pt}
    \renewcommand{\arraystretch}{0.6}
    {\small
    \begin{tabular}{c@{\hspace{0.05cm}} | @{\hspace{0.05cm}}c c c c}

          \multicolumn{1}{c@{}}{Input}
         & \multicolumn{1}{c}{Original}
         & Content$\uparrow$
         & Style$\uparrow$
         & C\&S$\uparrow$ \\

        \includegraphics[width=0.0525\textwidth]{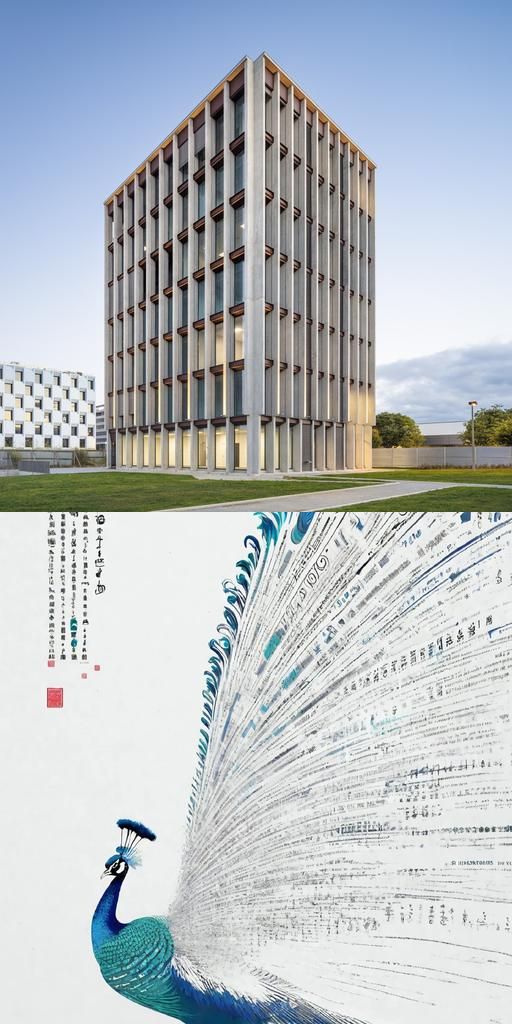} &
        \includegraphics[width=0.105\textwidth]{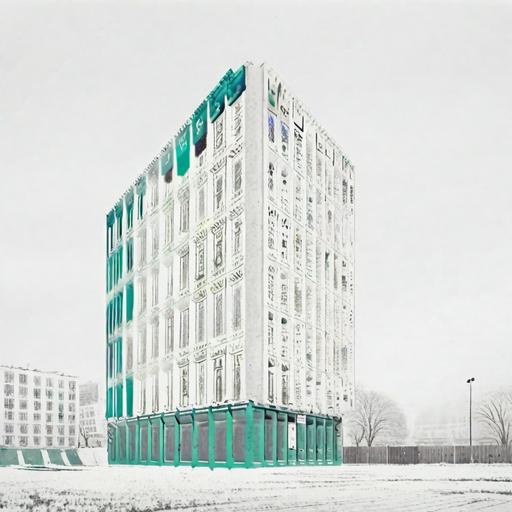} &
        \includegraphics[width=0.105\textwidth]{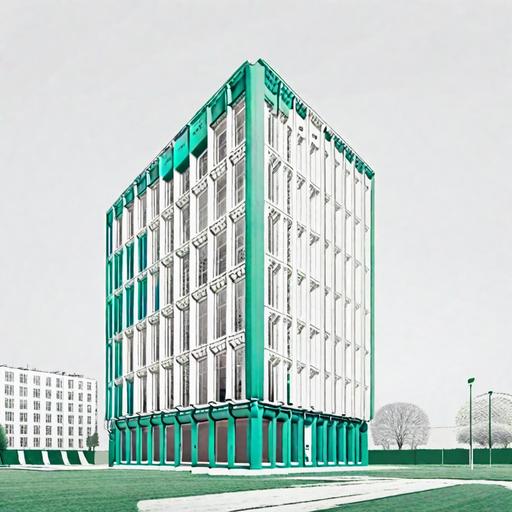} &
        \includegraphics[width=0.105\textwidth]{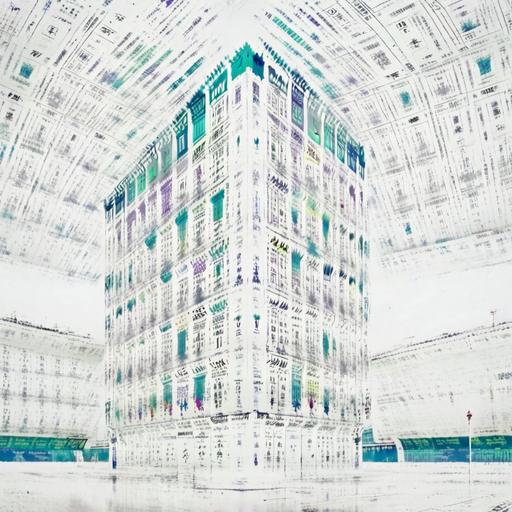} &
        \includegraphics[width=0.105\textwidth]{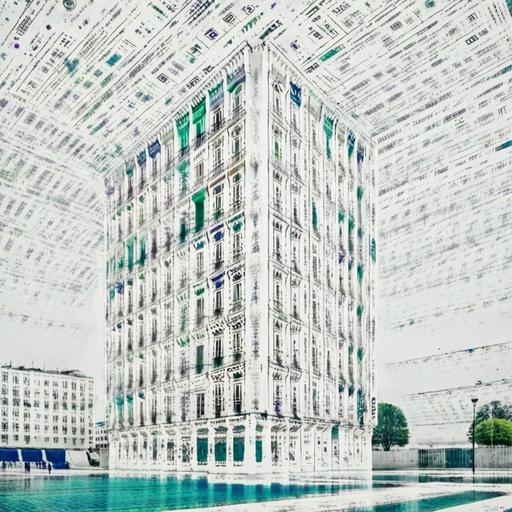} 
        \\ 
        
        \includegraphics[width=0.0525\textwidth]{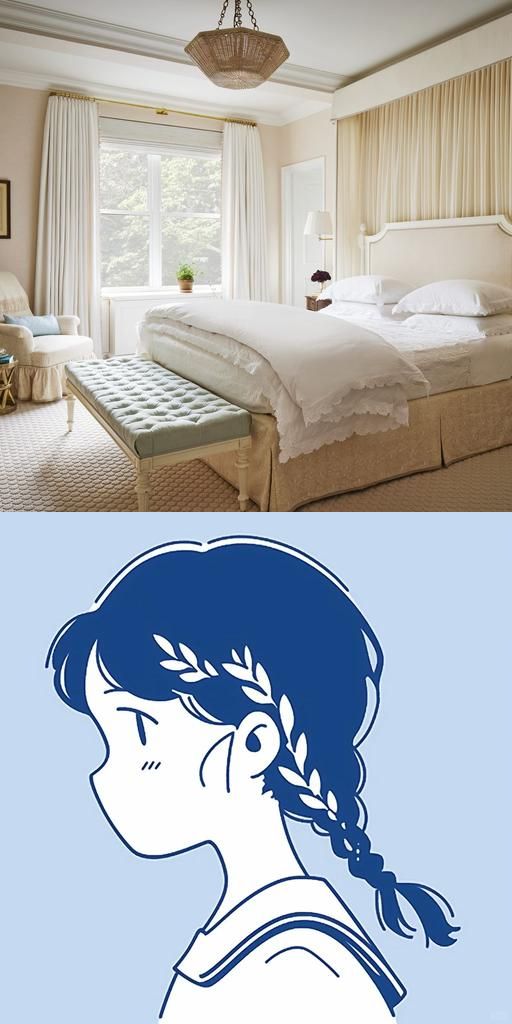} &
        \includegraphics[width=0.105\textwidth]{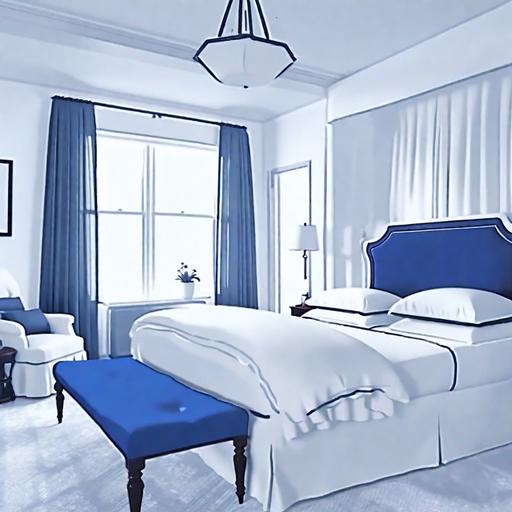} &
        \includegraphics[width=0.105\textwidth]{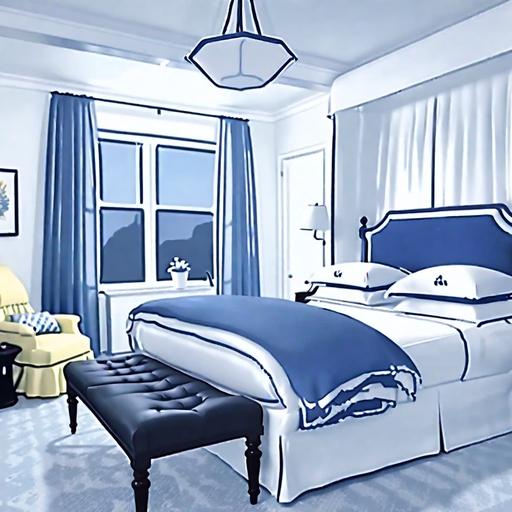} &
        \includegraphics[width=0.105\textwidth]{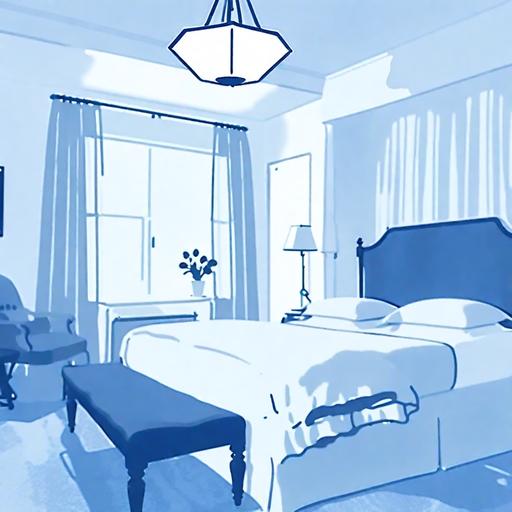} &
        \includegraphics[width=0.105\textwidth]{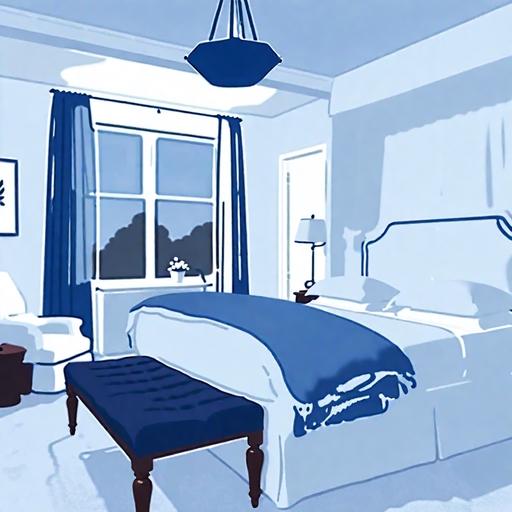} 
        \\ 
         
    \end{tabular}
    }   
    \caption{\textbf{Inference guidance.} Increasing the strengths of content and style guidance correspondingly enhances their impact on the generated images. Zoom in for best view.}
    \vspace{-0.3cm}
    \label{fig:guidance}
\end{figure}

\subsection{Inference Guidance} \label{sec:experiments_guidance}
In this section, we evaluate the proposed inference guidance described in \cref{sec:guidance} for controlling content and style strengths during inference. As shown in \cref{fig:guidance}, increasing the content and style strengths correspondingly enhances their impact on the generated images. Furthermore, a detailed comparison between our inference guidance and the approach of scaling LoRA weights is provided in Appendix~\ref{sec:appendix_guidance}. We observe that our method more effectively preserves the content structure when adjusting content strength. In terms of adjusting style strength, both methods are capable of generating high-quality stylized images.

\subsection{Ablation Study}
\label{sec:ablation}
\begin{figure}[t]
    \centering
    \setlength{\tabcolsep}{0.3pt}
    \renewcommand{\arraystretch}{0.6}
    {\small
    \begin{tabular}{c@{\hspace{0.05cm}} | @{\hspace{0.05cm}}c c c c}

         \multicolumn{1}{c@{}}{Input}
         & \multicolumn{1}{c}{Var A}
         & Var B
         & Var C
         & Full \\

        \includegraphics[width=0.0525\textwidth]{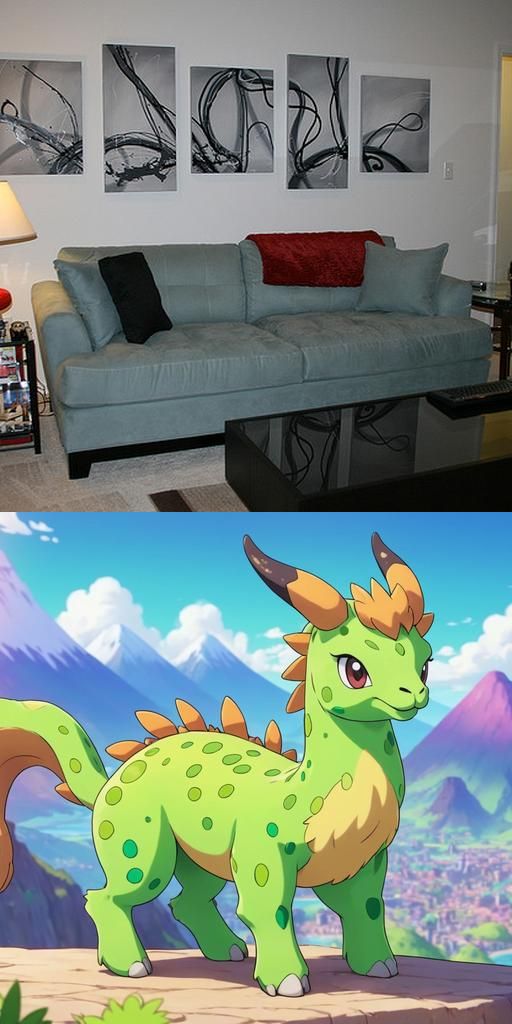} &
        \includegraphics[width=0.105\textwidth]{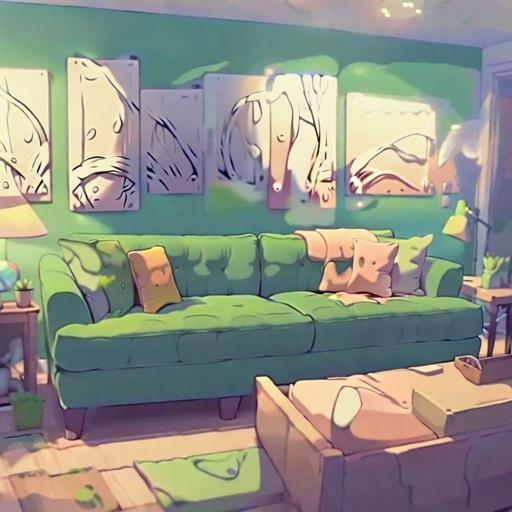} &
        \includegraphics[width=0.105\textwidth]{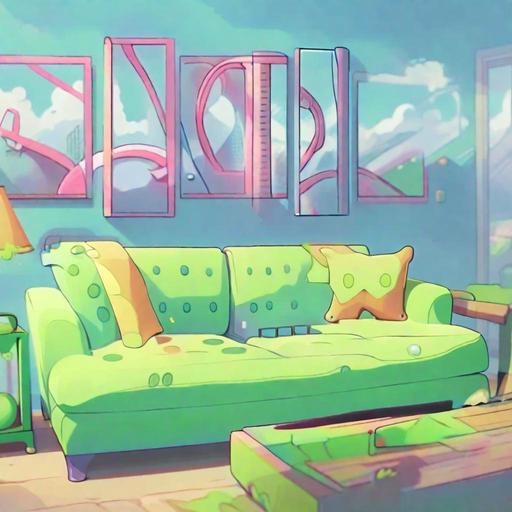} &
        \includegraphics[width=0.105\textwidth]{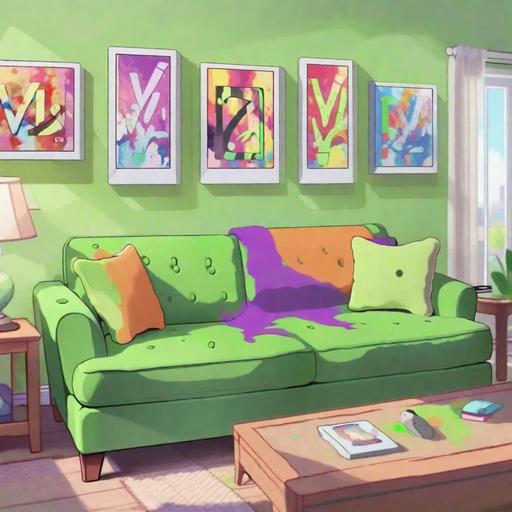} &
        \includegraphics[width=0.105\textwidth]{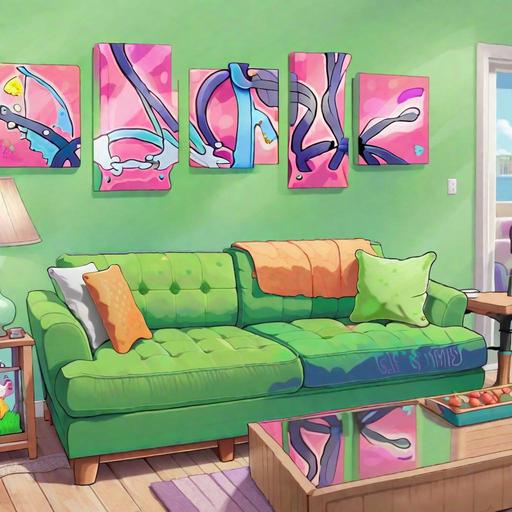} 
        \\ 
        
        \includegraphics[width=0.0525\textwidth]{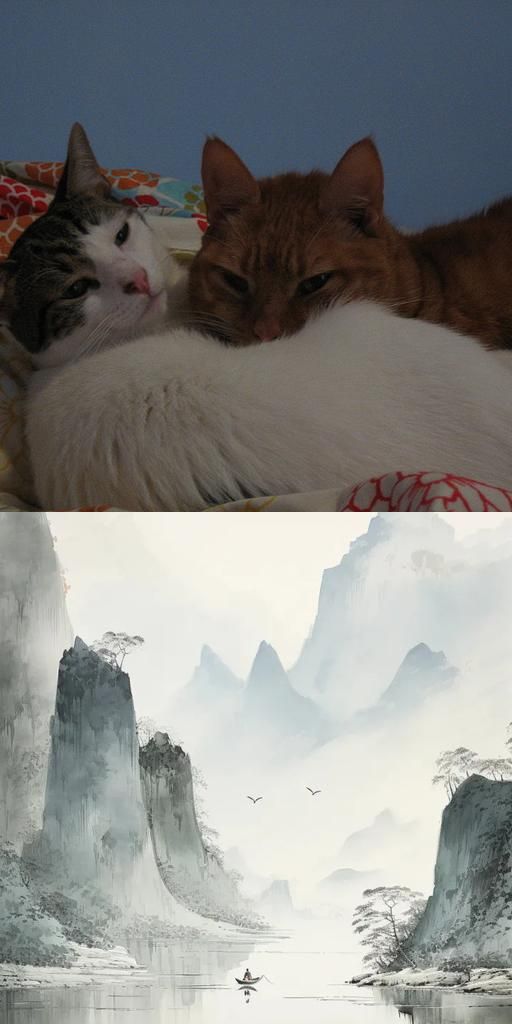} &
        \includegraphics[width=0.105\textwidth]{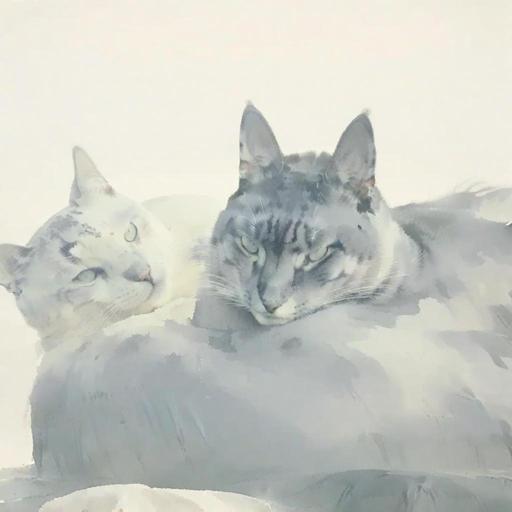} &
        \includegraphics[width=0.105\textwidth]{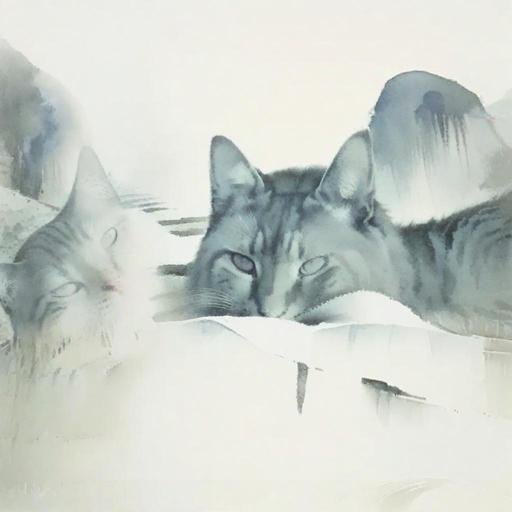} &
        \includegraphics[width=0.105\textwidth]{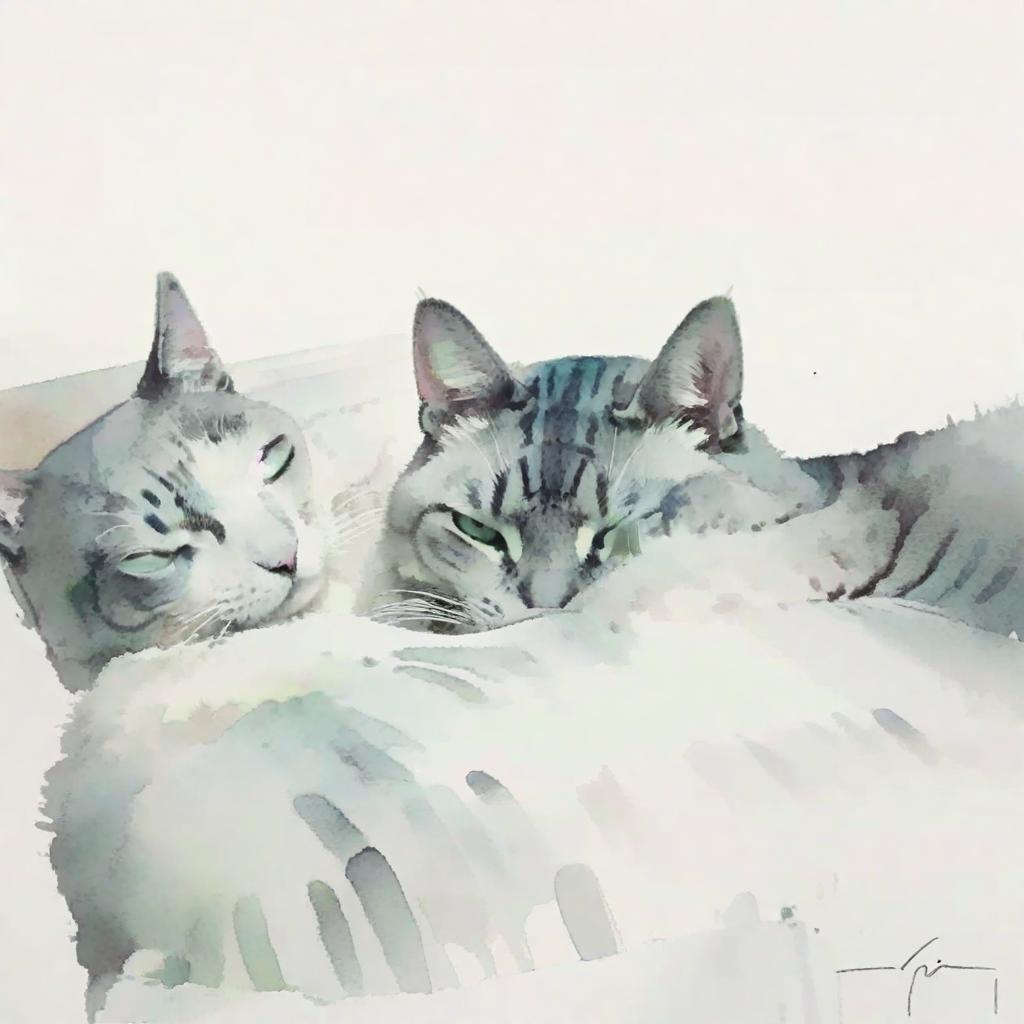} &
        \includegraphics[width=0.105\textwidth]{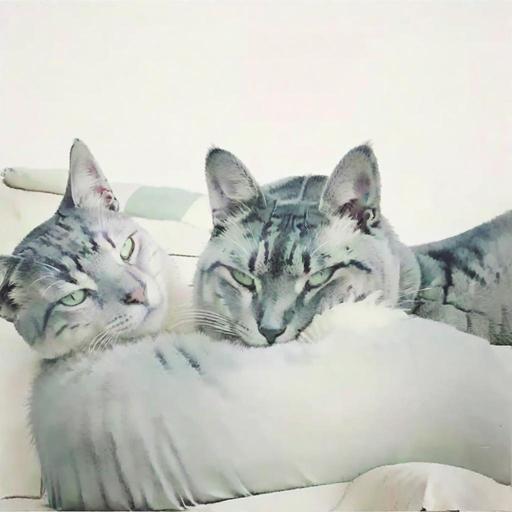} 
        \\ 
         
    \end{tabular}
    }   
    \caption{\textbf{Ablation study.} We evaluate three variants of our model: 1) replacing $x_0$-prediction with $\epsilon$-prediction (Var A), 2) removing the two-step training strategy for style LoRA (Var B), and 3) using $x_0$-prediction alone instead of loss transition for content LoRA (Var C). Zoom in for best view.}
    \vspace{-0.3cm}
    \label{fig:ablation}
\end{figure}

We conduct an ablation study to evaluate the effectiveness of each component of our method. Specifically, we assess three variants: 1) replacing $x_0$-prediction with $\epsilon$-prediction, 2) removing the two-step training strategy for style LoRA, and 3) employing $x_0$-prediction alone instead of loss transition for content LoRA. \cref{fig:ablation} presents a visual comparison of the stylized images generated by each variant. The results underscore the crucial role of each component. Without using $x_0$-prediction, the model fails to capture both the global structure of the content image and style features from the style image. Removing the two-step training strategy for style LoRA leads to significant content leakage issues. Moreover, employing $x_0$-prediction alone for content LoRA causes the model to struggle with capturing local details (e.g., the pictures hanging on the wall in the top row). More results of the ablation study are provided in Appendix~\ref{sec:appendix_ablation}.

\section{Conclusions and Limitations}
In this study, we introduced ConsisLoRA, a style transfer method designed to address critical challenges faced by existing LoRA-based methods, such as content inconsistency, style misalignment, and content leakage. By optimizing LoRA weights to predict the original image rather than noise, our approach significantly enhances both style and content consistency. Our two-step training strategy effectively separates the learning of content and style, facilitating the disentanglement of these elements. Additionally, our stepwise loss transition strategy ensures the preservation of both global structures and local details of the content image. Despite these advancements, our approach does exhibit some limitations. First, similar to other LoRA-based methods, our content LoRA often neglects the color of objects, a factor that may be crucial for certain applications. Second, our method faces challenges in preserving the identity of individuals, due to the limited capacity of LoRAs. We aim to focus our efforts on enhancing identity preservation of individuals in our future work.

{
    \small
    \bibliographystyle{ieeenat_fullname}
    \bibliography{main}
}

\clearpage
\appendix

\section{Implementation Details} \label{sec:appendix_implementation_details}
Our model leverages SDXL v1.0~\cite{SDXL}, with the model weights and text encoders fixed. We employ the Adam optimizer to tune the LoRA weights with a rank of 64. For the content image, training initially involves a learning rate of $2 \times 10^{-4}$ for 500 steps using the $\epsilon$-prediction loss, followed by a transition to the $x_0$-prediction loss at a learning rate of $1 \times 10^{-4}$ for an additional 1000 steps. For the style image, we first establish its content LoRA using the aforementioned strategy, then train a new style LoRA from scratch for 1000 steps using $x_0$-prediction. The entire training process requires approximately 12 minutes on a single Nvidia 4090 GPU. For baselines, including B-LoRA~\cite{B-LoRA}, StyleID~\cite{styleID}, and StyleAligned~\cite{styleAligned}, we utilize their official implementations. In the absence of an official implementation for ZipLoRA~\cite{ziplora}, we rely on a community-developed version~\cite{ziplora_github}. Since StyleAligned focuses solely on consistent style generation, we follow~\cite{styleAligned,B-LoRA} to use DreamBooth-LoRA~\cite{dreambooth-lora} to provide the content.

\section{Additional Qualitative Results} \label{sec:appendix_results}
\begin{figure*}[t]
    \centering
    \setlength{\tabcolsep}{0.85pt}
    \renewcommand{\arraystretch}{0.5}
    {\small
    \begin{tabular}{c c@{\hspace{0.1cm}} | @{\hspace{0.1cm}}c c c c c}
 
    Content & \multicolumn{1}{c@{}}{Style} & \multicolumn{1}{c}{Ours} & B-LoRA & ZipLoRA & StyleID & StyleAligned \\

    \includegraphics[width=0.137\textwidth]{images/cnt-sty/content/teddy_bear.jpg} &
    \includegraphics[width=0.137\textwidth]{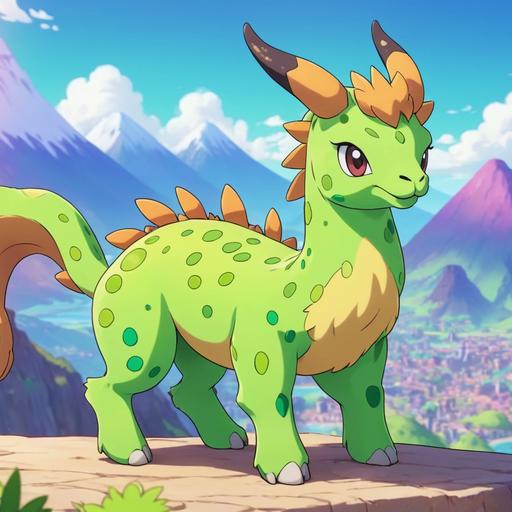} &
    \includegraphics[width=0.137\textwidth]{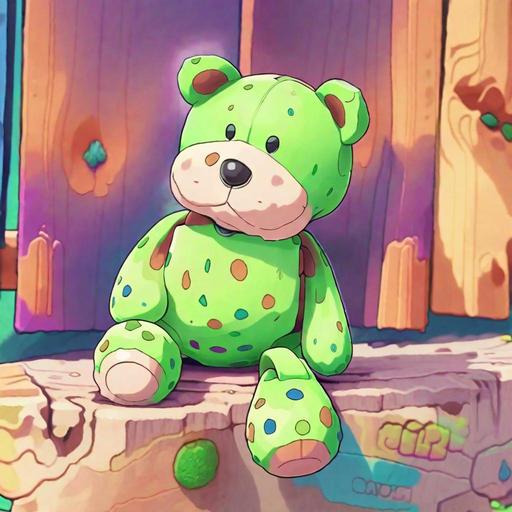} &
    \includegraphics[width=0.137\textwidth]{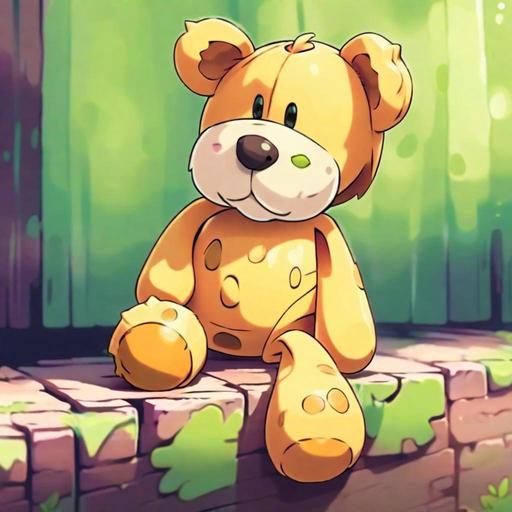} &
    \includegraphics[width=0.137\textwidth]{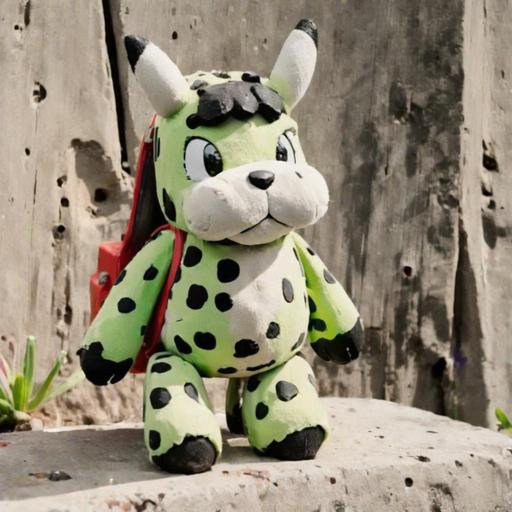} &
    \includegraphics[width=0.137\textwidth]{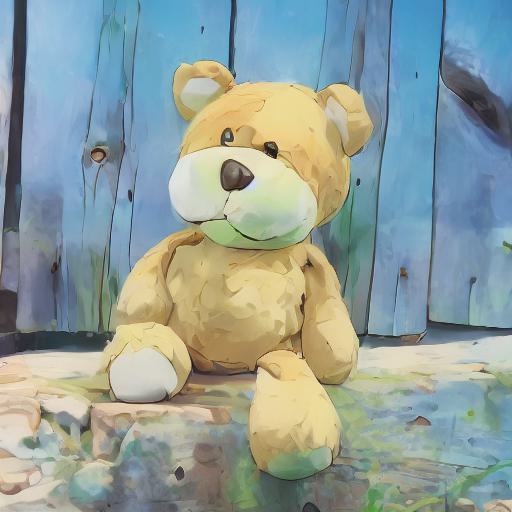} &
    \includegraphics[width=0.137\textwidth]{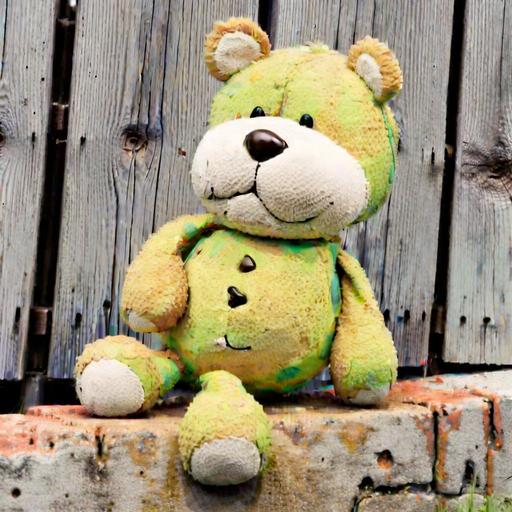} \\

    \includegraphics[width=0.137\textwidth]{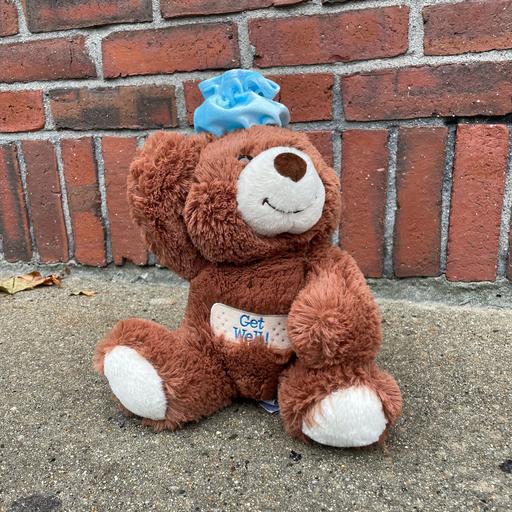} &
    \includegraphics[width=0.137\textwidth]{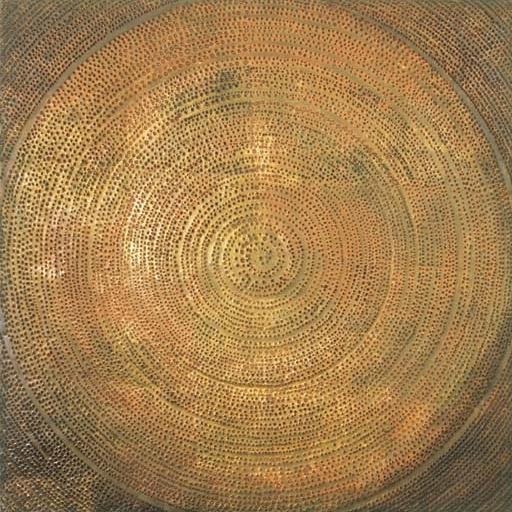} &
    \includegraphics[width=0.137\textwidth]{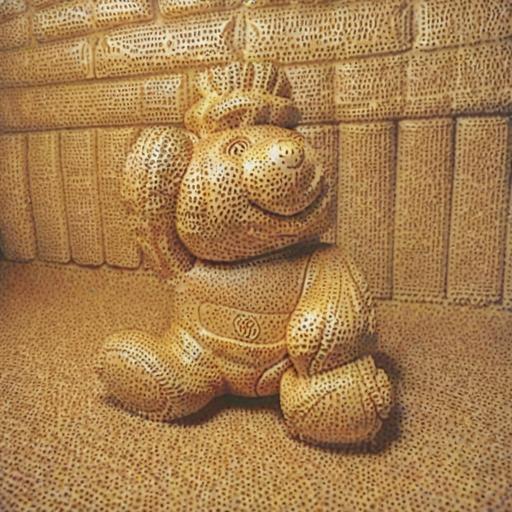} &
    \includegraphics[width=0.137\textwidth]{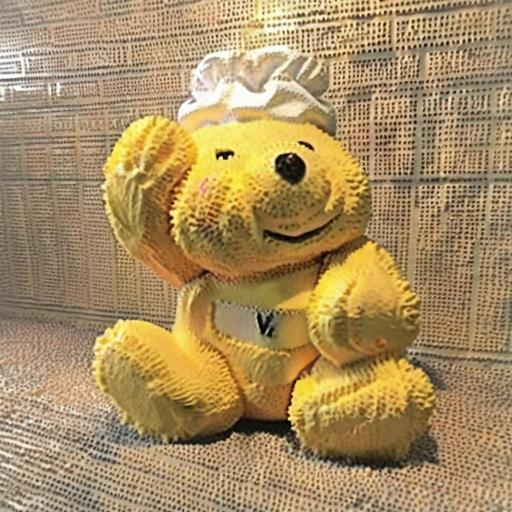} &
    \includegraphics[width=0.137\textwidth]{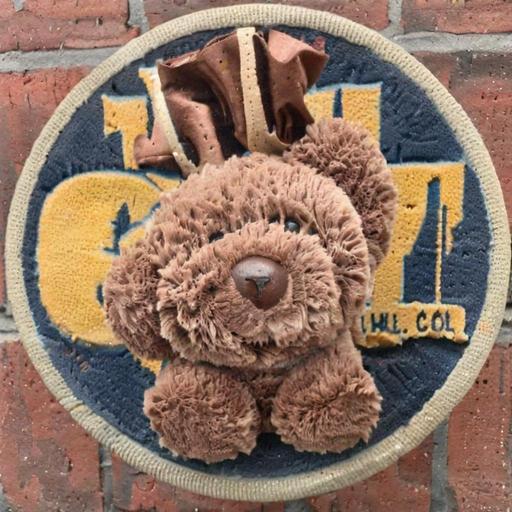} &
    \includegraphics[width=0.137\textwidth]{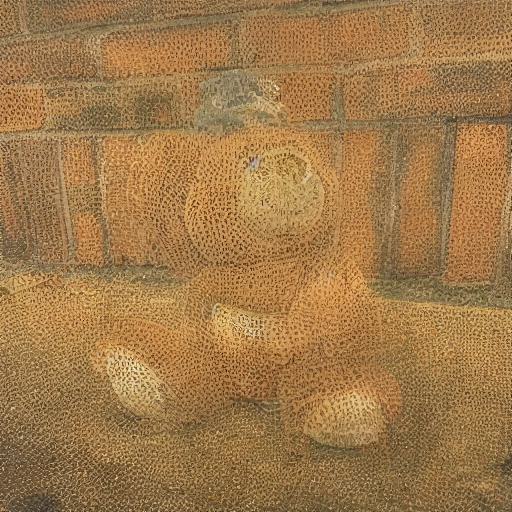} &
    \includegraphics[width=0.137\textwidth]{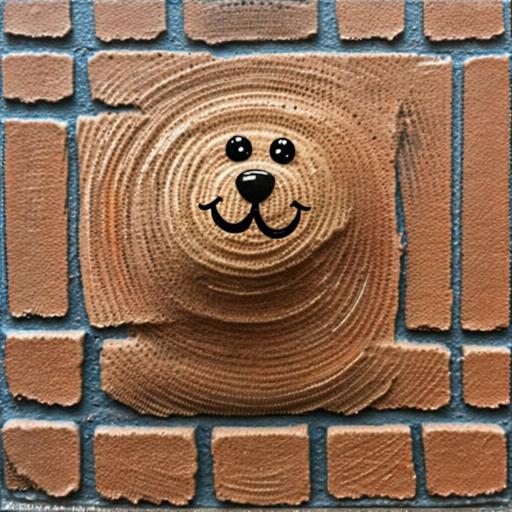} \\

    \includegraphics[width=0.137\textwidth]{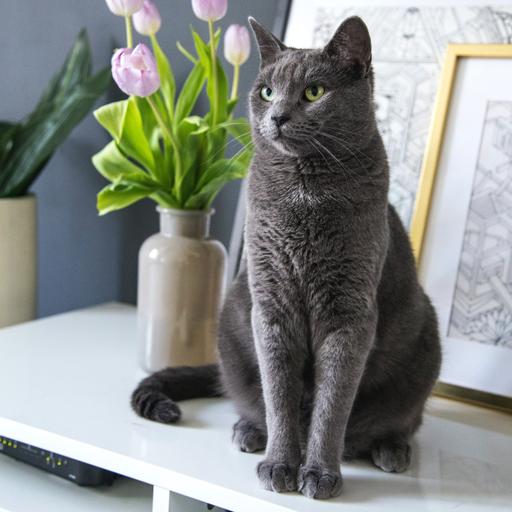} &
    \includegraphics[width=0.137\textwidth]{images/cnt-sty/style/zebra.jpg} &
    \includegraphics[width=0.137\textwidth]{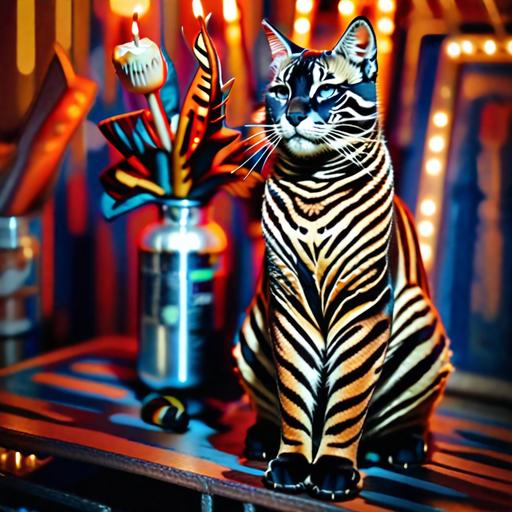} &
    \includegraphics[width=0.137\textwidth]{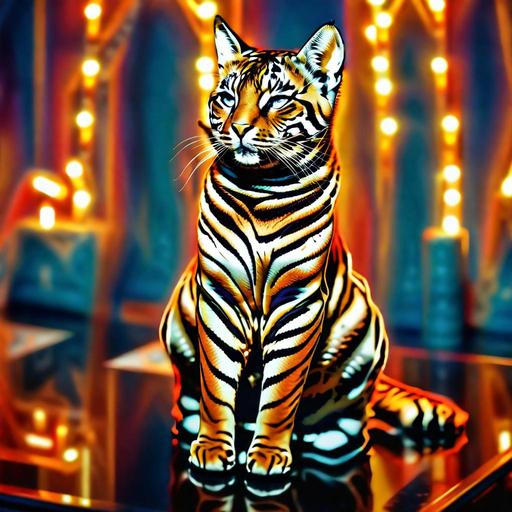} &
    \includegraphics[width=0.137\textwidth]{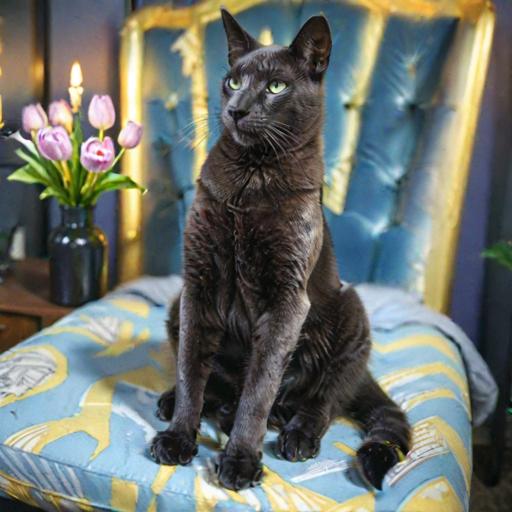} & 
    \includegraphics[width=0.137\textwidth]{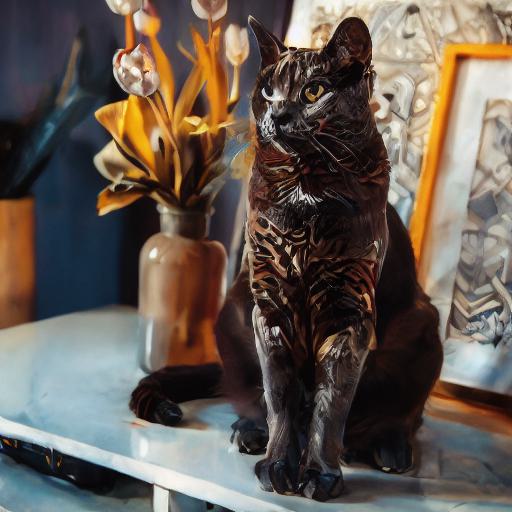} &
    \includegraphics[width=0.137\textwidth]{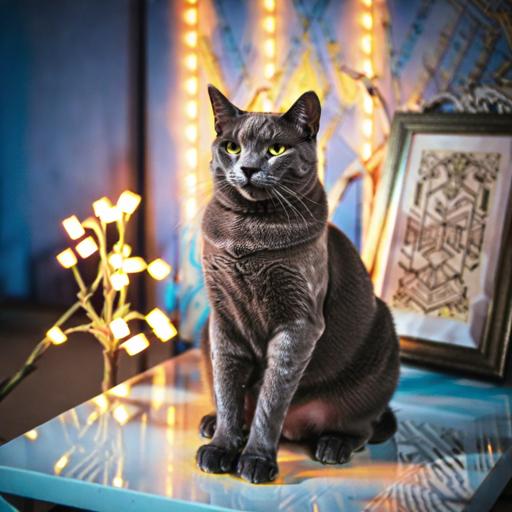} \\

    \includegraphics[width=0.137\textwidth]{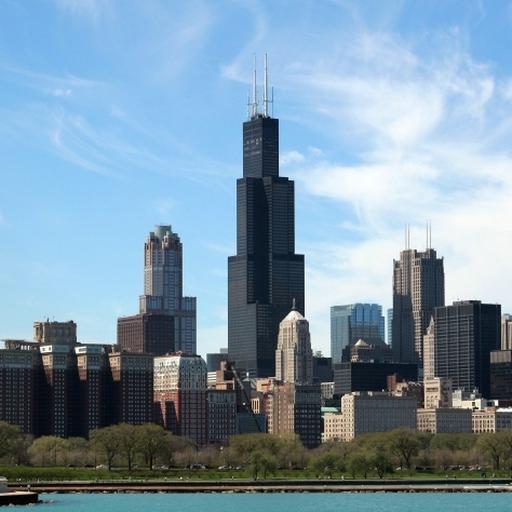} &
    \includegraphics[width=0.137\textwidth]{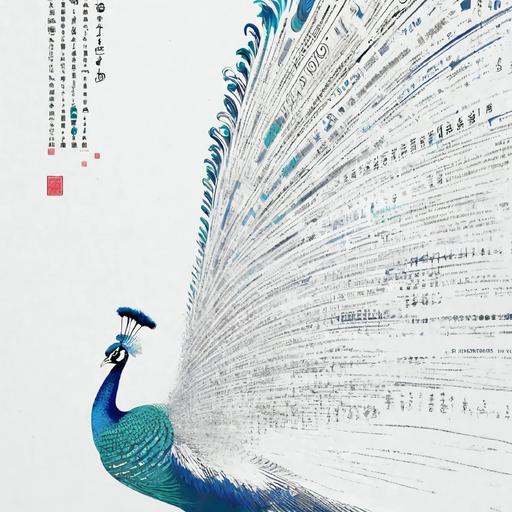} &
    \includegraphics[width=0.137\textwidth]{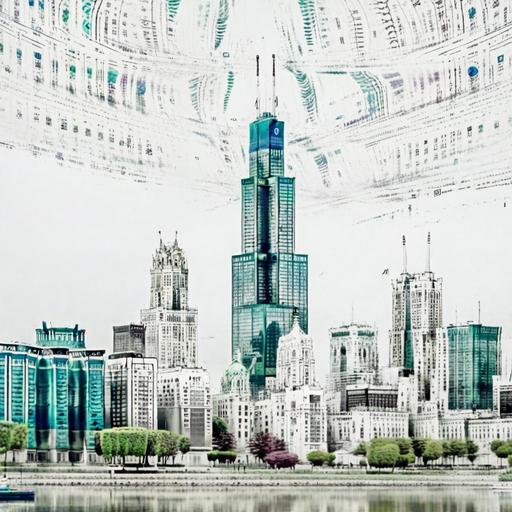} &
    \includegraphics[width=0.137\textwidth]{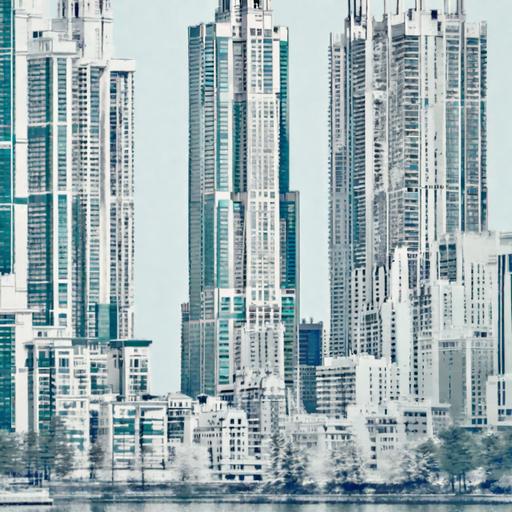} &
    \includegraphics[width=0.137\textwidth]{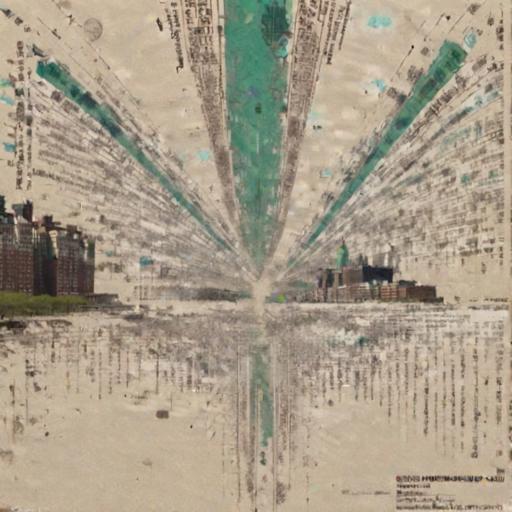} &
    \includegraphics[width=0.137\textwidth]{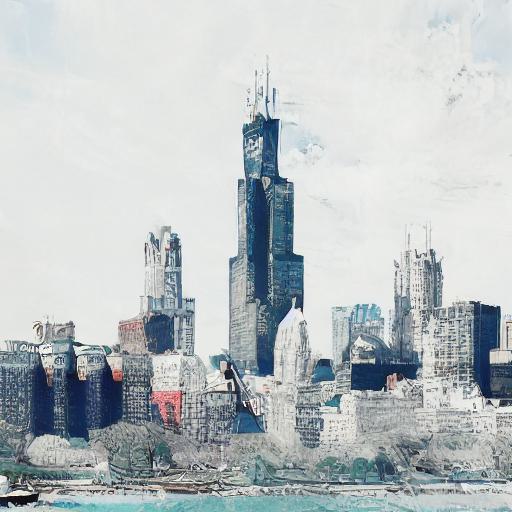} &
    \includegraphics[width=0.137\textwidth]{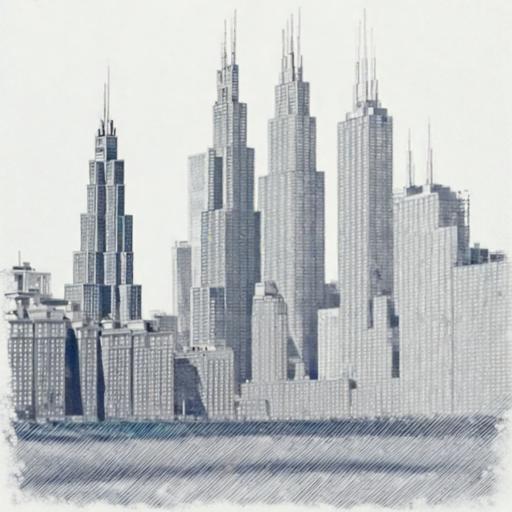} \\

    \includegraphics[width=0.137\textwidth]{images/cnt-sty/content/cornell.jpg} &
    \includegraphics[width=0.137\textwidth]{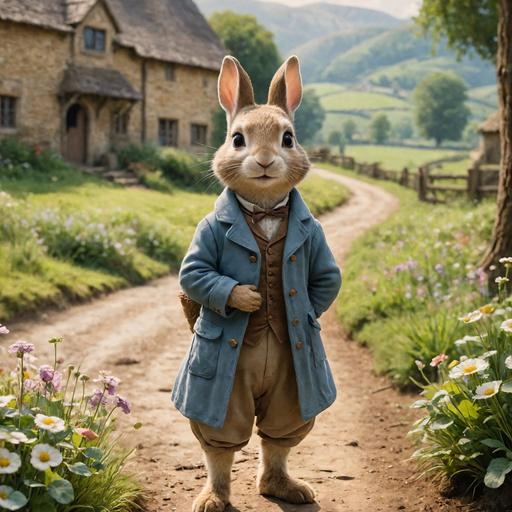} &
    \includegraphics[width=0.137\textwidth]{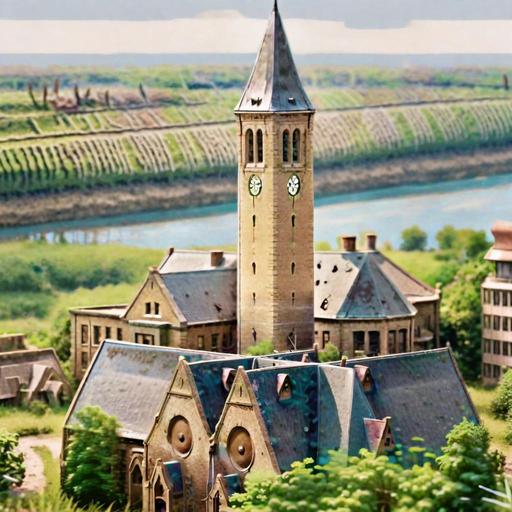} &
    \includegraphics[width=0.137\textwidth]{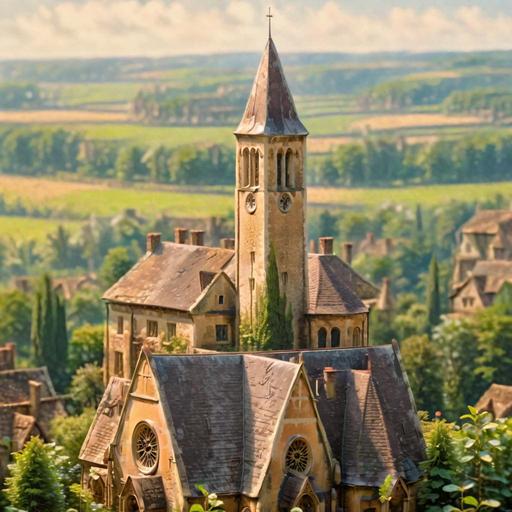} &
    \includegraphics[width=0.137\textwidth]{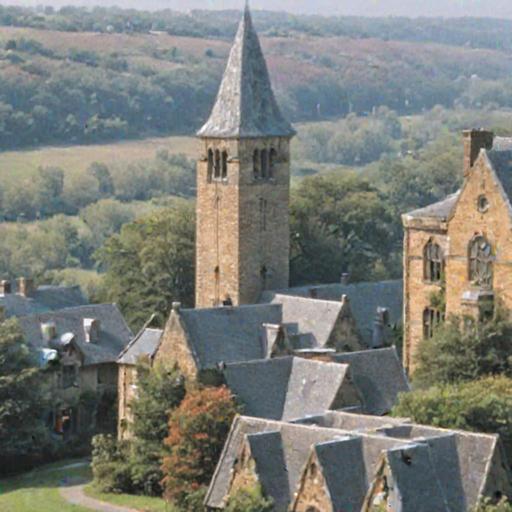} &
    \includegraphics[width=0.137\textwidth]{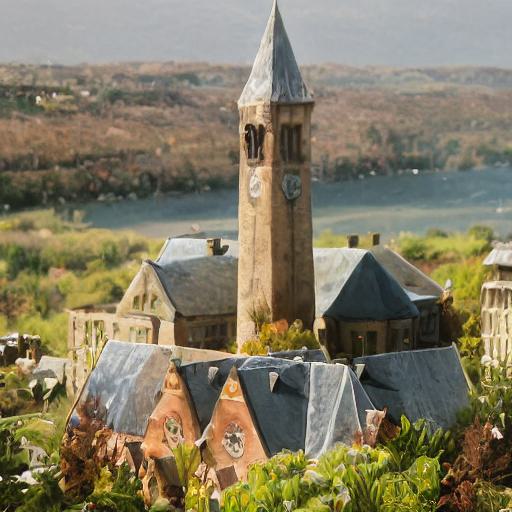} &
    \includegraphics[width=0.137\textwidth]{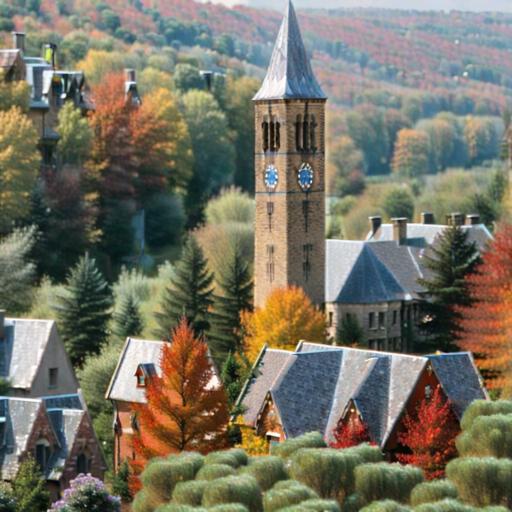} \\
    
    \includegraphics[width=0.137\textwidth]{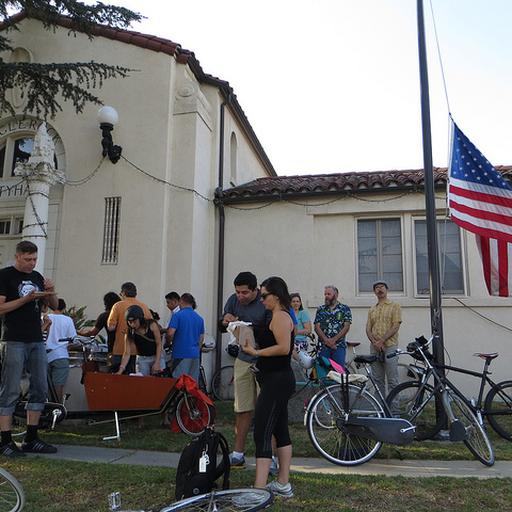} &
    \includegraphics[width=0.137\textwidth]{images/cnt-sty/style/pirate.jpg} &
    \includegraphics[width=0.137\textwidth]{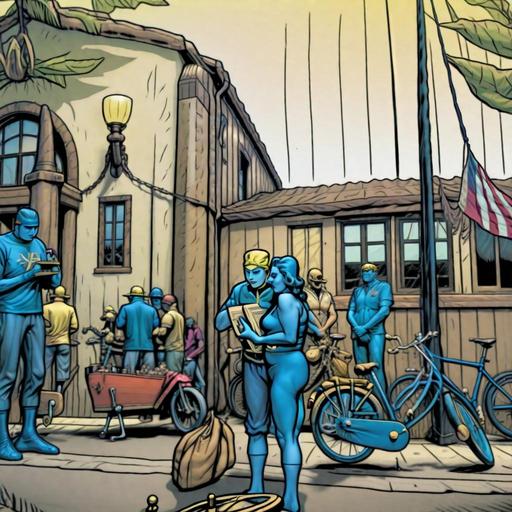} &
    \includegraphics[width=0.137\textwidth]{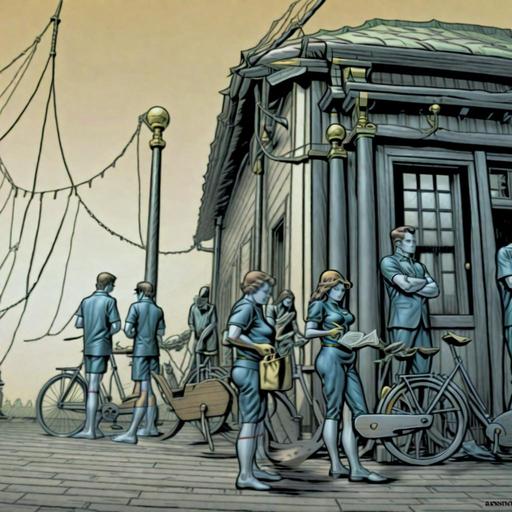} &
    \includegraphics[width=0.137\textwidth]{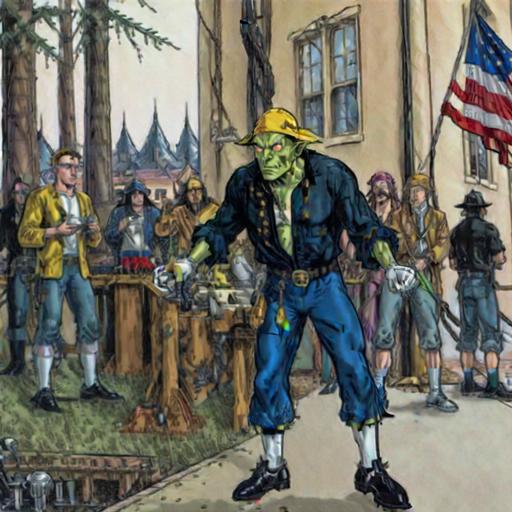} &
    \includegraphics[width=0.137\textwidth]{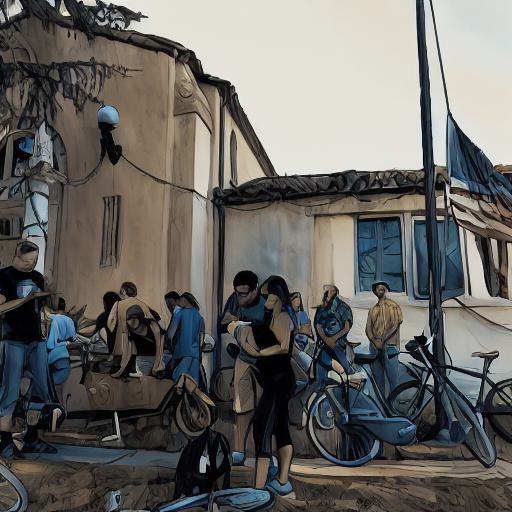} &
    \includegraphics[width=0.137\textwidth]{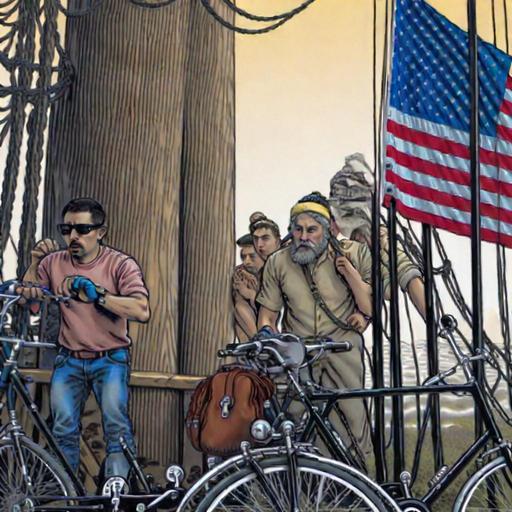} \\

    \includegraphics[width=0.137\textwidth]{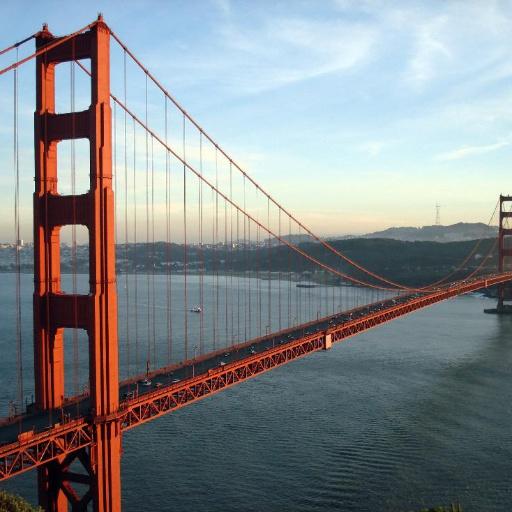} &
    \includegraphics[width=0.137\textwidth]{images/cnt-sty/style/pig.jpg} &
    \includegraphics[width=0.137\textwidth]{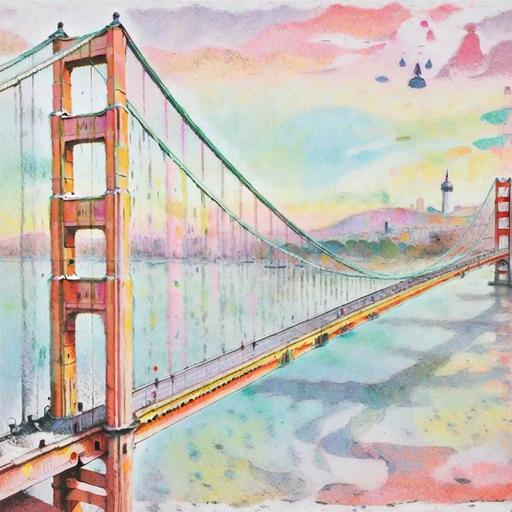} &
    \includegraphics[width=0.137\textwidth]{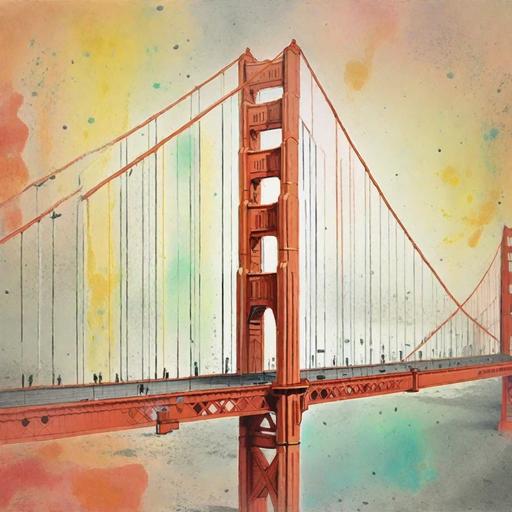} &
    \includegraphics[width=0.137\textwidth]{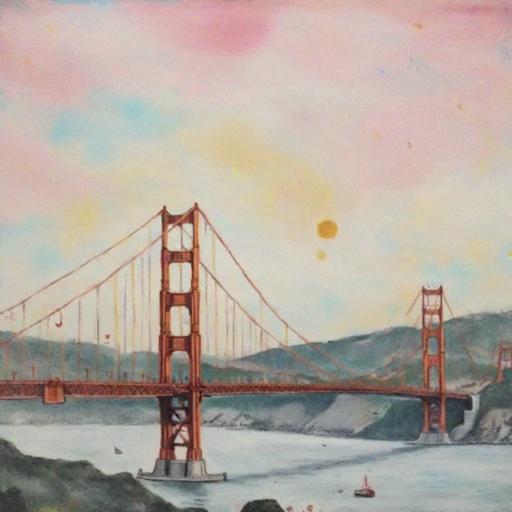} &
    \includegraphics[width=0.137\textwidth]{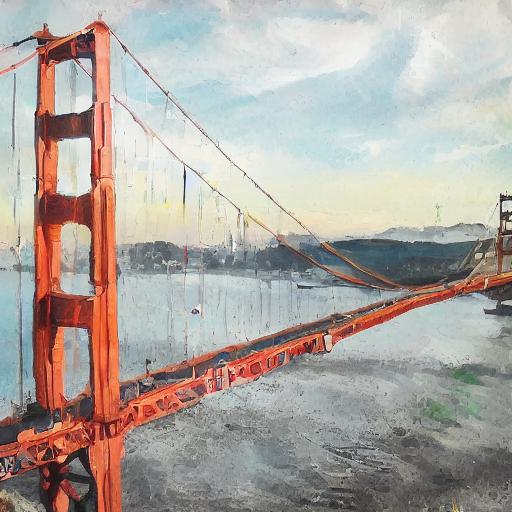} &
    \includegraphics[width=0.137\textwidth]{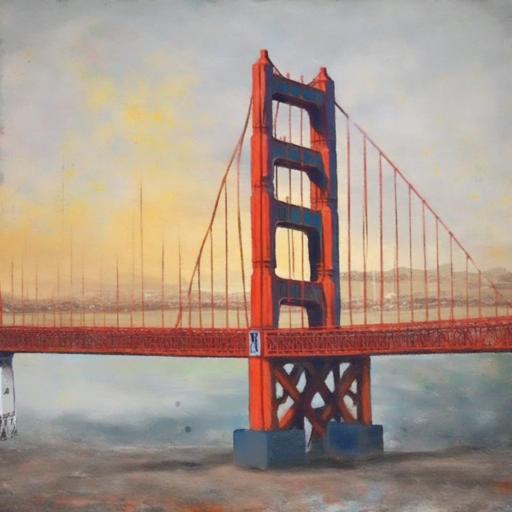} \\
    
    \includegraphics[width=0.137\textwidth]{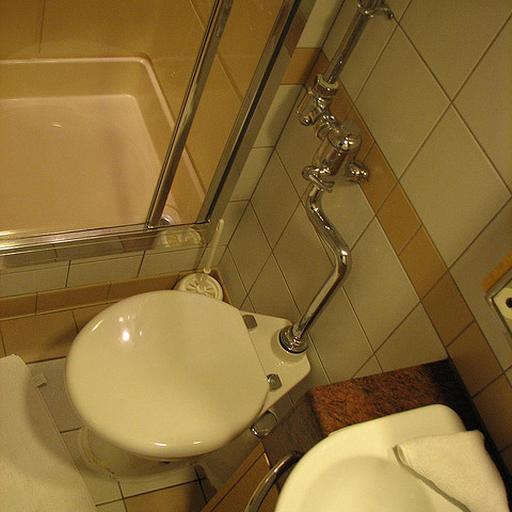} &
    \includegraphics[width=0.137\textwidth]{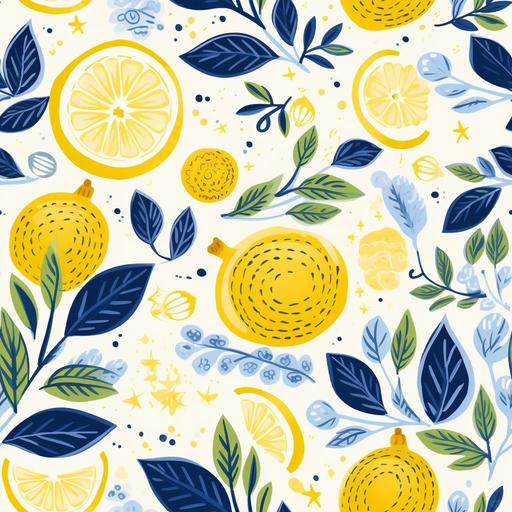} &
    \includegraphics[width=0.137\textwidth]{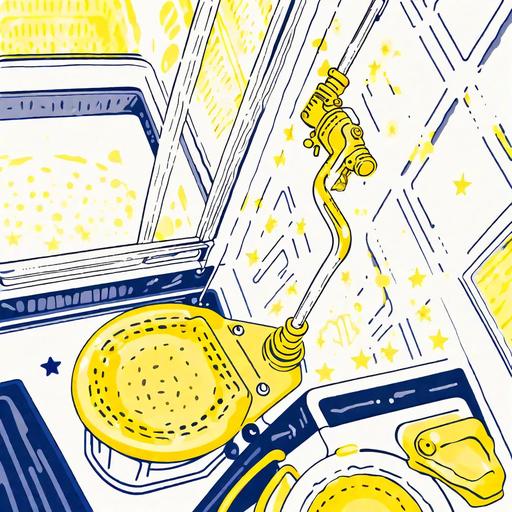} &
    \includegraphics[width=0.137\textwidth]{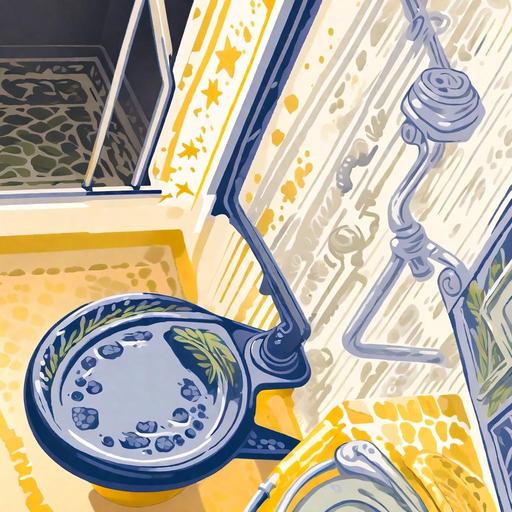} &
    \includegraphics[width=0.137\textwidth]{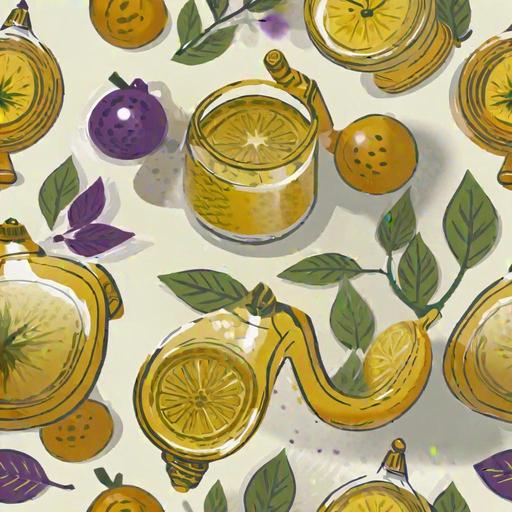} &
    \includegraphics[width=0.137\textwidth]{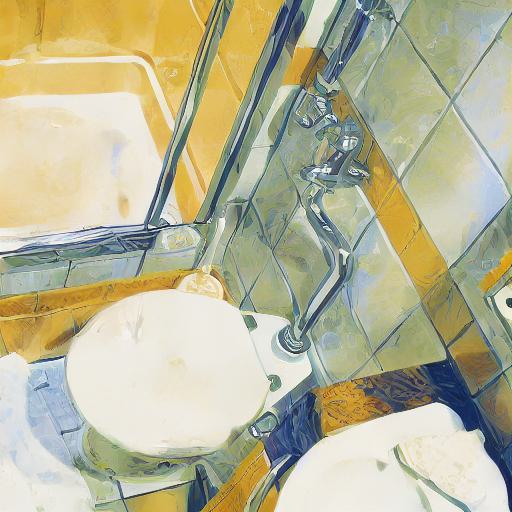} &
    \includegraphics[width=0.137\textwidth]{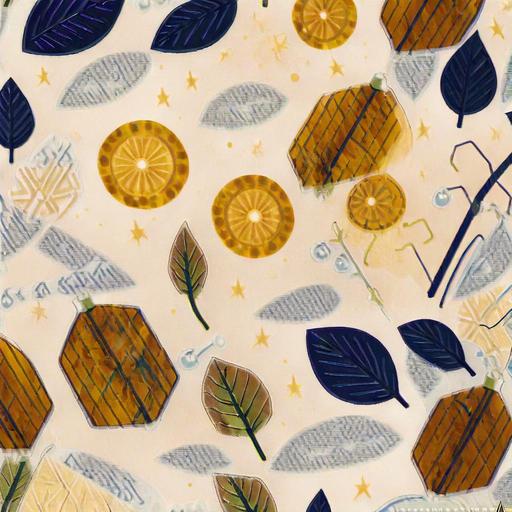} \\
        
    \end{tabular}
    }   
    \caption{\textbf{Additional qualitative comparison.} We present style transfer results of our method and four baseline methods, including B-LoRA~\cite{B-LoRA}, ZipLoRA~\cite{ziplora}, StyleID~\cite{styleID}, and StyleAligned~\cite{styleAligned}. Our method demonstrates superior performance in both content preservation and style alignment.}
    \label{fig:additional_qualitative_comparison}
\end{figure*}

\begin{figure*}[t]
    \centering
    \setlength{\tabcolsep}{0.85pt}
    \renewcommand{\arraystretch}{0.5}
    {\small
    \begin{tabular}{c@{\hspace{0.1cm}} | @{\hspace{0.1cm}}c c c c c c}
        
        \quad \ \ \ \ Content \ \ \ \raisebox{0.36in}{\rotatebox[origin=t]{90}{Style}}&
        \includegraphics[width=0.137\textwidth]{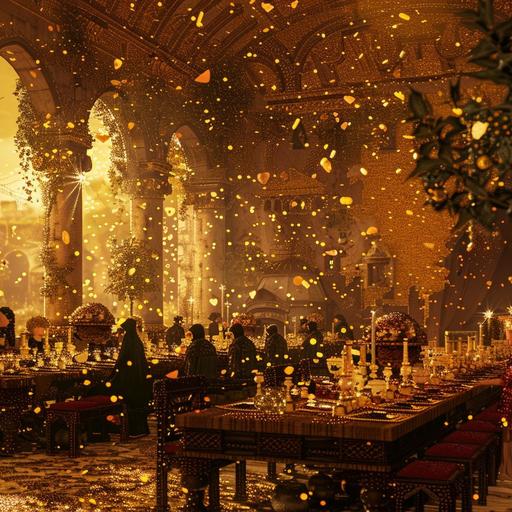} &
        \includegraphics[width=0.137\textwidth]{images/cnt-sty/style/rain_princess.jpg} &
        \includegraphics[width=0.137\textwidth]{images/cnt-sty/style/abstract_painting.jpg} &
        \includegraphics[width=0.137\textwidth]{images/cnt-sty/style/cartoon_line.jpg} &
        \includegraphics[width=0.137\textwidth]{images/cnt-sty/style/rabbit.jpg} &
        \includegraphics[width=0.137\textwidth]{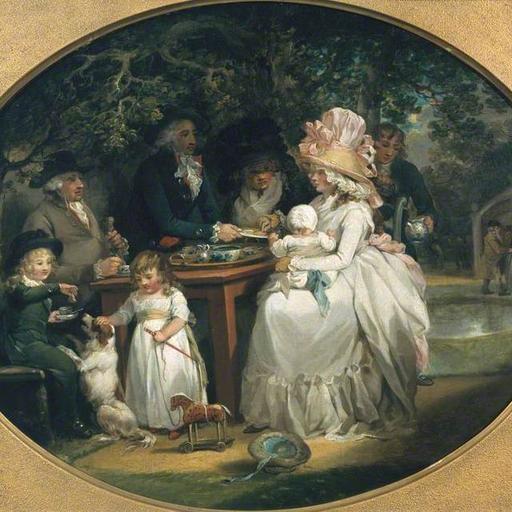}
        \\   
        
        \noalign{\vskip 0.07cm}\hline\noalign{\vskip 0.07cm}

        \includegraphics[width=0.137\textwidth]{images/cnt-sty/content/teddy_bear.jpg} &
        \includegraphics[width=0.137\textwidth]{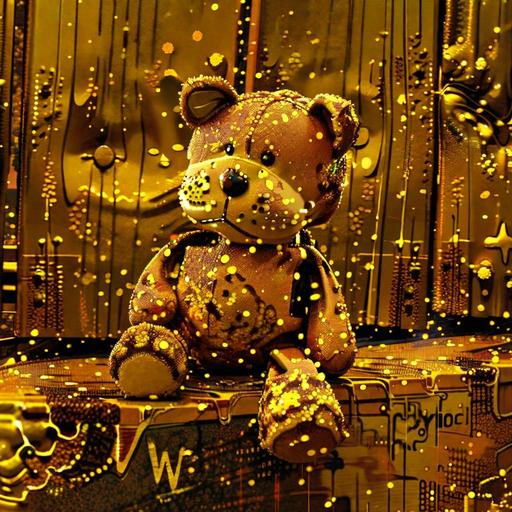} &
        \includegraphics[width=0.137\textwidth]{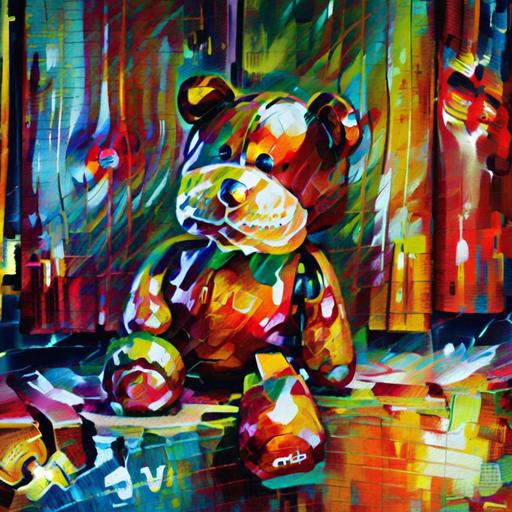} &
        \includegraphics[width=0.137\textwidth]{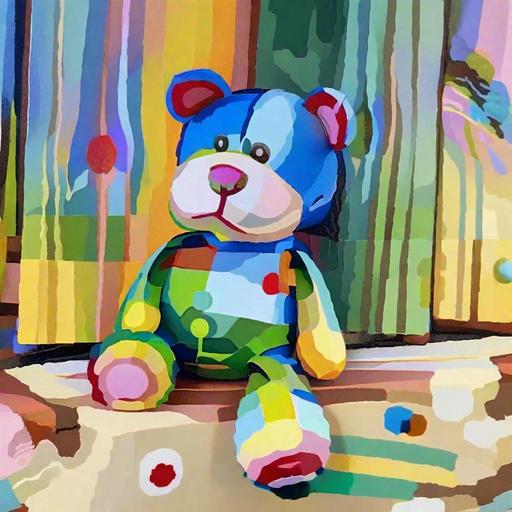} &
        \includegraphics[width=0.137\textwidth]{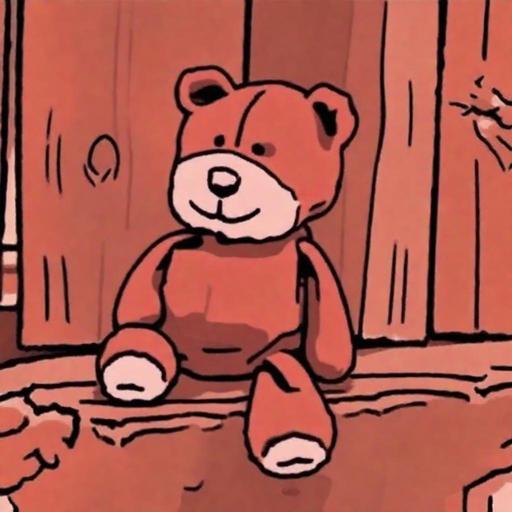} &
        \includegraphics[width=0.137\textwidth]{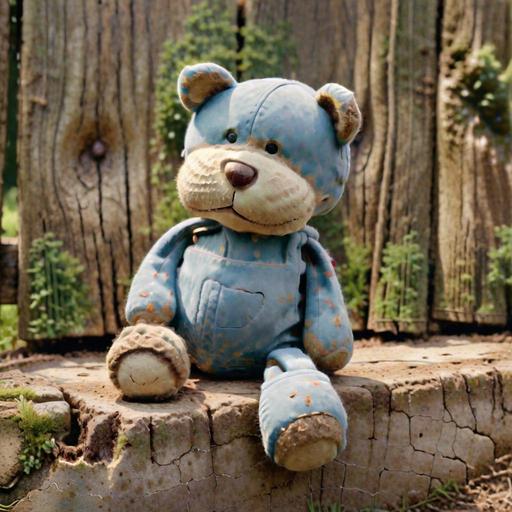} &
        \includegraphics[width=0.137\textwidth]{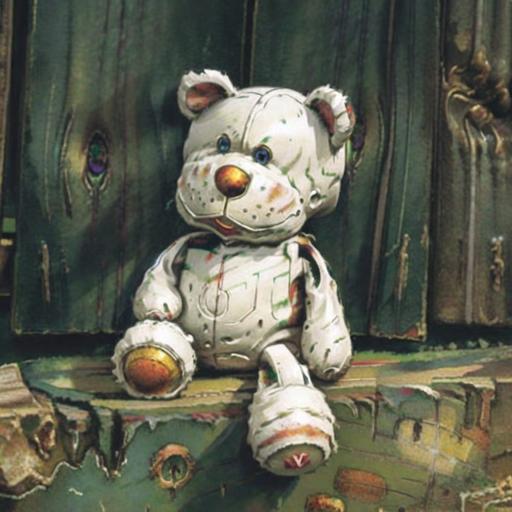}
        \\
        
        \includegraphics[width=0.137\textwidth]{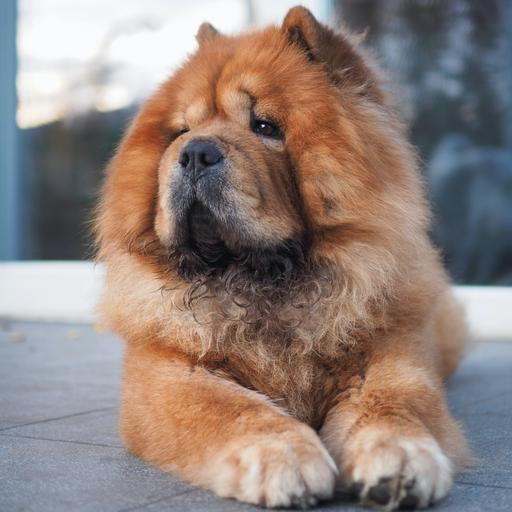} &
        \includegraphics[width=0.137\textwidth]{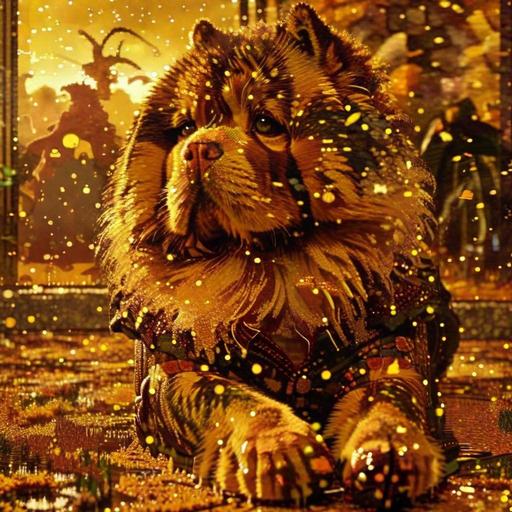} &
        \includegraphics[width=0.137\textwidth]{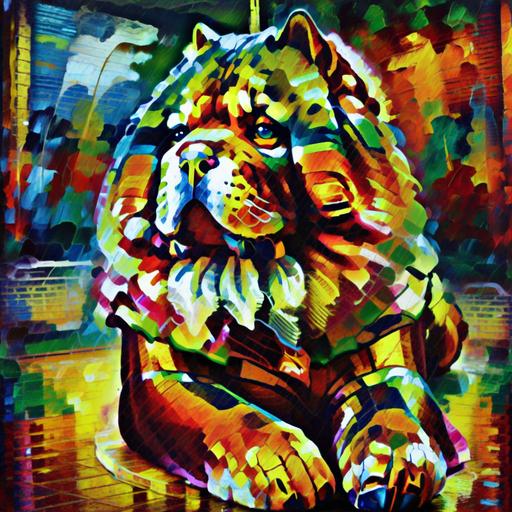} &
        \includegraphics[width=0.137\textwidth]{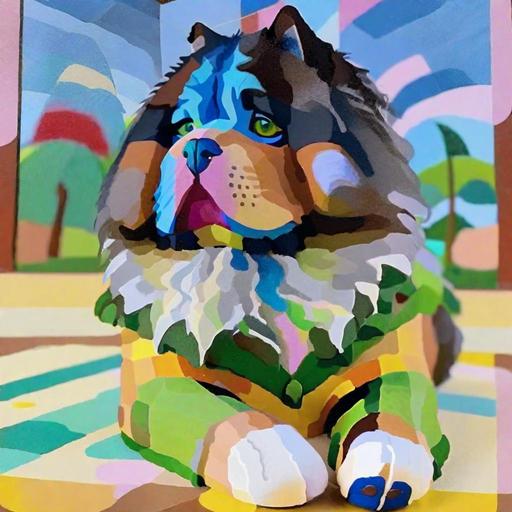} &
        \includegraphics[width=0.137\textwidth]{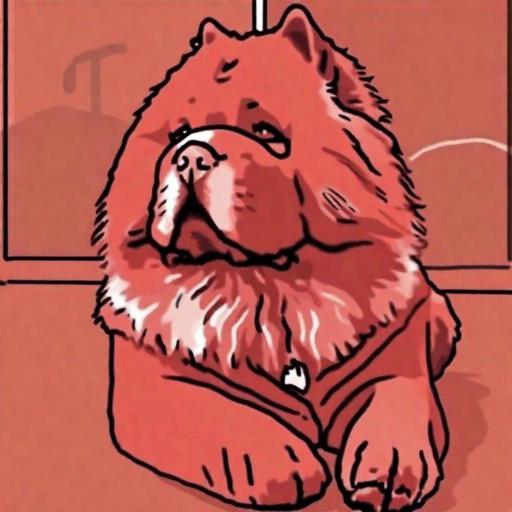} &
        \includegraphics[width=0.137\textwidth]{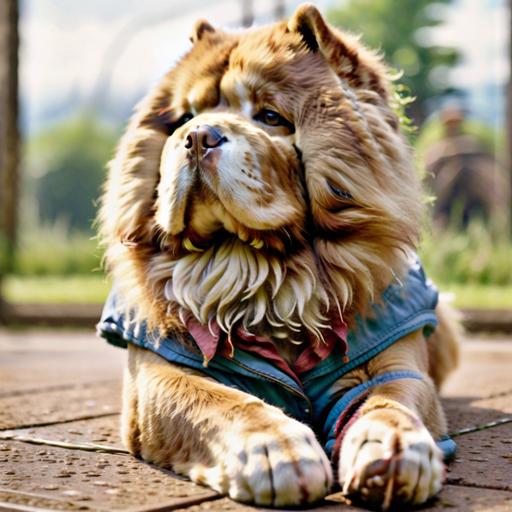} &
        \includegraphics[width=0.137\textwidth]{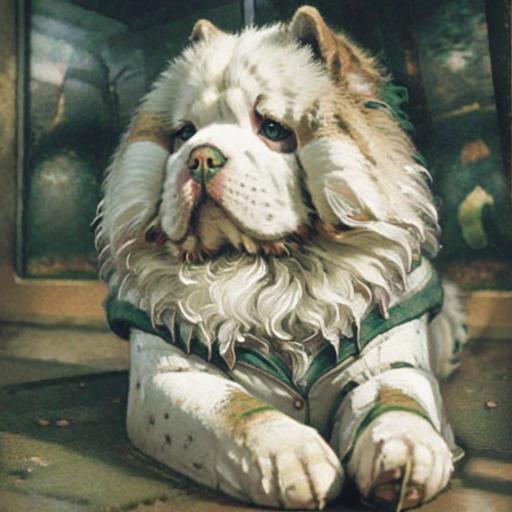}
        \\

        \includegraphics[width=0.137\textwidth]{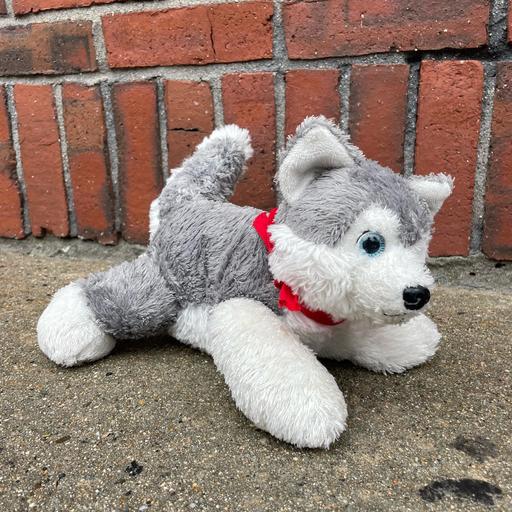} &
        \includegraphics[width=0.137\textwidth]{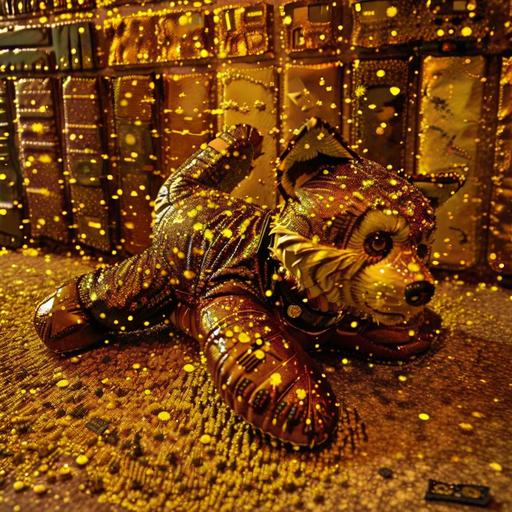} &
        \includegraphics[width=0.137\textwidth]{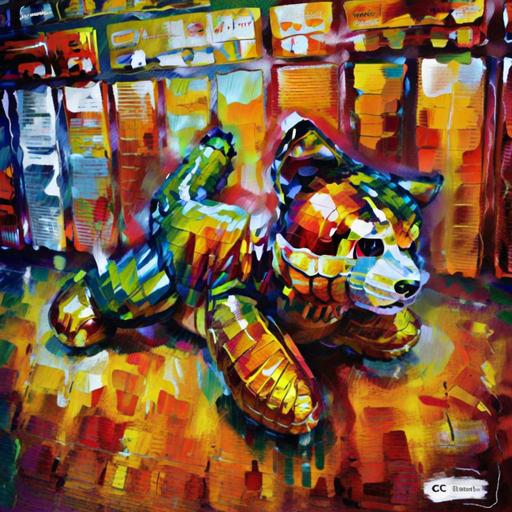} &
        \includegraphics[width=0.137\textwidth]{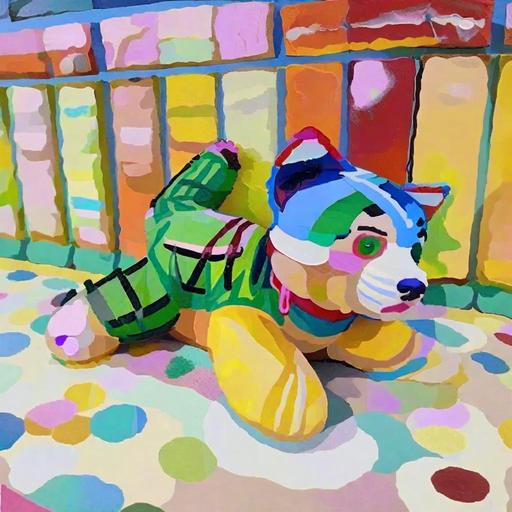} &
        \includegraphics[width=0.137\textwidth]{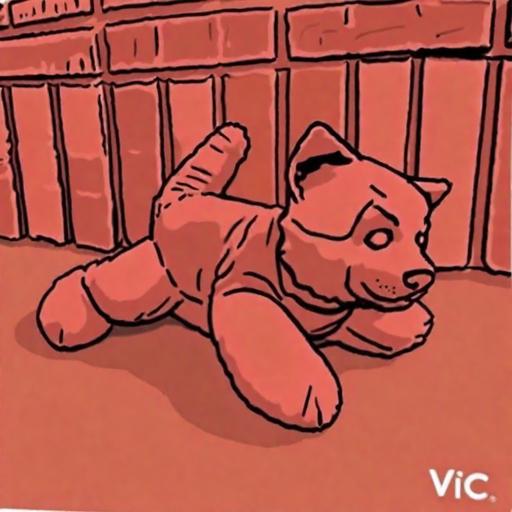} &
        \includegraphics[width=0.137\textwidth]{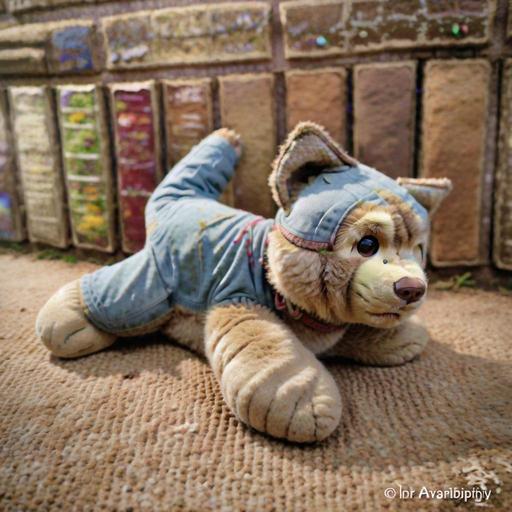} &
        \includegraphics[width=0.137\textwidth]{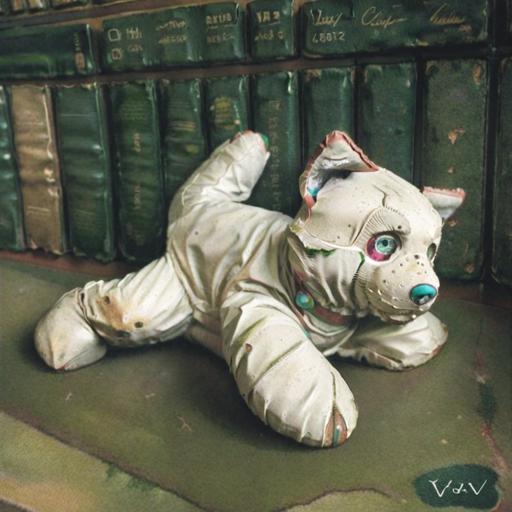} 
        \\
        
        \includegraphics[width=0.137\textwidth]{images/cnt-sty/content/sofa.jpg} &
        \includegraphics[width=0.137\textwidth]{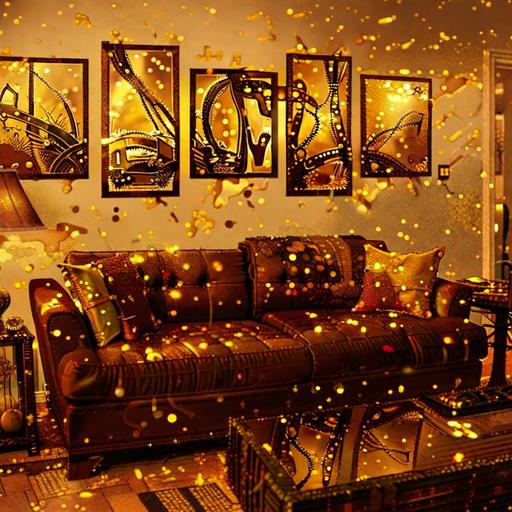} &
        \includegraphics[width=0.137\textwidth]{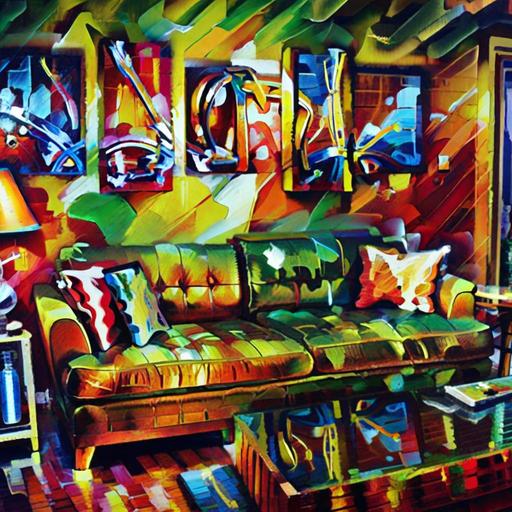} &
        \includegraphics[width=0.137\textwidth]{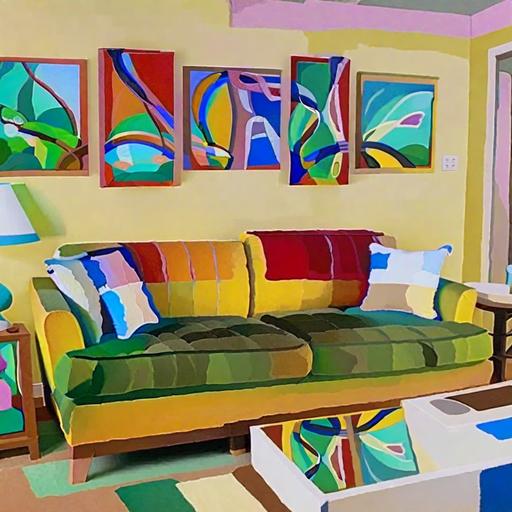} &
        \includegraphics[width=0.137\textwidth]{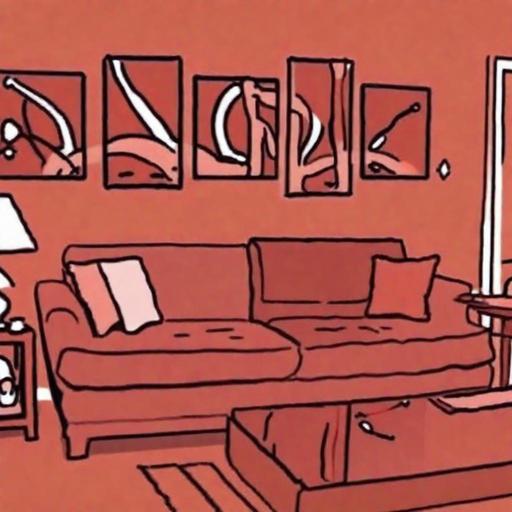} &
        \includegraphics[width=0.137\textwidth]{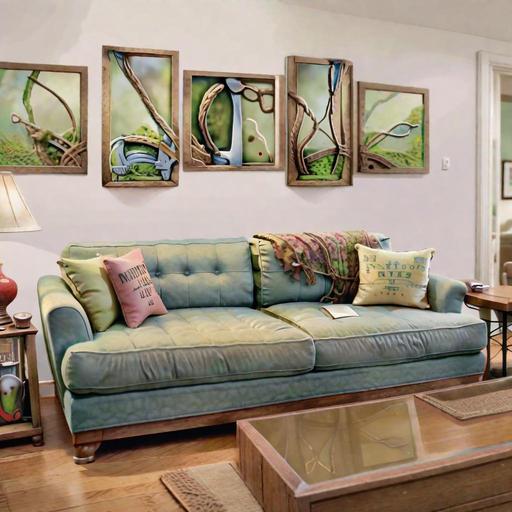} &
        \includegraphics[width=0.137\textwidth]{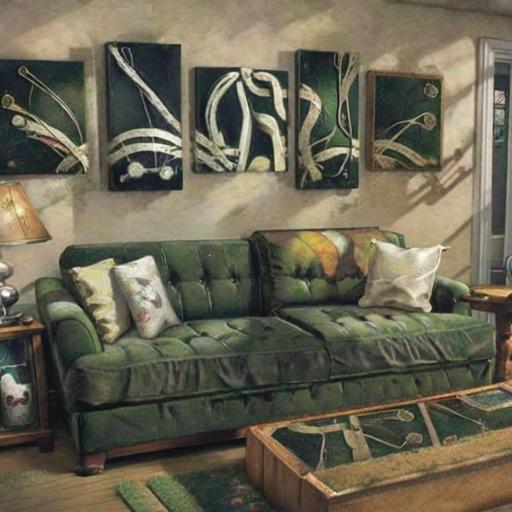} 
        \\

        \includegraphics[width=0.137\textwidth]{images/cnt-sty/content/toilet.jpg} &
        \includegraphics[width=0.137\textwidth]{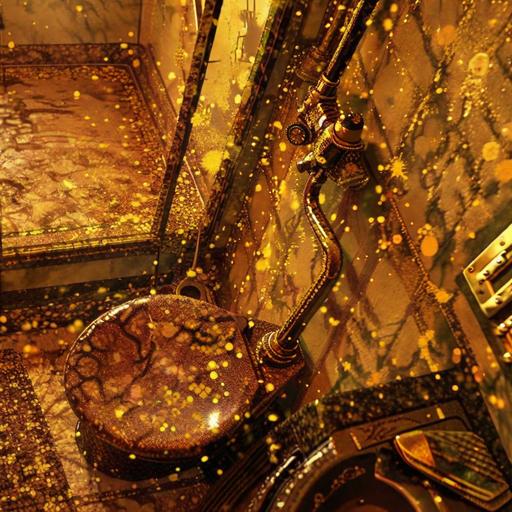} &
        \includegraphics[width=0.137\textwidth]{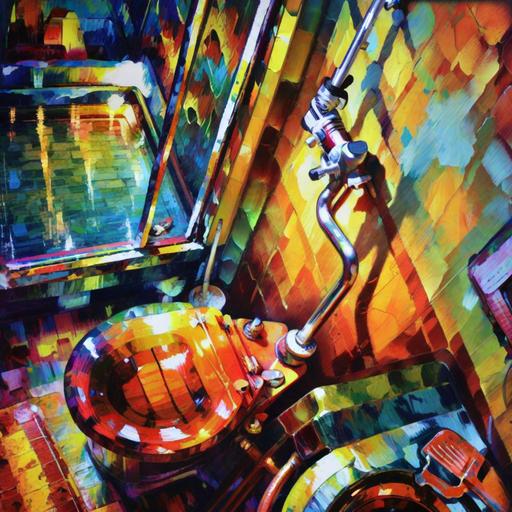} &
        \includegraphics[width=0.137\textwidth]{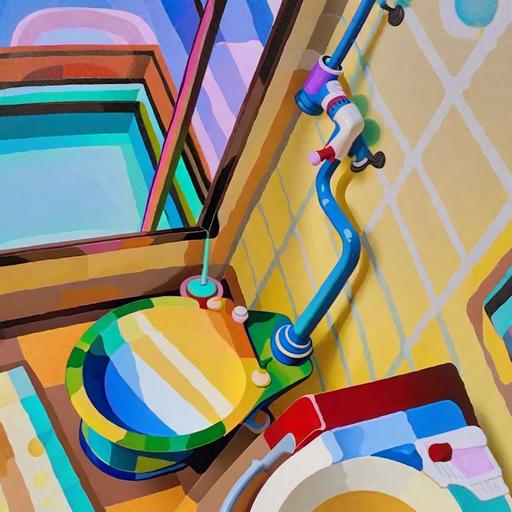} &
        \includegraphics[width=0.137\textwidth]{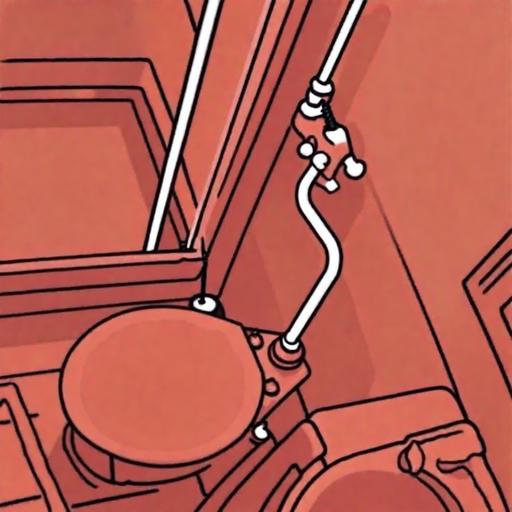} &
        \includegraphics[width=0.137\textwidth]{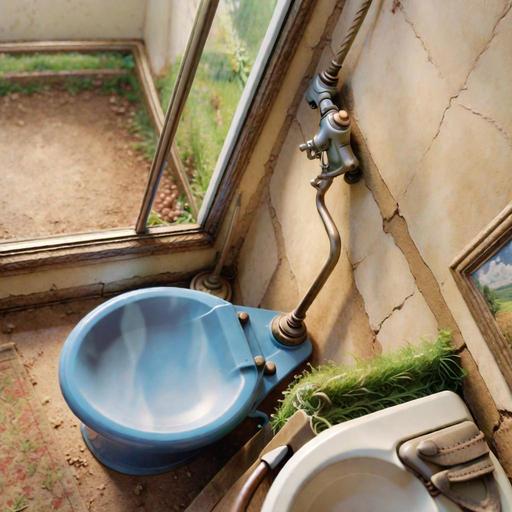} &
        \includegraphics[width=0.137\textwidth]{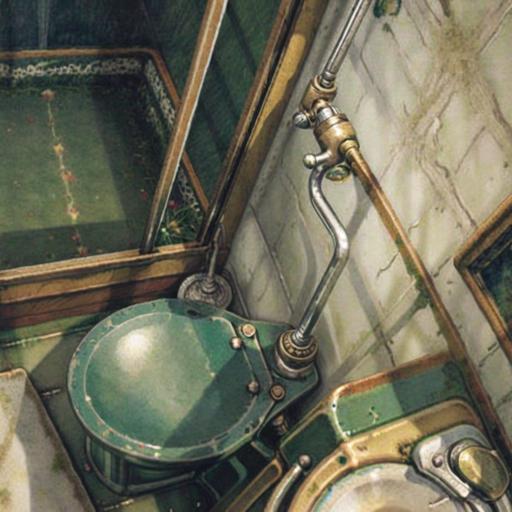} 
        \\

        \includegraphics[width=0.137\textwidth]{images/cnt-sty/content/sailboat.jpg} &
        \includegraphics[width=0.137\textwidth]{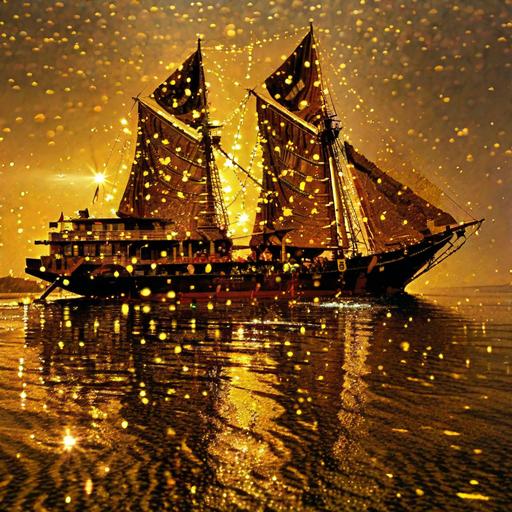} &
        \includegraphics[width=0.137\textwidth]{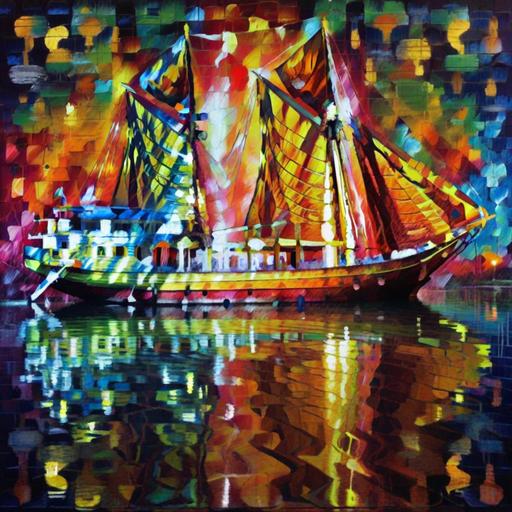} &
        \includegraphics[width=0.137\textwidth]{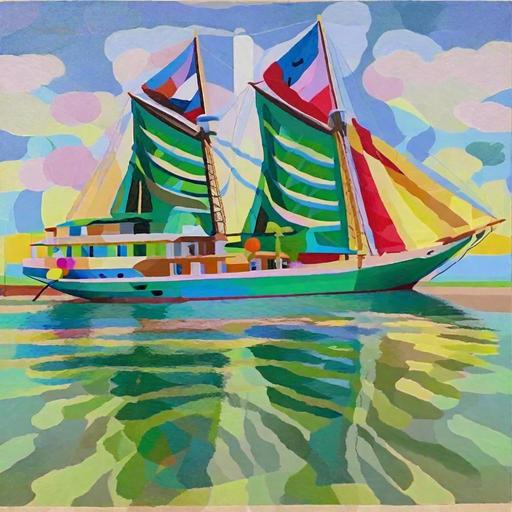} &
        \includegraphics[width=0.137\textwidth]{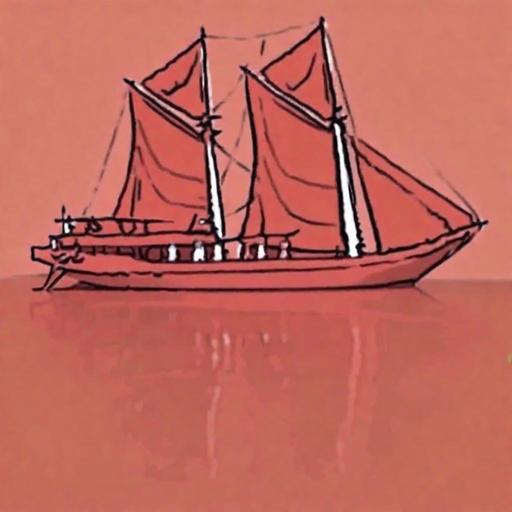} &
        \includegraphics[width=0.137\textwidth]{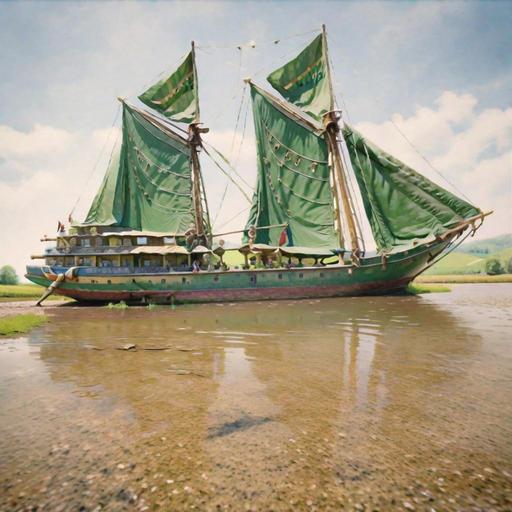} &
        \includegraphics[width=0.137\textwidth]{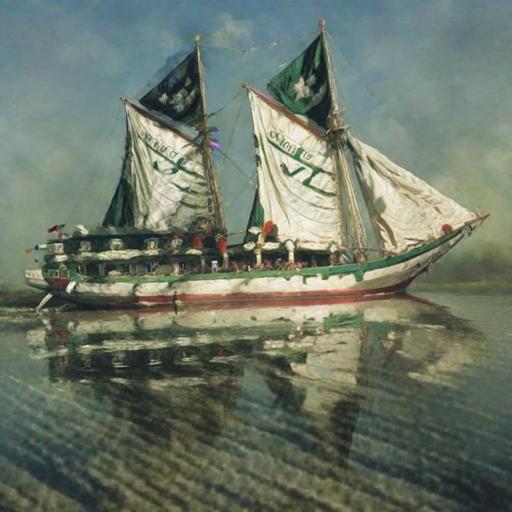} 
        \\

    \end{tabular}
    }   
    
    \caption{Additional style transfer results by ConsisLoRA.}
    \label{fig:additional_qualitative_results_2}
\end{figure*}

\begin{figure*}[t]
    \centering
    \setlength{\tabcolsep}{0.85pt}
    \renewcommand{\arraystretch}{0.5}
    {\small
    \begin{tabular}{c@{\hspace{0.1cm}} | @{\hspace{0.1cm}}c c c c c c}
        
        \quad \ \ \ \ Content \ \ \ \raisebox{0.36in}{\rotatebox[origin=t]{90}{Style}}&
        \includegraphics[width=0.137\textwidth]{images/cnt-sty/style/pig.jpg} &
        \includegraphics[width=0.137\textwidth]{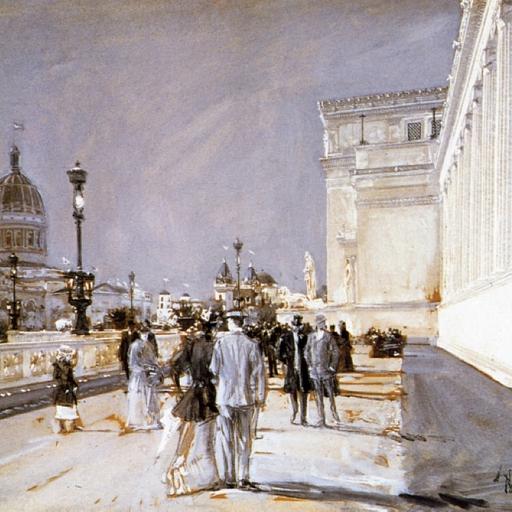} &
        \includegraphics[width=0.137\textwidth]{images/cnt-sty/style/orange.jpg} &   
        \includegraphics[width=0.137\textwidth]{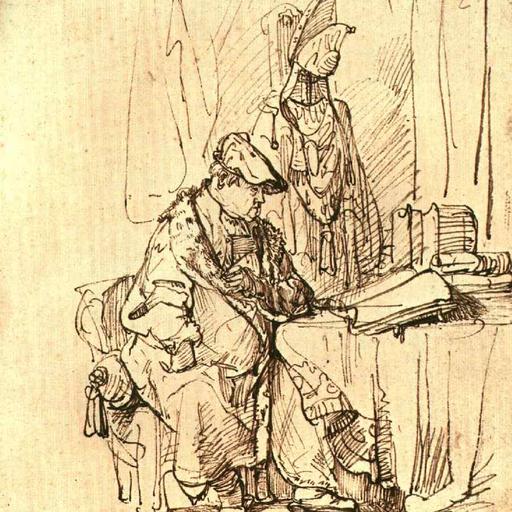} &
        \includegraphics[width=0.137\textwidth]{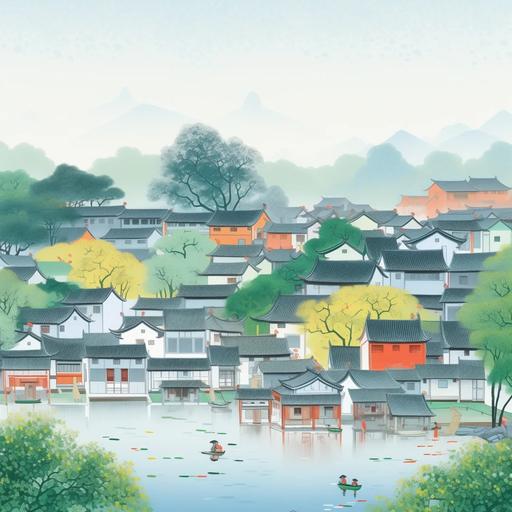} &
        \includegraphics[width=0.137\textwidth]{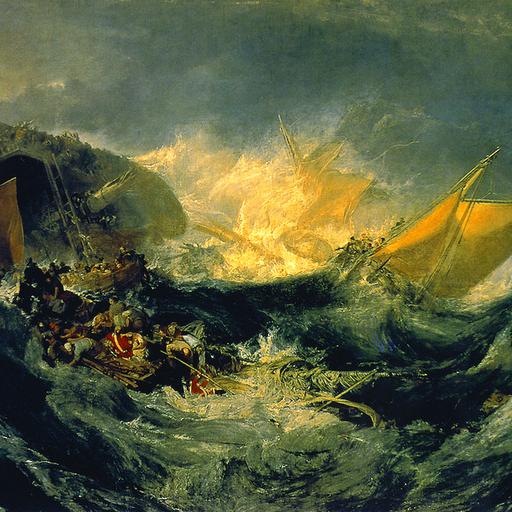} \\
        
        \noalign{\vskip 0.07cm}\hline\noalign{\vskip 0.07cm}

        \includegraphics[width=0.137\textwidth]{images/cnt-sty/content/modern.jpg} &
        \includegraphics[width=0.137\textwidth]{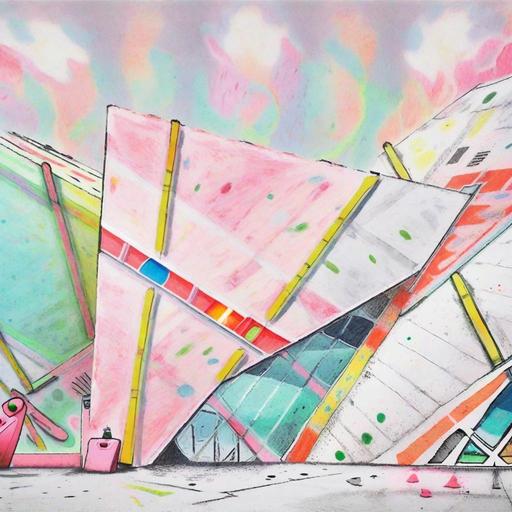} &
        \includegraphics[width=0.137\textwidth]{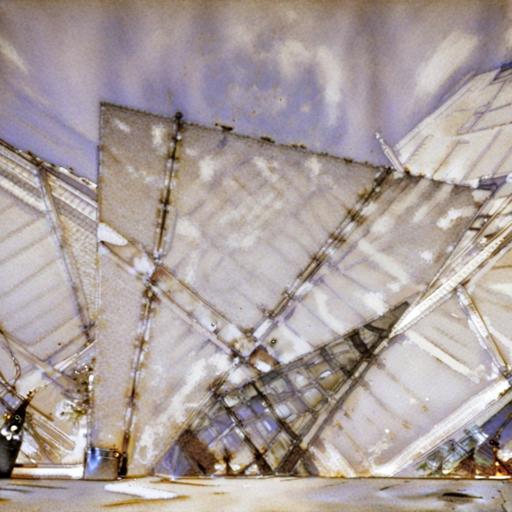} &
        \includegraphics[width=0.137\textwidth]{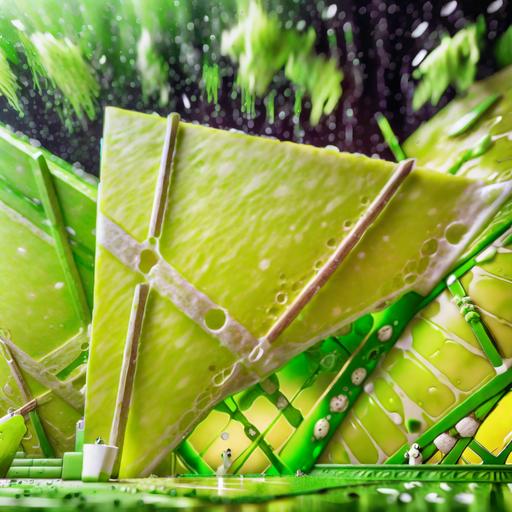} &
        \includegraphics[width=0.137\textwidth]{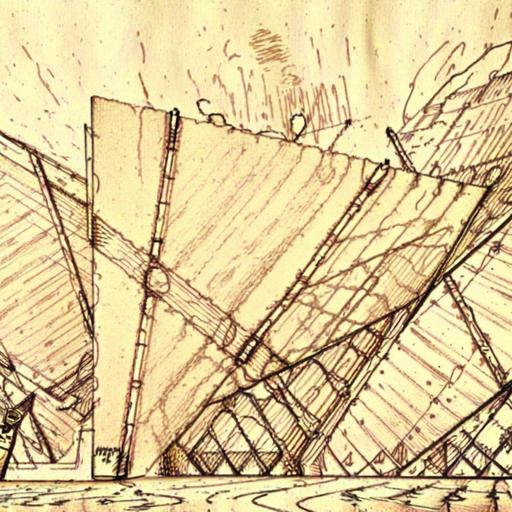} &
        \includegraphics[width=0.137\textwidth]{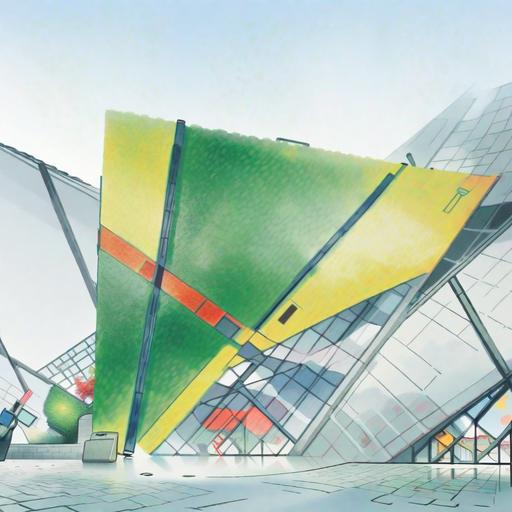} &
        \includegraphics[width=0.137\textwidth]{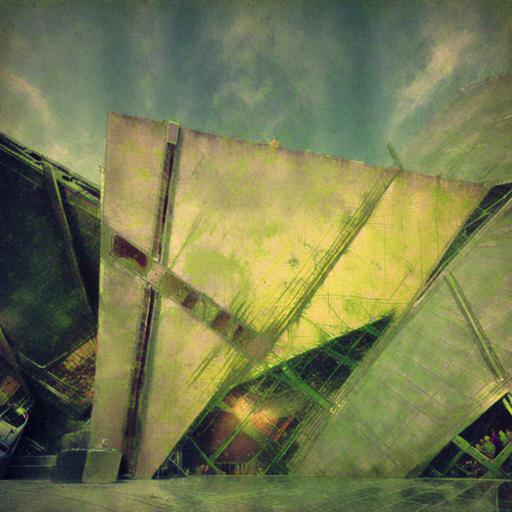} \\

        \includegraphics[width=0.137\textwidth]{images/cnt-sty/content/building.jpg} &
        \includegraphics[width=0.137\textwidth]{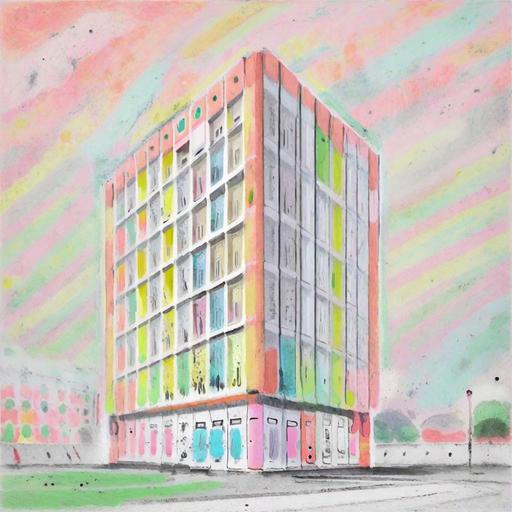} &
        \includegraphics[width=0.137\textwidth]{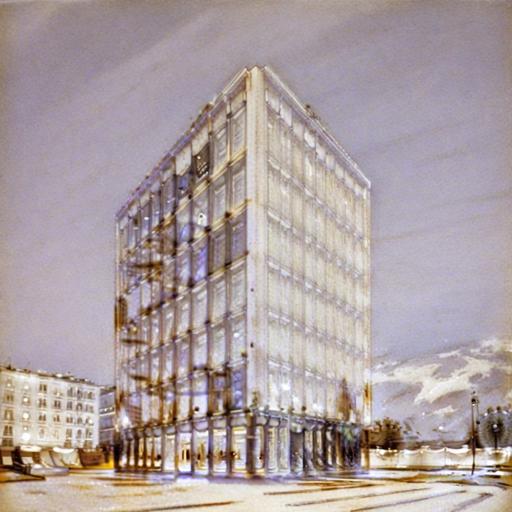} &
        \includegraphics[width=0.137\textwidth]{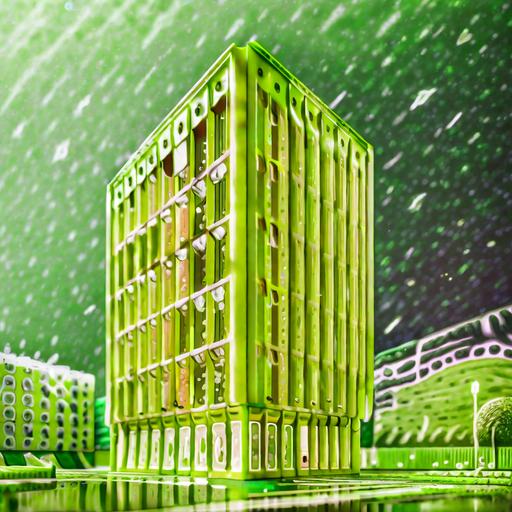} &
        \includegraphics[width=0.137\textwidth]{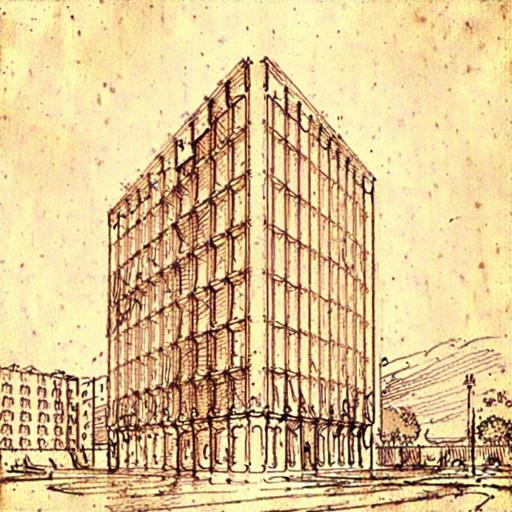} &
        \includegraphics[width=0.137\textwidth]{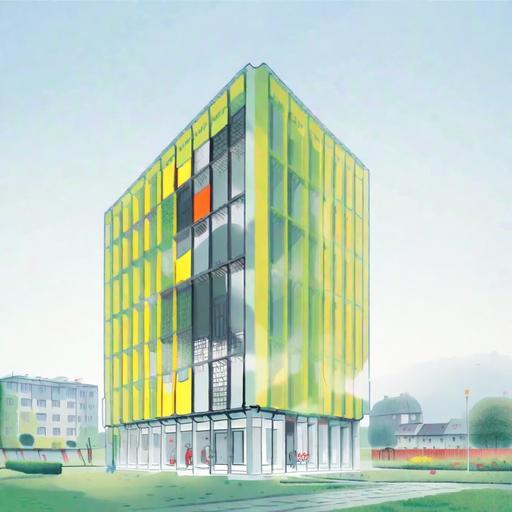} &
        \includegraphics[width=0.137\textwidth]{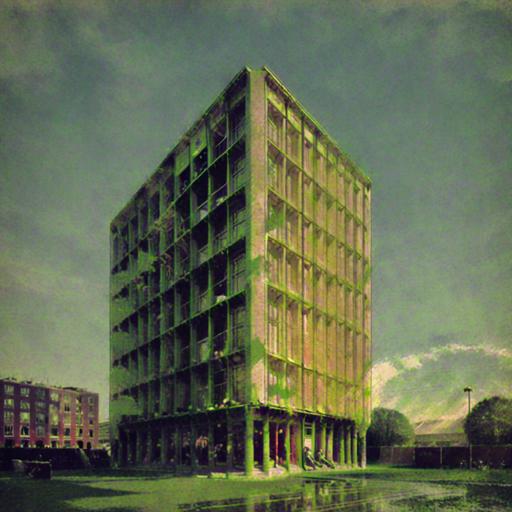} \\
        
        \includegraphics[width=0.137\textwidth]{images/cnt-sty/content/cornell.jpg} &
        \includegraphics[width=0.137\textwidth]{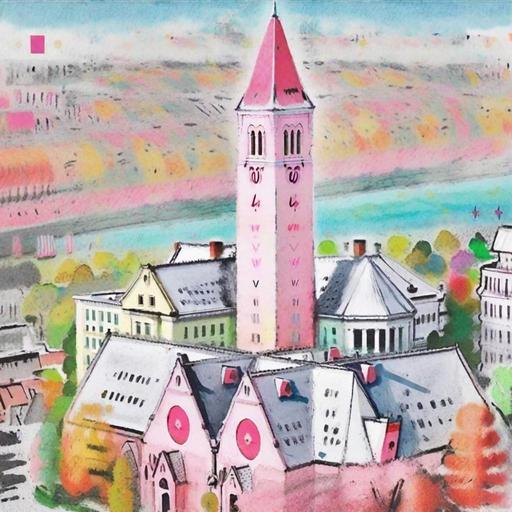} &
        \includegraphics[width=0.137\textwidth]{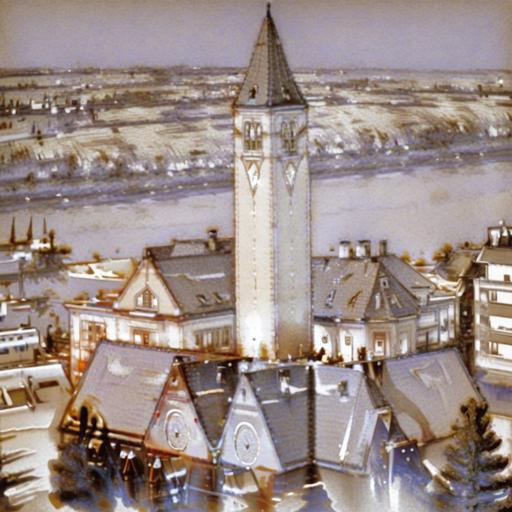} &
        \includegraphics[width=0.137\textwidth]{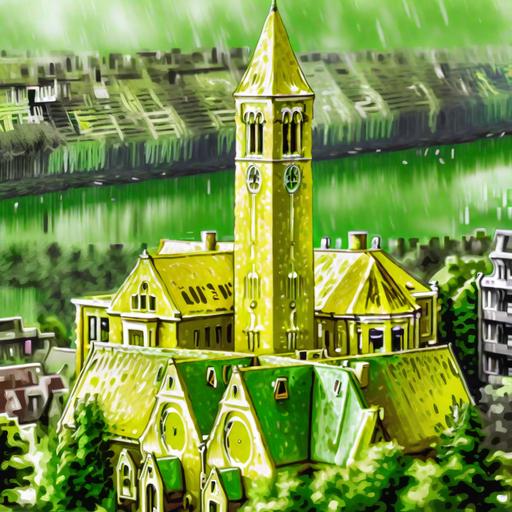} &
        \includegraphics[width=0.137\textwidth]{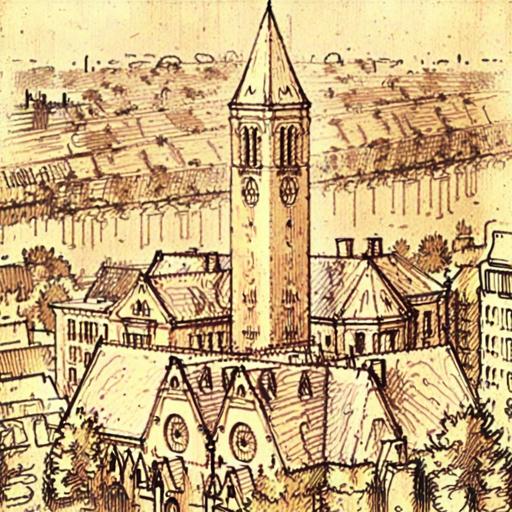} &
        \includegraphics[width=0.137\textwidth]{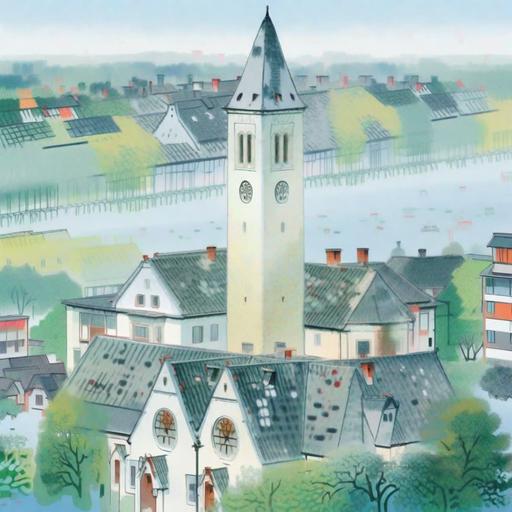} &
        \includegraphics[width=0.137\textwidth]{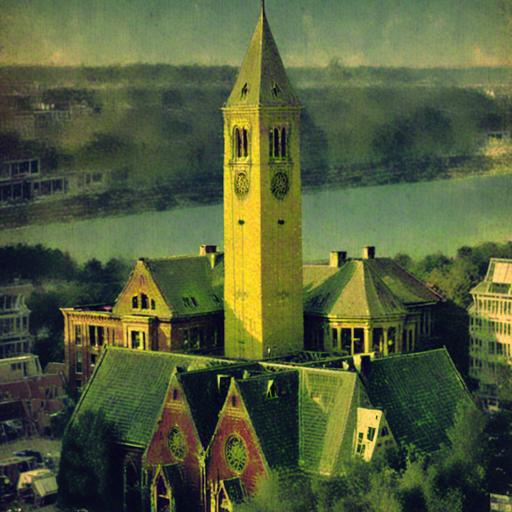} \\

        \includegraphics[width=0.137\textwidth]{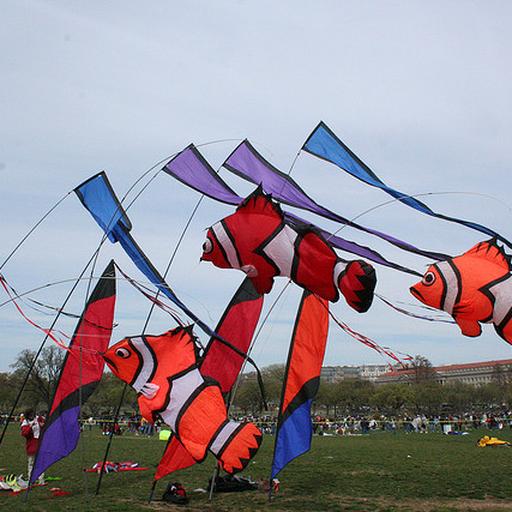} &
        \includegraphics[width=0.137\textwidth]{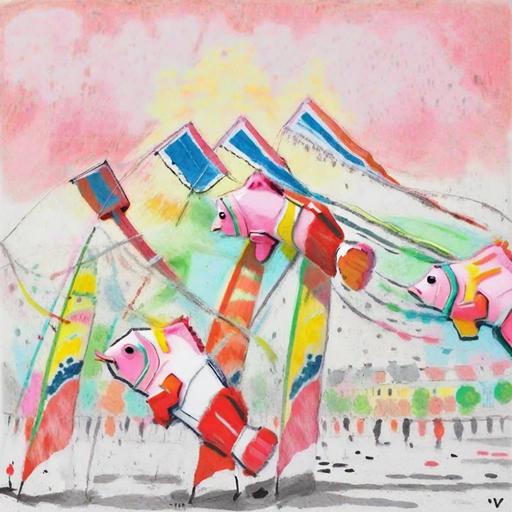} &
        \includegraphics[width=0.137\textwidth]{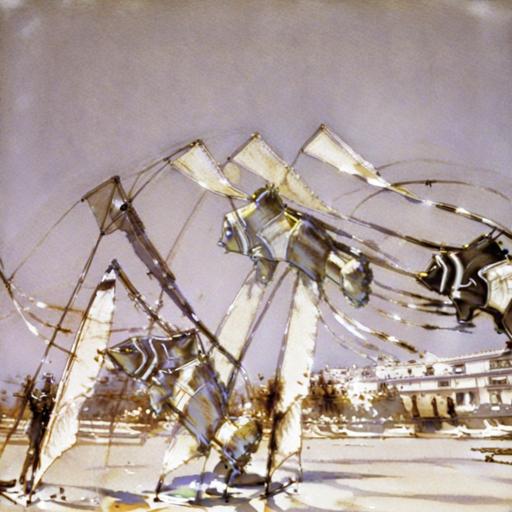} &
        \includegraphics[width=0.137\textwidth]{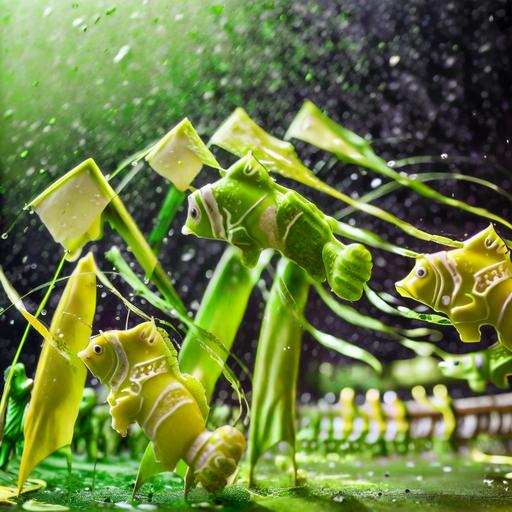} &
        \includegraphics[width=0.137\textwidth]{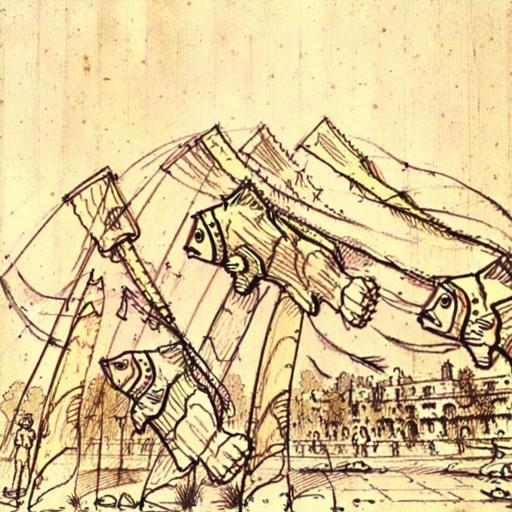} &
        \includegraphics[width=0.137\textwidth]{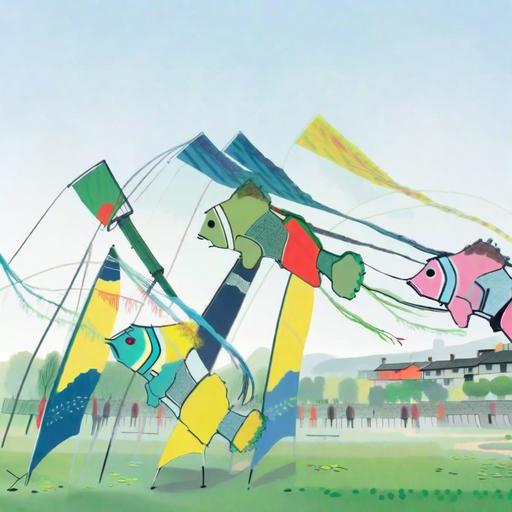} &
        \includegraphics[width=0.137\textwidth]{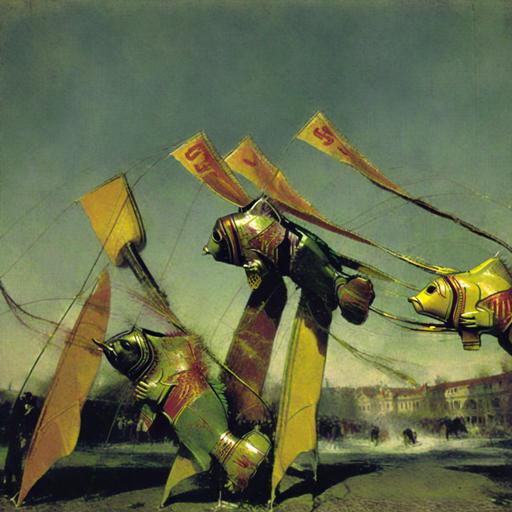} \\
        
        \includegraphics[width=0.137\textwidth]{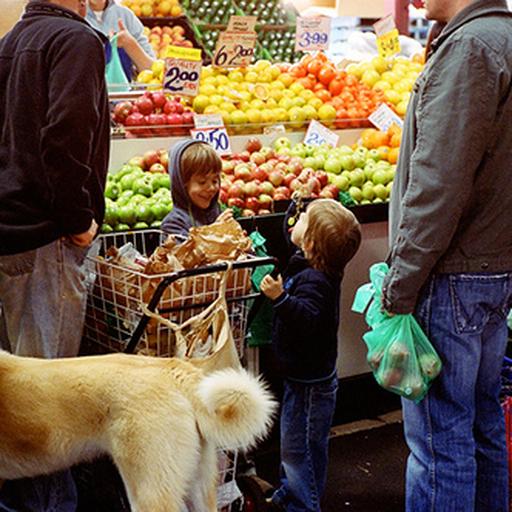} &
        \includegraphics[width=0.137\textwidth]{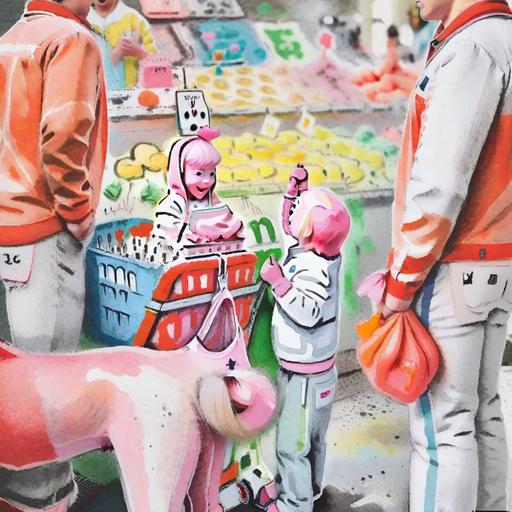} &
        \includegraphics[width=0.137\textwidth]{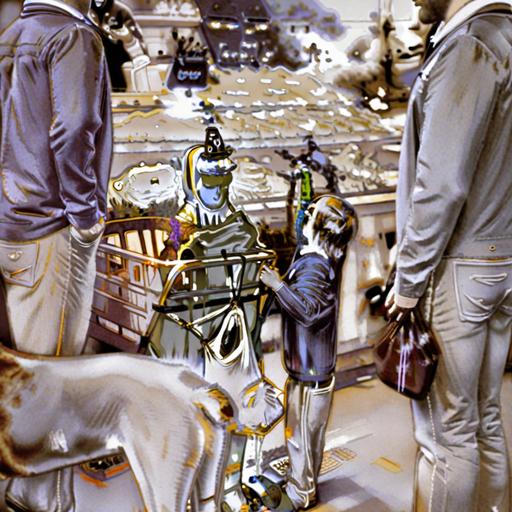} &
        \includegraphics[width=0.137\textwidth]{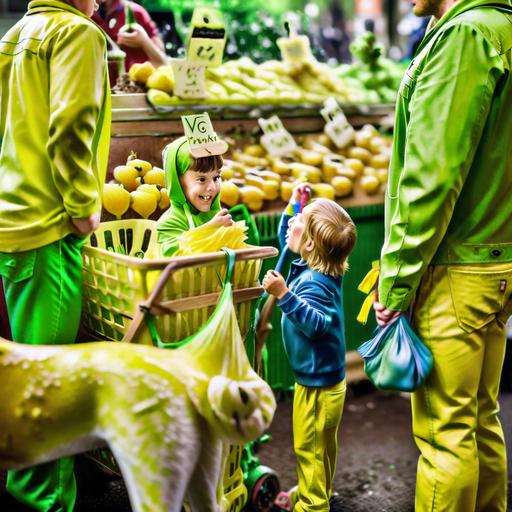} &
        \includegraphics[width=0.137\textwidth]{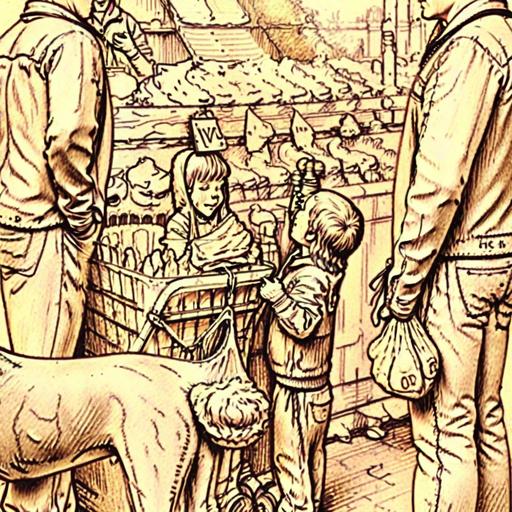} &
        \includegraphics[width=0.137\textwidth]{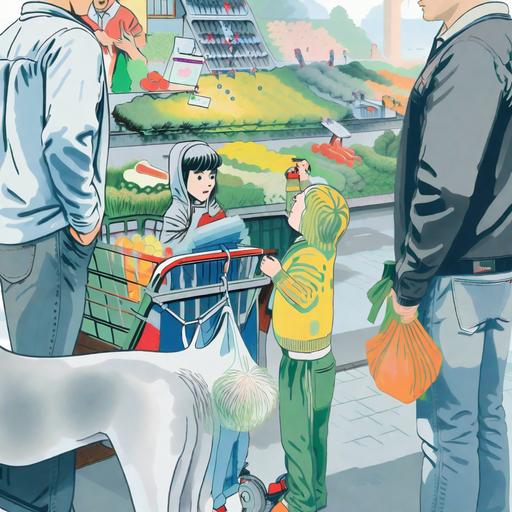} &
        \includegraphics[width=0.137\textwidth]{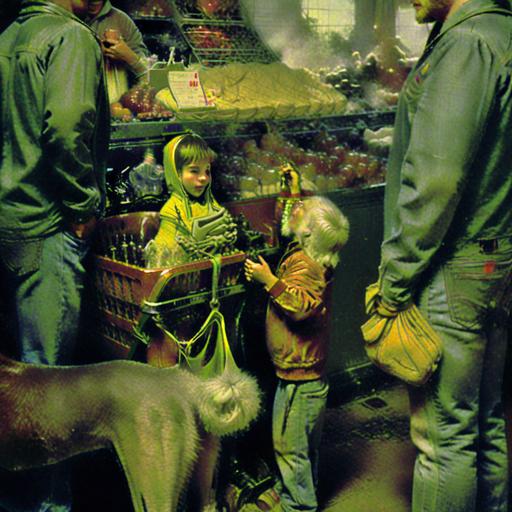} \\
        
        \noalign{\vskip 0.04cm}\hline\noalign{\vskip 0.1cm}
    
        \multicolumn{1}{c}{Content input} & \multicolumn{1}{c}{``Pixel art''}  & ``Ice'' & ``Vintage''   &  ``Sketch cartoon''& ``Cyberpunk'' &  ``Gothic dark'' \\
      
        \includegraphics[width=0.137\textwidth]{images/cnt-sty/content/dog.jpg} &
        \includegraphics[width=0.137\textwidth]{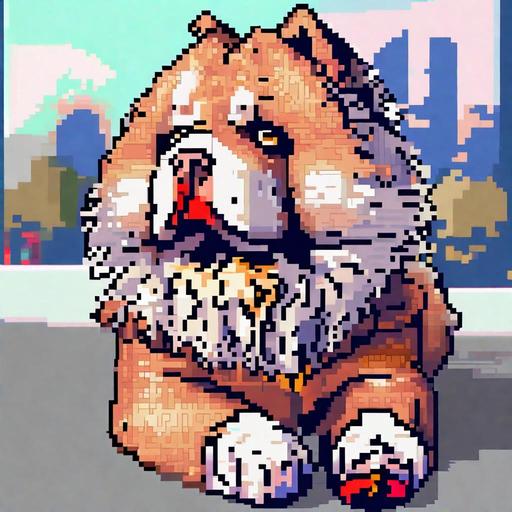} &
        \includegraphics[width=0.137\textwidth]{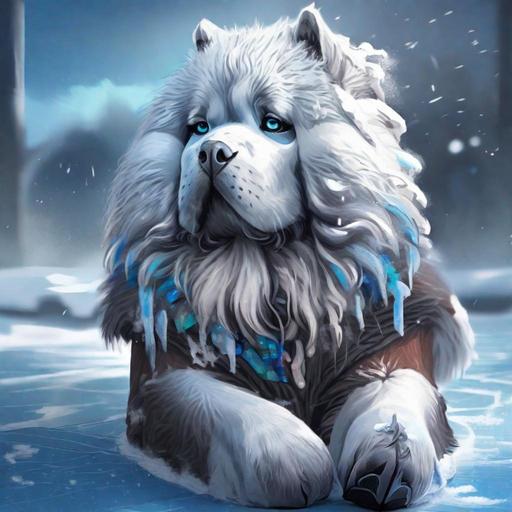} &
        \includegraphics[width=0.137\textwidth]{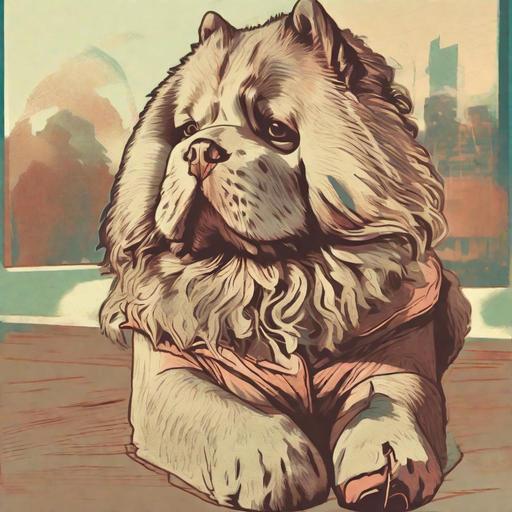} &
        \includegraphics[width=0.137\textwidth]{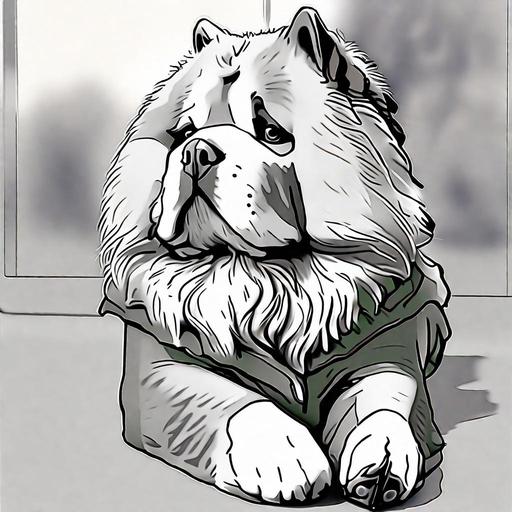} &
        \includegraphics[width=0.137\textwidth]{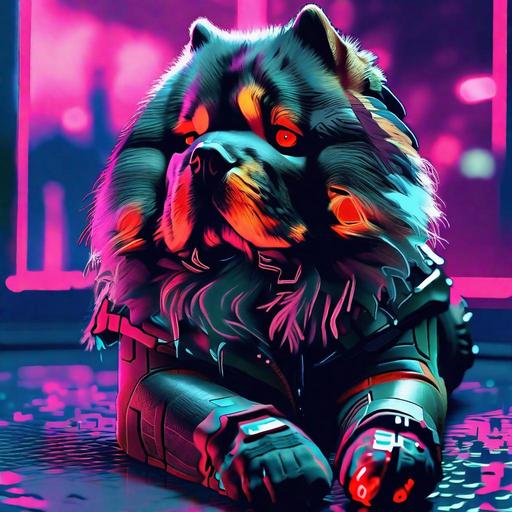} &
        \includegraphics[width=0.137\textwidth]{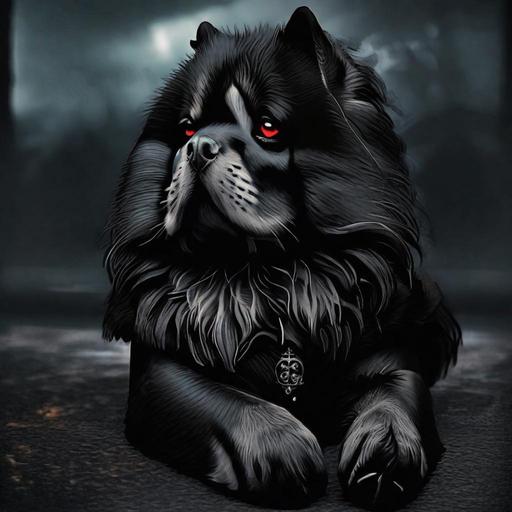}
        \\ 

        \noalign{\vskip 0.04cm}\hline\noalign{\vskip 0.1cm}

        \multicolumn{1}{c}{Style input} & \multicolumn{1}{c}{``Dog''} & ``Car'' & ``Train'' & ``Bench'' & ``Classroom'' & ``Bedroom'' \\
        
        \includegraphics[width=0.137\textwidth]{images/cnt-sty/style/palace.jpg} &
        \includegraphics[width=0.137\textwidth]{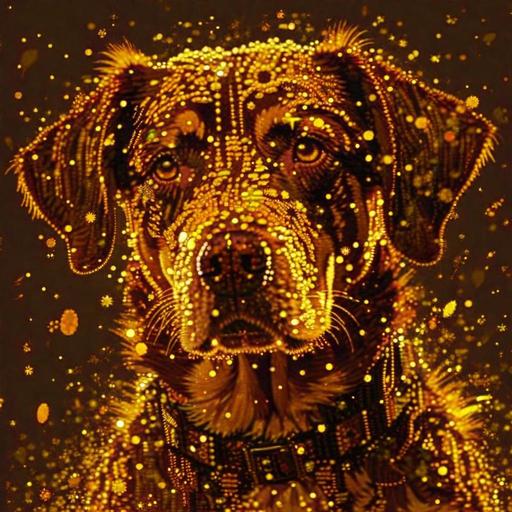} &
        \includegraphics[width=0.137\textwidth]{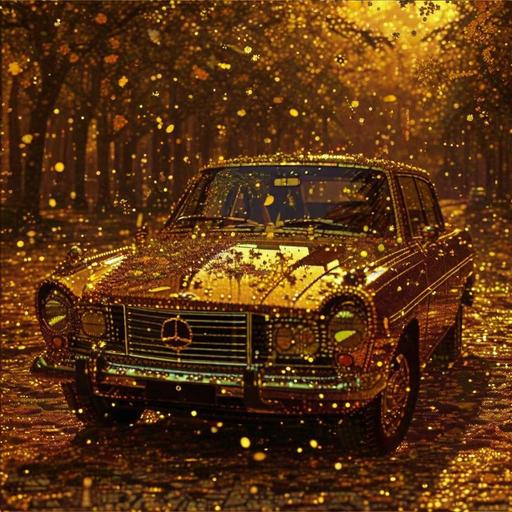} &
        \includegraphics[width=0.137\textwidth]{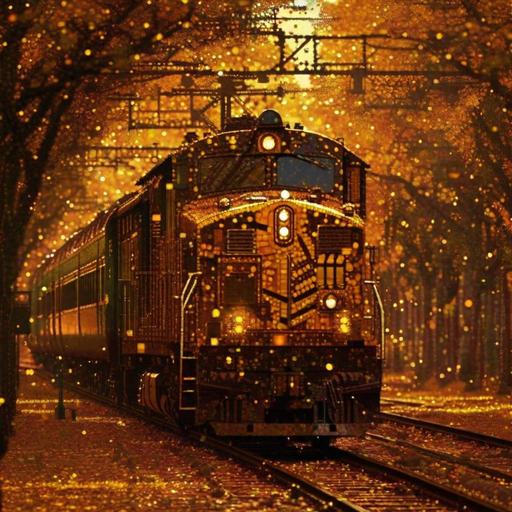} &
        \includegraphics[width=0.137\textwidth]{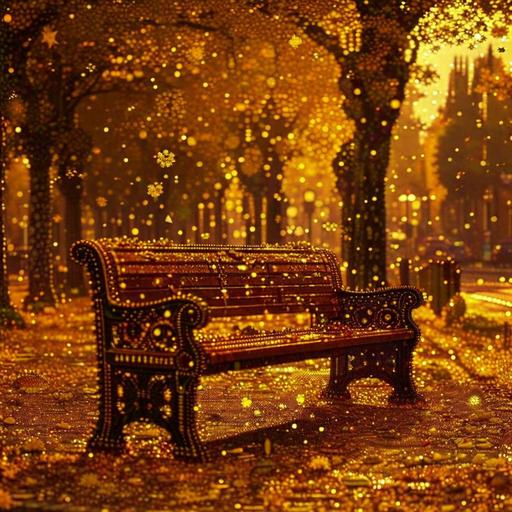} &
        \includegraphics[width=0.137\textwidth]{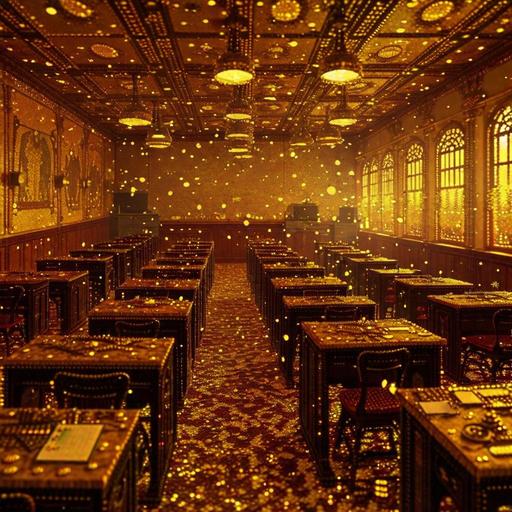} &
        \includegraphics[width=0.137\textwidth]{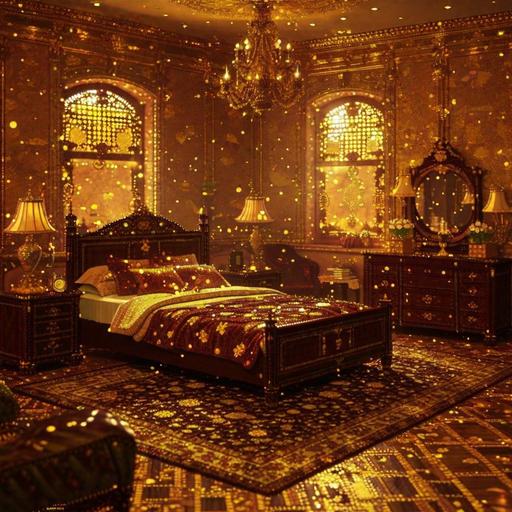} 
        \\    
    \end{tabular}
    }   
    
    \caption{Additional qualitative results by ConsisLoRA for three image stylization tasks, including style transfer (top), text-based image stylization (middle), and consistent style generation (bottom).}
    \label{fig:additional_qualitative_results_1}
\end{figure*}

In \cref{fig:additional_qualitative_comparison}, we present additional qualitative comparisons against the baseline methods using various pairs of style and content images. Our method demonstrates superior performance in both content preservation and style alignment compared to the baselines. Moreover, in \cref{fig:additional_qualitative_results_1,fig:additional_qualitative_results_2}, we provide additional qualitative results generated by ConsisLoRA for different image stylization applications.

\section{Content and Style Decomposition}\label{sec:appendix_decomposition}
\begin{figure*}[t]
 \centering
 \includegraphics[width=1\linewidth]{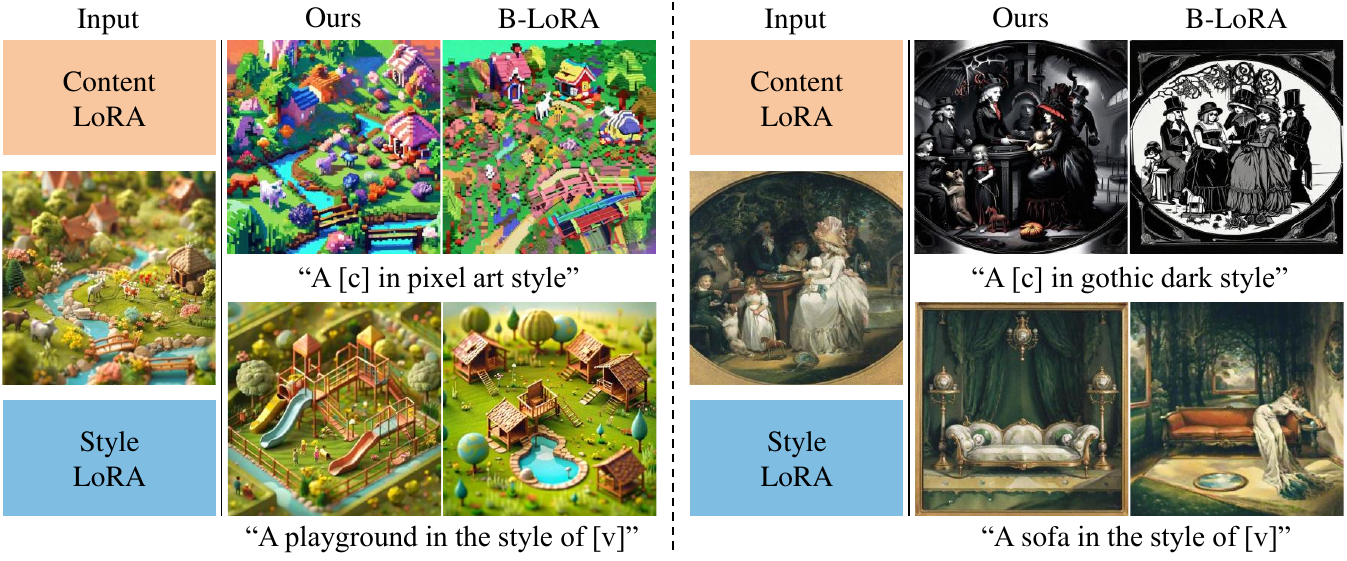}
\caption{Additional results of content and style decomposition.}
\label{fig:additional_decomposition}
\end{figure*}

In \cref{fig:additional_decomposition}, we present an additional decomposition comparison between our method and B-LoRA. As illustrated, our method shows clear advantages in accurately capturing both the content structure and style features.

\section{Additional Ablation Study} \label{sec:appendix_ablation}
\begin{figure*}[t]
    \centering
    \setlength{\tabcolsep}{0.85pt}
    \renewcommand{\arraystretch}{0.5}
    {\small
    \begin{tabular}{c c@{\hspace{0.1cm}} | @{\hspace{0.1cm}}c c c c}

         Content Image
         & \multicolumn{1}{c@{}}{Style Image}
         & \multicolumn{1}{c}{w/o $x_0$-prediction}
         & w/o two-step training
         & w/o loss transition
         & Full \\

        \includegraphics[width=0.16\textwidth]{images/cnt-sty/content/eiffel_tower.jpg} &
        \includegraphics[width=0.16\textwidth]{images/cnt-sty/style/eyes.jpg} &
        \includegraphics[width=0.16\textwidth]{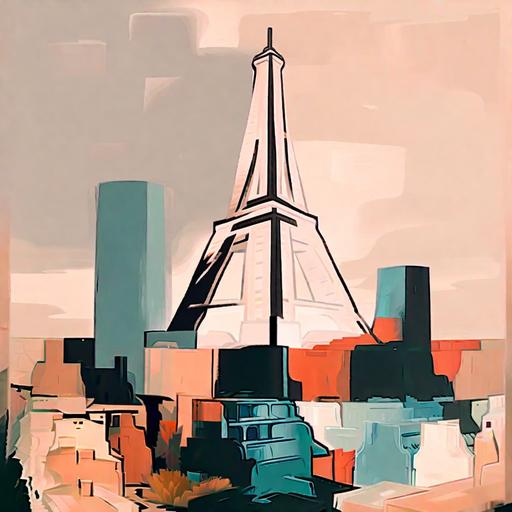} &
        \includegraphics[width=0.16\textwidth]{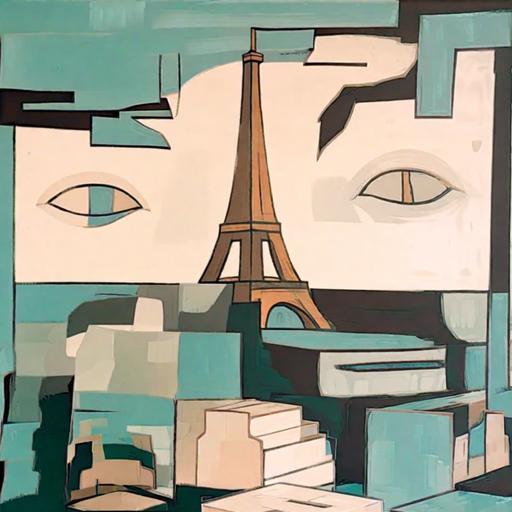} &
        \includegraphics[width=0.16\textwidth]{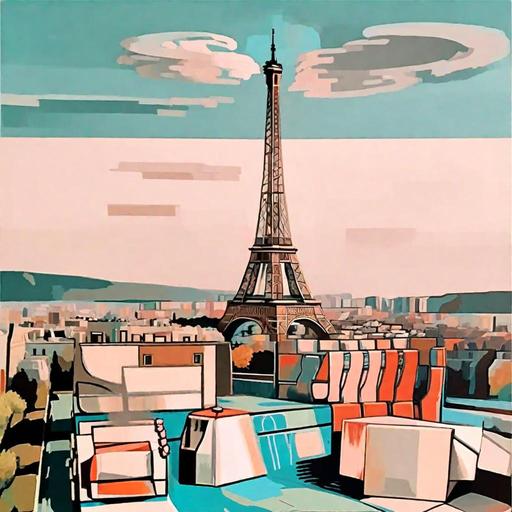} &
        \includegraphics[width=0.16\textwidth]{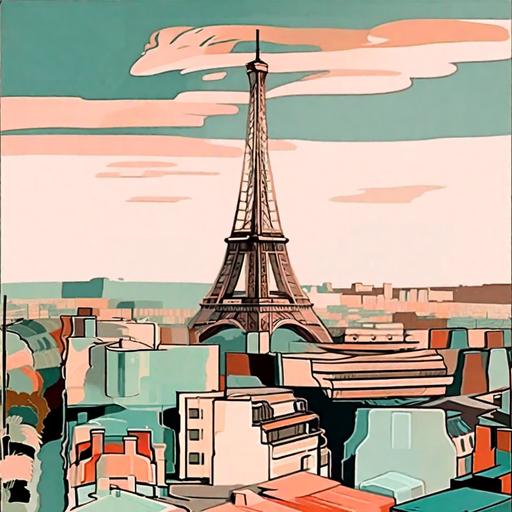} 
        \\

        \includegraphics[width=0.16\textwidth]{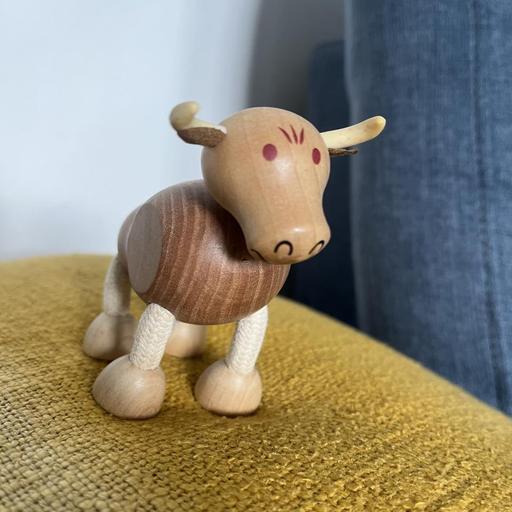} &
        \includegraphics[width=0.16\textwidth]{images/cnt-sty/style/pig.jpg} &
        \includegraphics[width=0.16\textwidth]{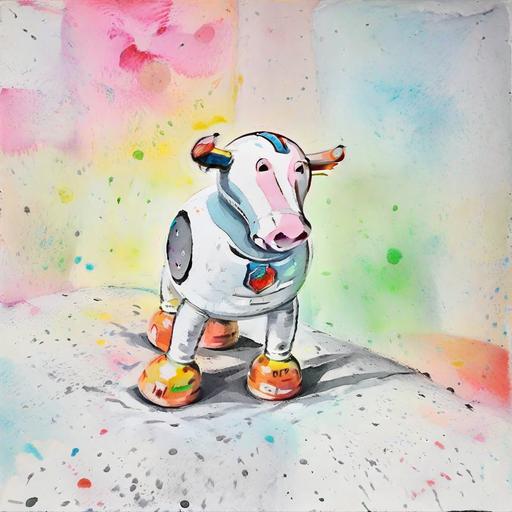} &
        \includegraphics[width=0.16\textwidth]{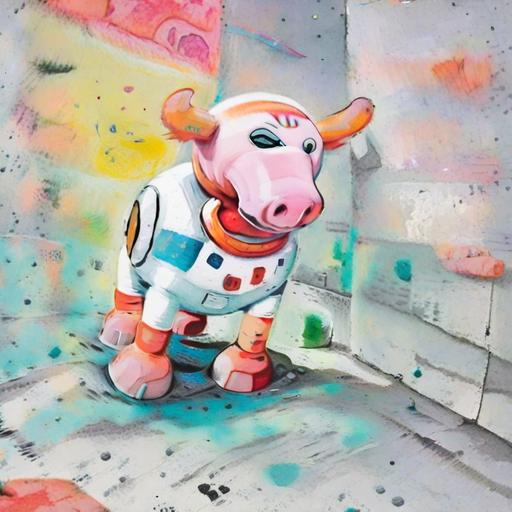} &
        \includegraphics[width=0.16\textwidth]{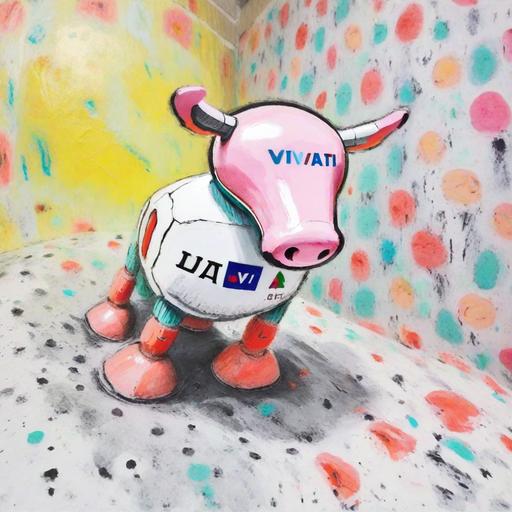} &
        \includegraphics[width=0.16\textwidth]{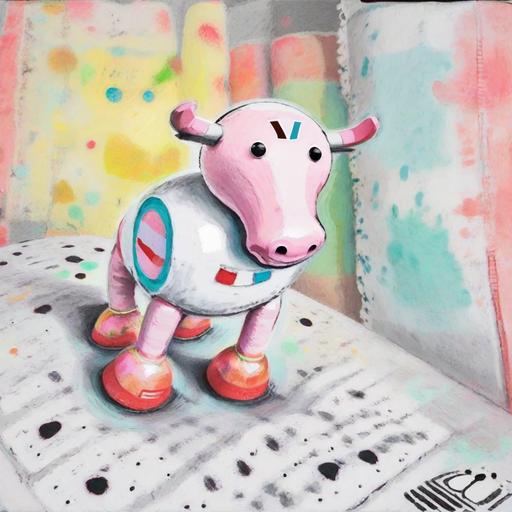} 
        \\ 

        \includegraphics[width=0.16\textwidth]{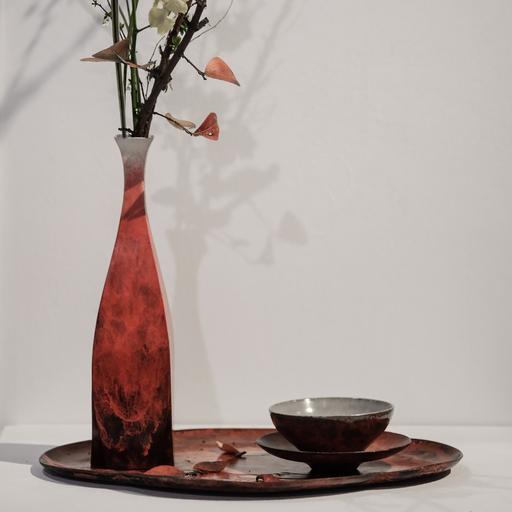} &
        \includegraphics[width=0.16\textwidth]{images/cnt-sty/style/pirate.jpg} &
        \includegraphics[width=0.16\textwidth]{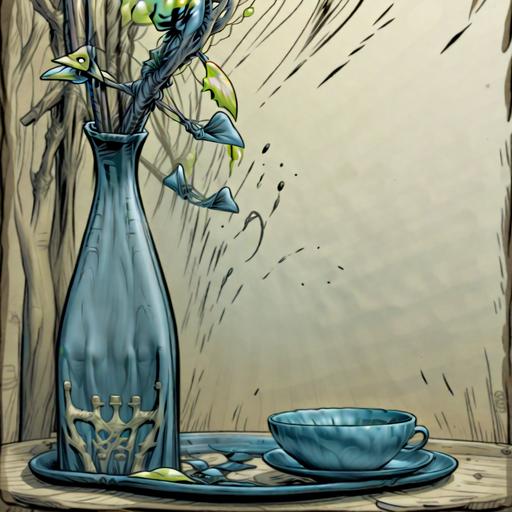} &
        \includegraphics[width=0.16\textwidth]{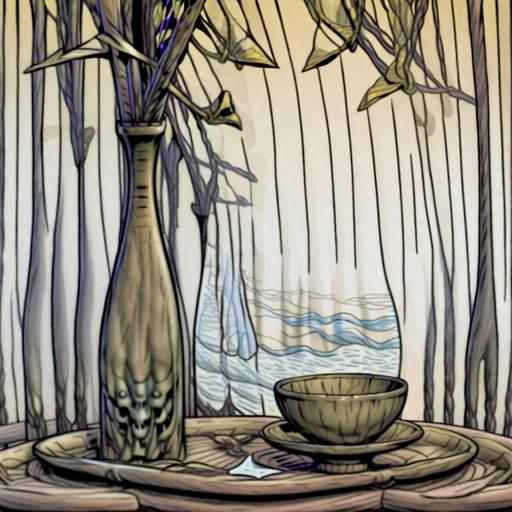} &
        \includegraphics[width=0.16\textwidth]{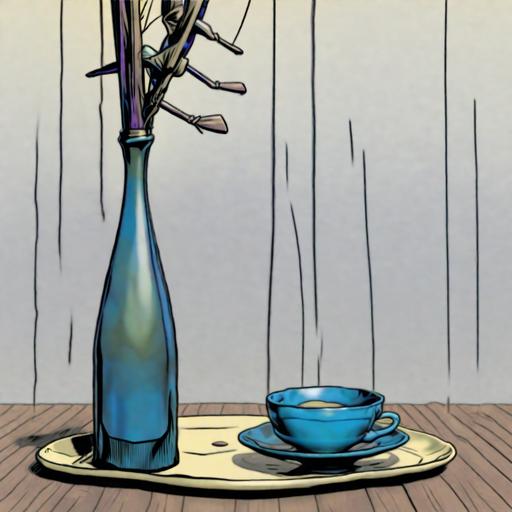} &
        \includegraphics[width=0.16\textwidth]{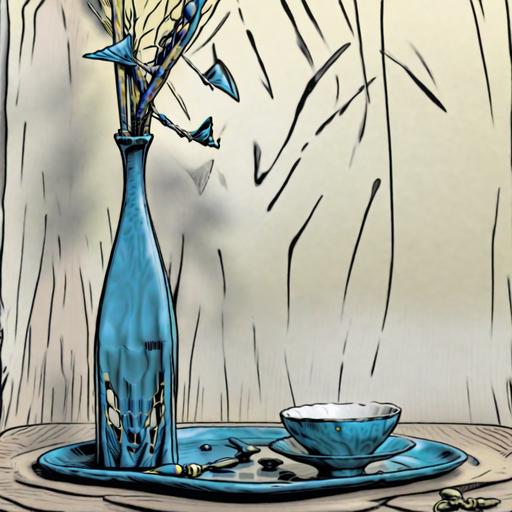} 
        \\ 
         
    \end{tabular}
    }   
    \caption{\textbf{Additional results of the ablation study.} We evaluate three variants of our model: replacing $x_0$-prediction with $\epsilon$-prediction (w/o $x_0$-prediction), 2) removing the two-step training strategy for style LoRA (w/o two-step training), and 3) using x0-prediction alone instead of loss transition for content LoRA (w/o loss transition).}
    \label{fig:additional_ablation}
\end{figure*}

As described in \cref{sec:ablation}, we conduct an ablation study by removing
each component of our method.
Specifically, we evaluate three variants: 1) replacing $x_0$-prediction with $\epsilon$-prediction (w/o $x_0$-prediction), 2) removing the two-step training strategy for style LoRA (w/o two-step training), and 3) using x0-prediction alone instead of loss transition for content LoRA (w/o loss transition).
In \cref{fig:additional_ablation}, we provide additional ablation study results for each variant. The results demonstrate the crucial role of each component.

\section{Comparing $\epsilon$-prediction with $x_0$-prediction} 
\label{sec:appendix_epsilon_x0}
In \cref{fig:epsilon_x0}, we present a qualitative comparison of four different loss schemes used for training the content LoRA: 1) using $\epsilon$-prediction only, 2) using $x_0$-prediction only, 3) transitioning from $x_0$-prediction to $\epsilon$-prediction ($x_0 \rightarrow \epsilon$), and 4) transitioning from $\epsilon$-prediction to $x_0$-prediction ($\epsilon \rightarrow x_0$). Similar to B-LoRA, $\epsilon$-prediction alone does not effectively capture the global structure of the content image. In contrast, $x_0$-prediction alone more accurately captures the global structure but falls short in retaining some local details. The transition of $x_0 \rightarrow \epsilon$ achieves a compromised performance between $\epsilon$-prediction and $x_0$-prediction. The proposed transition of $\epsilon \rightarrow x_0$ achieves the best performance in content preservation, accurately capturing both global structure and local details. This suggests that directly switching the original pre-training loss to a new loss may make the model struggle to adapt.

\section{Inference Guidance} \label{sec:appendix_guidance}
\begin{figure*}[t]
    \centering
    \setlength{\tabcolsep}{0.85pt}
    \renewcommand{\arraystretch}{0.5}
    {\small
    \begin{tabular}{c c@{\hspace{0.1cm}} | @{\hspace{0.1cm}}c c c c c}
 
    Content & \multicolumn{1}{c@{}}{Style} & & \multicolumn{2}{c}{Inference guidance strength (content$\uparrow$)} & &   \\ 

    \includegraphics[width=0.137\textwidth]{images/cnt-sty/content/modern.jpg} &
    \includegraphics[width=0.137\textwidth]{images/cnt-sty/style/rabbit.jpg}&
    \includegraphics[width=0.137\textwidth]{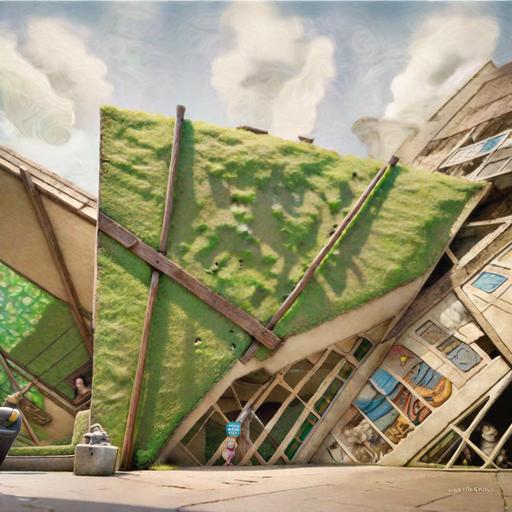} &
    \includegraphics[width=0.137\textwidth]{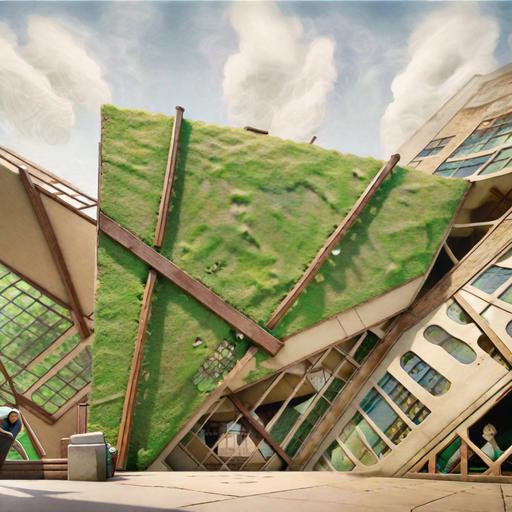} &
    \includegraphics[width=0.137\textwidth]{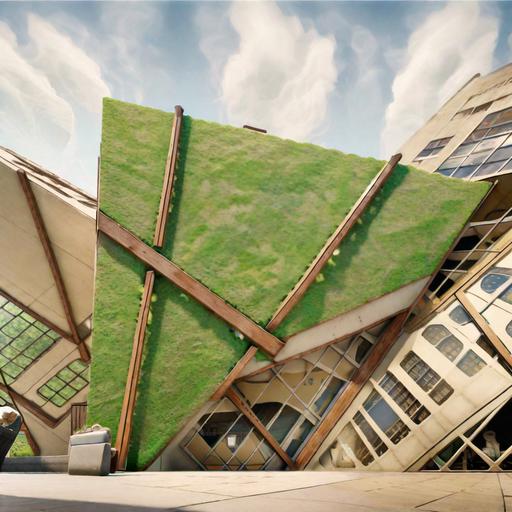} &
    \includegraphics[width=0.137\textwidth]{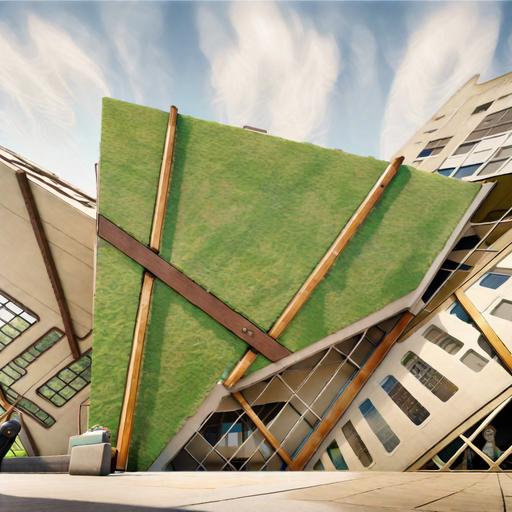} &
    \includegraphics[width=0.137\textwidth]{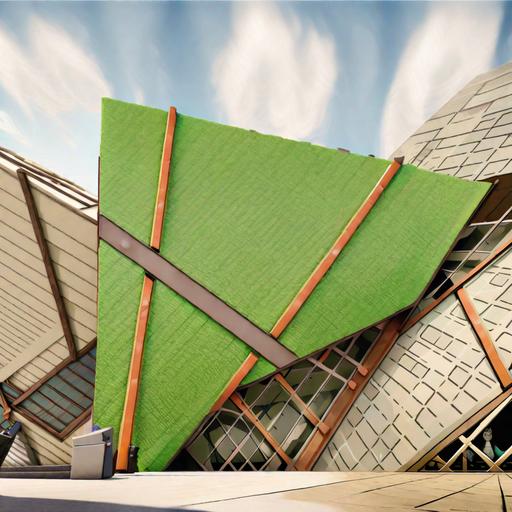} \\
    
    & \multicolumn{1}{c@{}}{} & \multicolumn{1}{c}{0.0} & 1.0 & 2.0 & 3.0 & 4.0  \\ [1.5ex]

    Content& \multicolumn{1}{c@{}}{Style} & & \multicolumn{2}{c}{LoRA weight scaling (content$\uparrow$)} & &  \\ 

    \includegraphics[width=0.137\textwidth]{images/cnt-sty/content/modern.jpg} &
    \includegraphics[width=0.137\textwidth]{images/cnt-sty/style/rabbit.jpg}&
    \includegraphics[width=0.137\textwidth]{images/appendix/ours/modern-stylized-rabbit.jpg} &
    \includegraphics[width=0.137\textwidth]{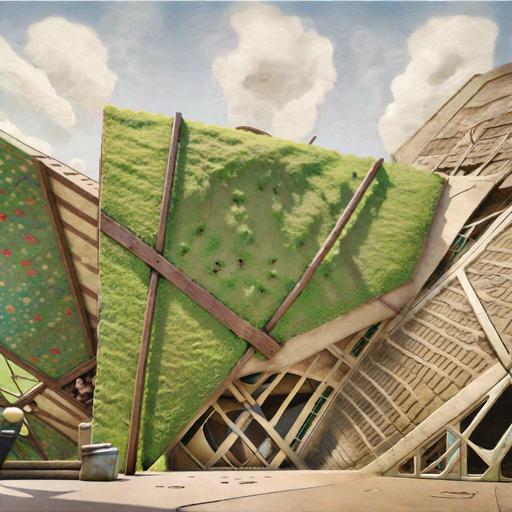} &
    \includegraphics[width=0.137\textwidth]{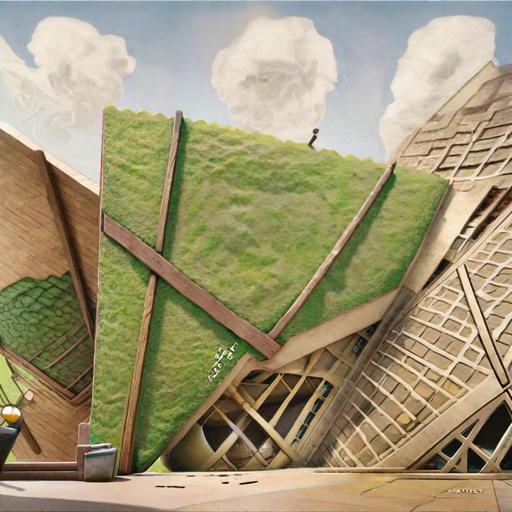} &
    \includegraphics[width=0.137\textwidth]{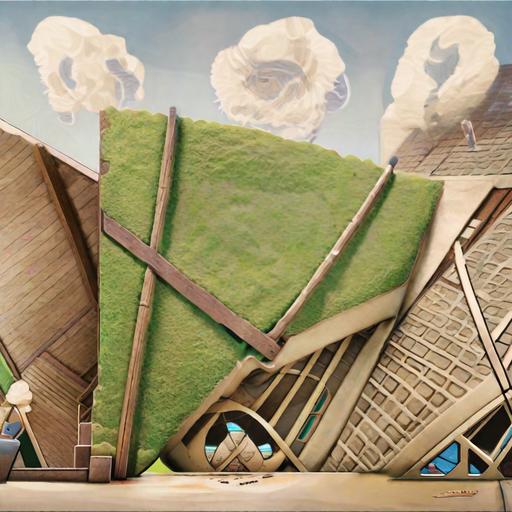} &
    \includegraphics[width=0.137\textwidth]{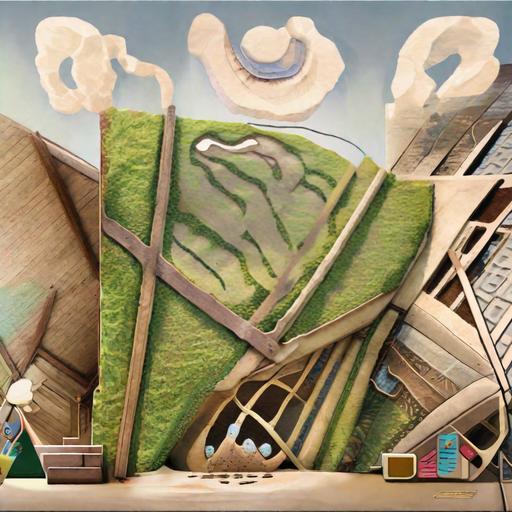} \\ 

    & \multicolumn{1}{c}{} & \multicolumn{1}{c}{1.0} & 1.1 & 1.2 & 1.3 & 1.4  \\
    
    \noalign{\vskip 1.5ex}\hline\noalign{\vskip 1.5ex}

     Content & \multicolumn{1}{c@{}}{Style} & & \multicolumn{2}{c}{Inference guidance strength (style$\uparrow$)} & &  \\ 

    \includegraphics[width=0.137\textwidth]{images/cnt-sty/content/modern.jpg} &
    \includegraphics[width=0.137\textwidth]{images/cnt-sty/style/rabbit.jpg} &
    \includegraphics[width=0.137\textwidth]{images/appendix/ours/modern-stylized-rabbit.jpg} &
    \includegraphics[width=0.137\textwidth]{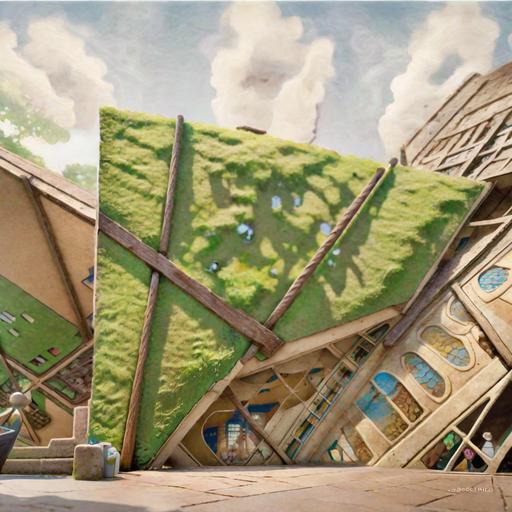} &
    \includegraphics[width=0.137\textwidth]{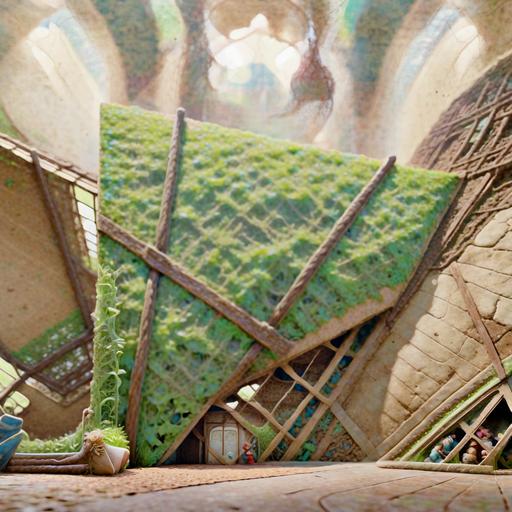} & 
    \includegraphics[width=0.137\textwidth]{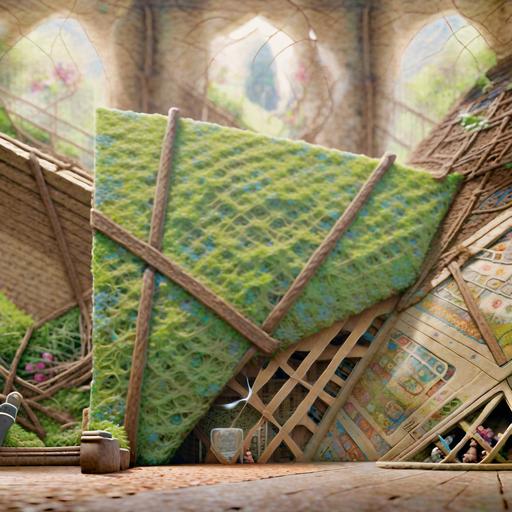} &
    \includegraphics[width=0.137\textwidth]{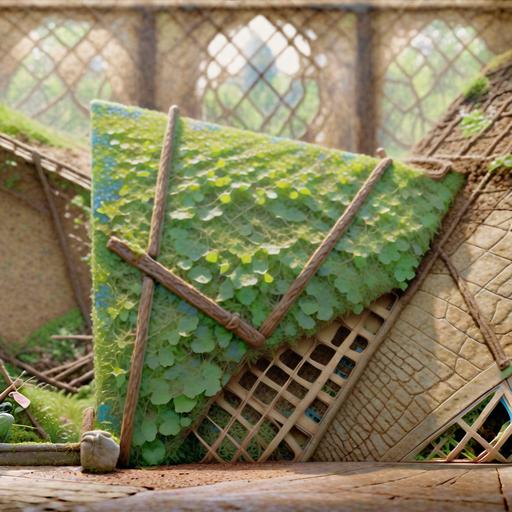} \\
    
    & \multicolumn{1}{c@{}}{} & \multicolumn{1}{c}{0.0} & 1.0 & 2.0 & 3.0 & 4.0  \\ [1.5ex] 
    
    Content & \multicolumn{1}{c@{}}{Style} & & \multicolumn{2}{c}{LoRA weight scaling (style$\uparrow$)} & &  \\

    \includegraphics[width=0.137\textwidth]{images/cnt-sty/content/modern.jpg} &
    \includegraphics[width=0.137\textwidth]{images/cnt-sty/style/rabbit.jpg} &
    \includegraphics[width=0.137\textwidth]{images/appendix/ours/modern-stylized-rabbit.jpg} &
    \includegraphics[width=0.137\textwidth]{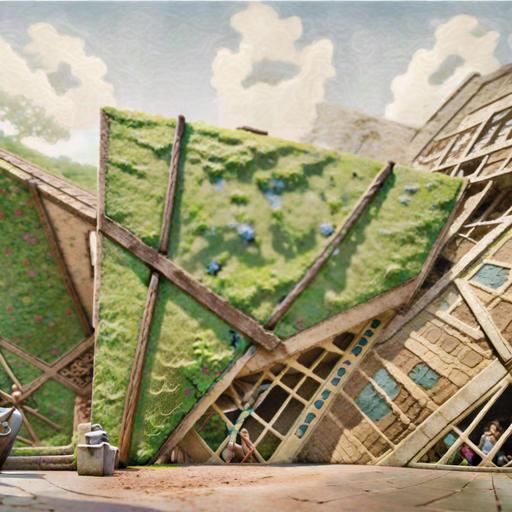} &
    \includegraphics[width=0.137\textwidth]{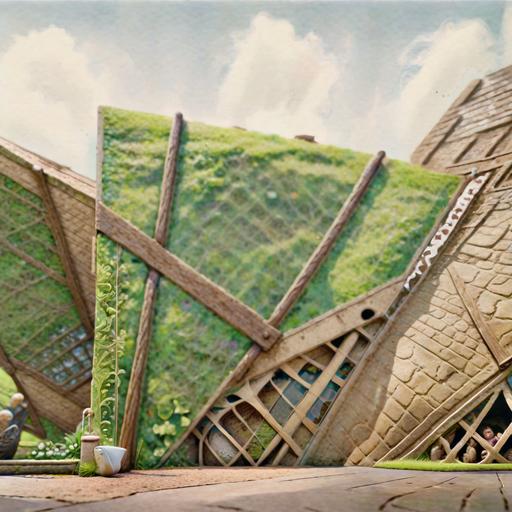} &
    \includegraphics[width=0.137\textwidth]{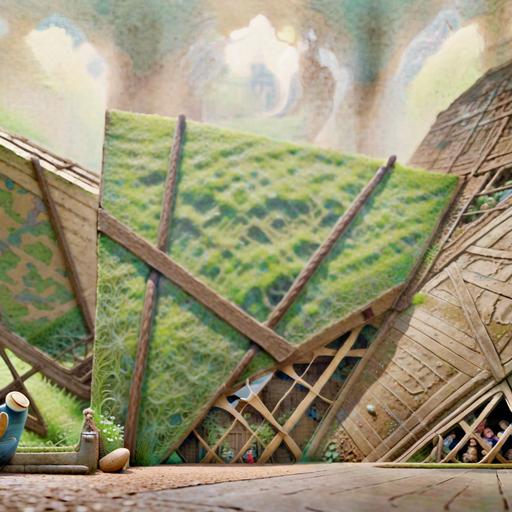} &
    \includegraphics[width=0.137\textwidth]{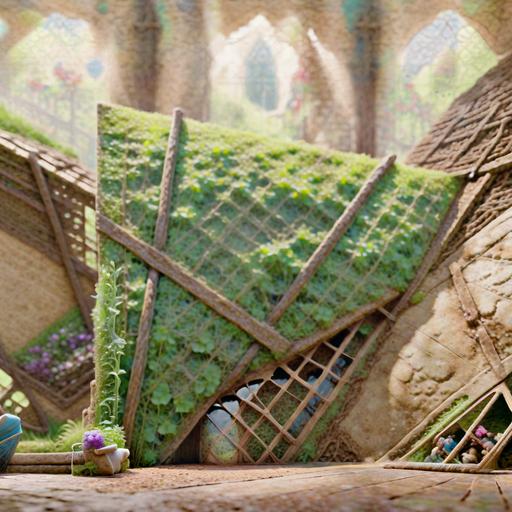} \\
    
    & \multicolumn{1}{c}{} & \multicolumn{1}{c}{1.0} & 1.1 & 1.2 & 1.3 & 1.4  \\

    \end{tabular}
    }   
    \caption{
    Qualitative comparison between the proposed inference guidance and scaling LoRA weight. 
    }
    \label{fig:guidance_comparison1}
\end{figure*}

\begin{figure*}[t]
    \centering
    \setlength{\tabcolsep}{0.85pt}
    \renewcommand{\arraystretch}{0.5}
    {\small
    \begin{tabular}{c c@{\hspace{0.1cm}} | @{\hspace{0.1cm}}c c c c c}
 
    Content & \multicolumn{1}{c@{}}{Style} & & \multicolumn{2}{c}{Inference guidance strength (content$\uparrow$)} & &   \\ 
    
    \includegraphics[width=0.137\textwidth]{images/cnt-sty/content/sofa.jpg} &
    \includegraphics[width=0.137\textwidth]{images/cnt-sty/style/pirate.jpg} &
    \includegraphics[width=0.137\textwidth]{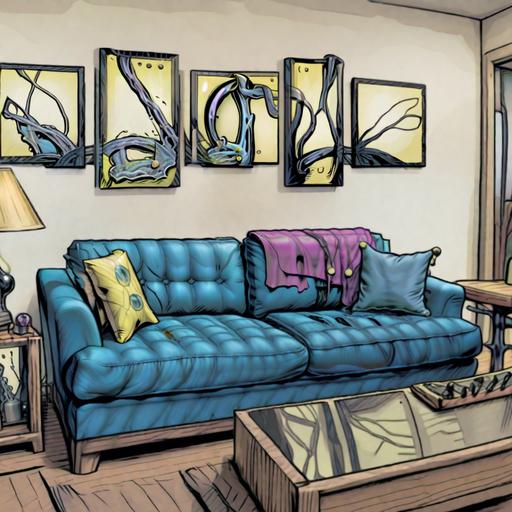} &
    \includegraphics[width=0.137\textwidth]{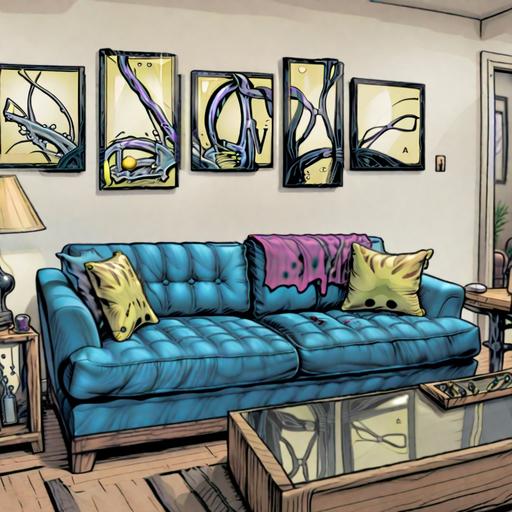} &
    \includegraphics[width=0.137\textwidth]{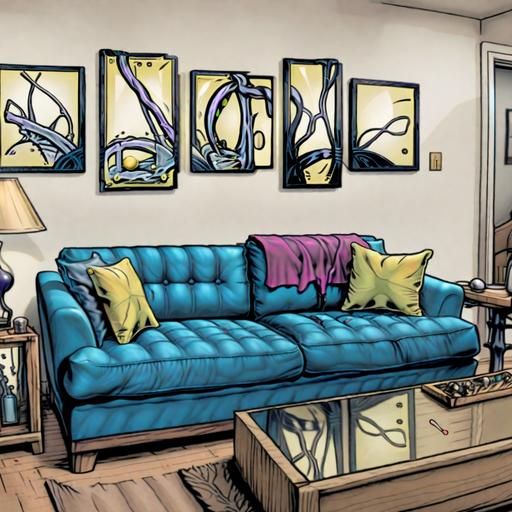} &
    \includegraphics[width=0.137\textwidth]{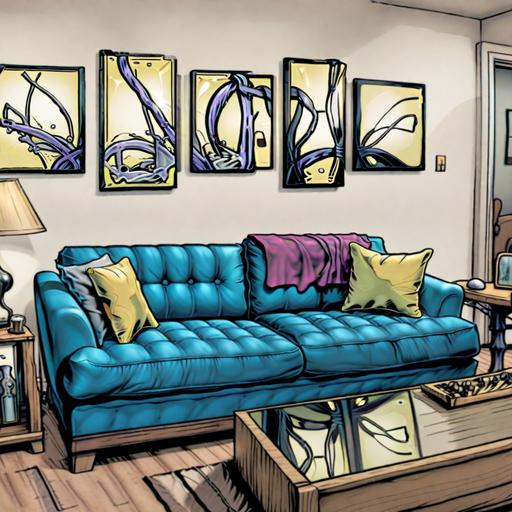} &
    \includegraphics[width=0.137\textwidth]{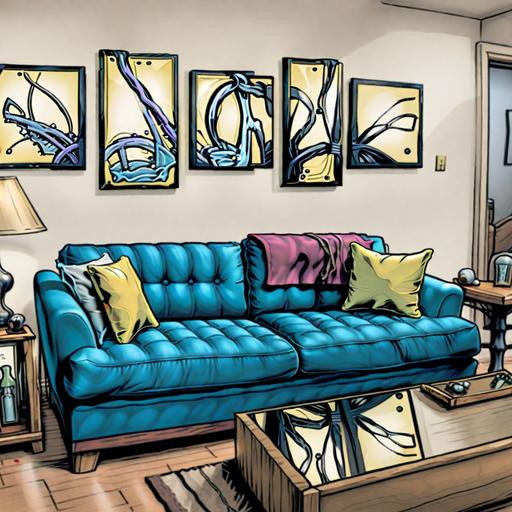} \\
    
    & \multicolumn{1}{c@{}}{} & \multicolumn{1}{c}{0.0} & 1.0 & 2.0 & 3.0 & 4.0  \\ [1.5ex]

    Content& \multicolumn{1}{c@{}}{Style} & & \multicolumn{2}{c}{LoRA weight scaling (content$\uparrow$)} & &  \\ 

    \includegraphics[width=0.137\textwidth]{images/cnt-sty/content/sofa.jpg} &
    \includegraphics[width=0.137\textwidth]{images/cnt-sty/style/pirate.jpg}&
    \includegraphics[width=0.137\textwidth]{images/appendix/ours/sofa-stylized-pirate.jpg} &
    \includegraphics[width=0.137\textwidth]{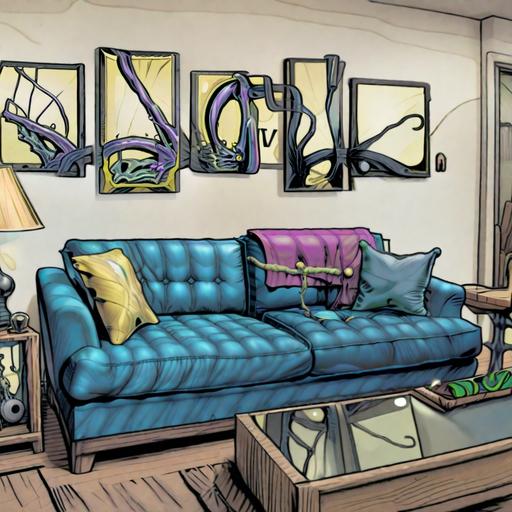} &
    \includegraphics[width=0.137\textwidth]{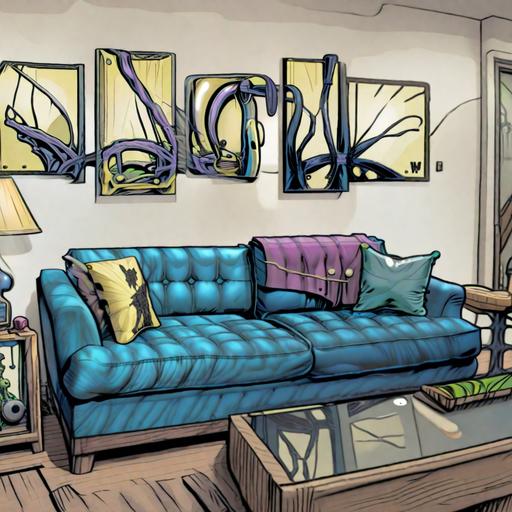} &
    \includegraphics[width=0.137\textwidth]{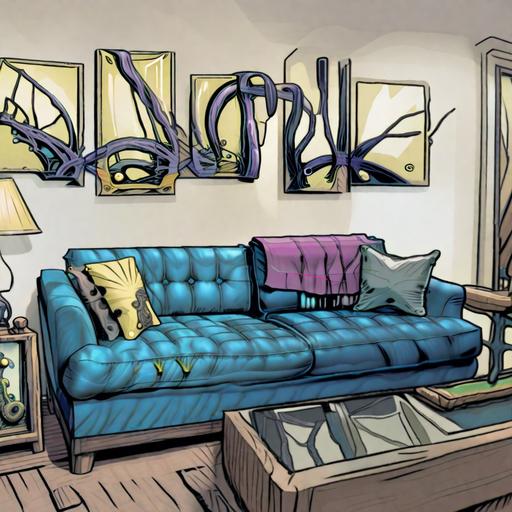} &
    \includegraphics[width=0.137\textwidth]{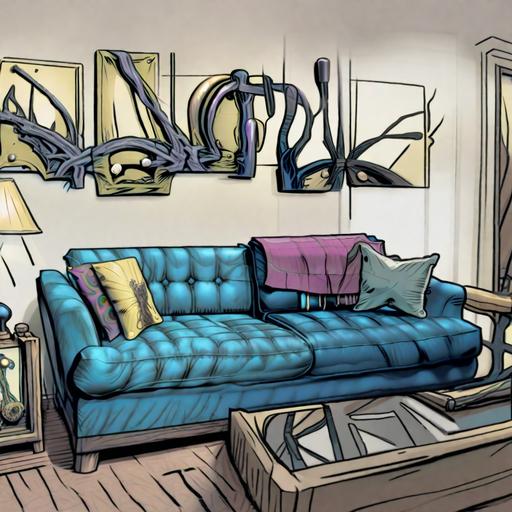} \\ 

    & \multicolumn{1}{c}{} & \multicolumn{1}{c}{1.0} & 1.1 & 1.2 & 1.3 & 1.4  \\
    
    \noalign{\vskip 1.5ex}\hline\noalign{\vskip 1.5ex}

     Content & \multicolumn{1}{c@{}}{Style} & & \multicolumn{2}{c}{Inference guidance strength (style$\uparrow$)} & &  \\ 

    \includegraphics[width=0.137\textwidth]{images/cnt-sty/content/sofa.jpg} &
    \includegraphics[width=0.137\textwidth]{images/cnt-sty/style/pirate.jpg} &
    \includegraphics[width=0.137\textwidth]{images/appendix/ours/sofa-stylized-pirate.jpg} &
    \includegraphics[width=0.137\textwidth]{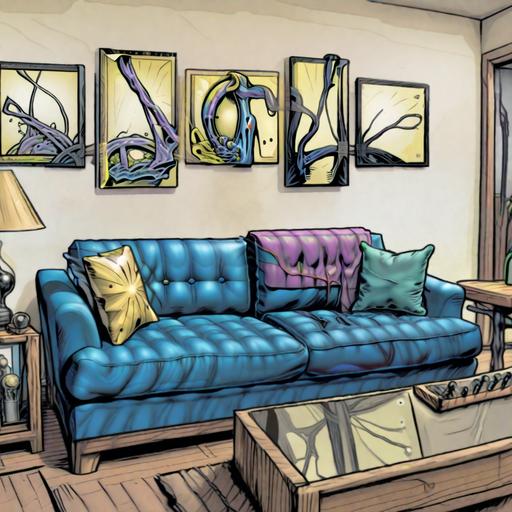} &
    \includegraphics[width=0.137\textwidth]{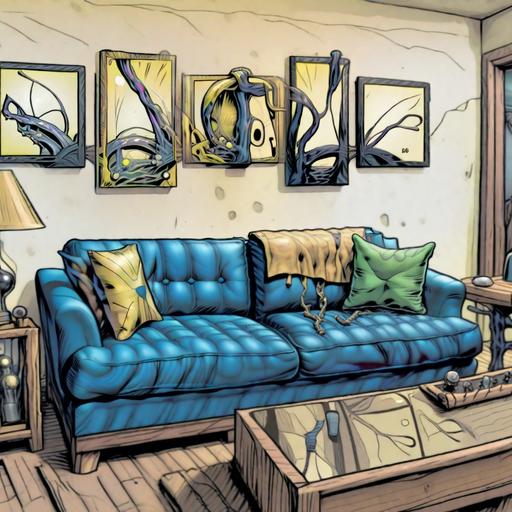} & 
    \includegraphics[width=0.137\textwidth]{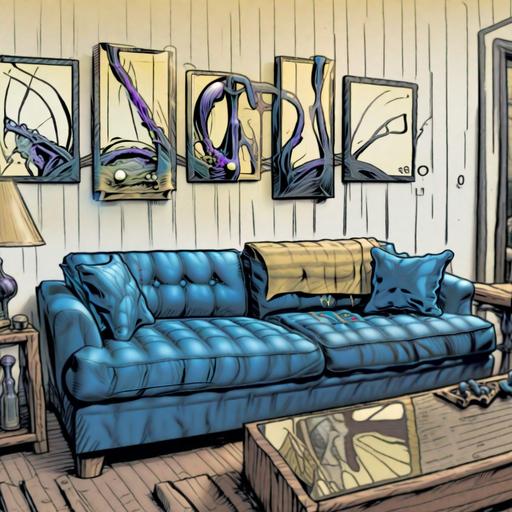} &
    \includegraphics[width=0.137\textwidth]{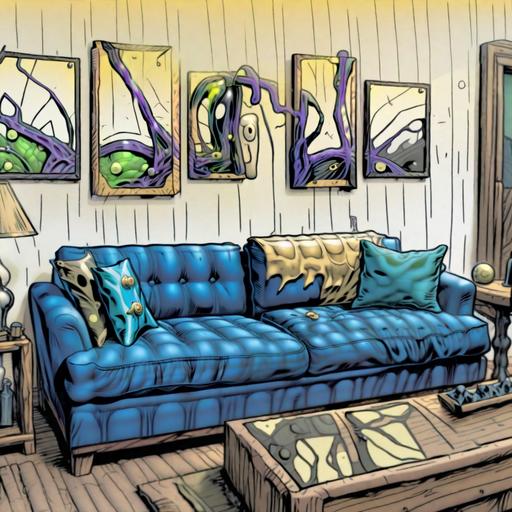} \\
    
    & \multicolumn{1}{c@{}}{} & \multicolumn{1}{c}{0.0} & 1.0 & 2.0 & 3.0 & 4.0  \\ [1.5ex] 
    
    Content & \multicolumn{1}{c@{}}{Style} & & \multicolumn{2}{c}{LoRA weight scaling (style$\uparrow$)} & &  \\

    \includegraphics[width=0.137\textwidth]{images/cnt-sty/content/sofa.jpg} &
    \includegraphics[width=0.137\textwidth]{images/cnt-sty/style/pirate.jpg} &
    \includegraphics[width=0.137\textwidth]{images/appendix/ours/sofa-stylized-pirate.jpg} &
    \includegraphics[width=0.137\textwidth]{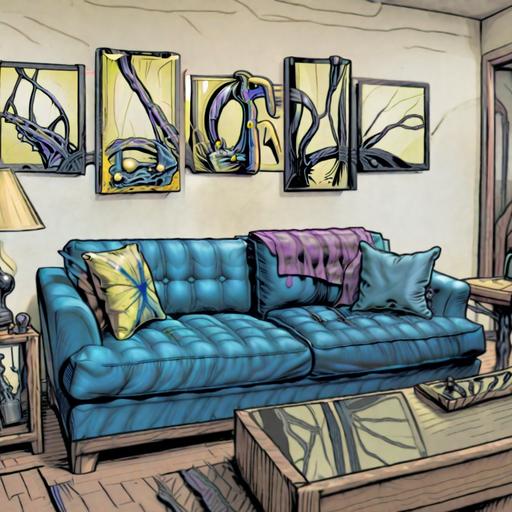} &
    \includegraphics[width=0.137\textwidth]{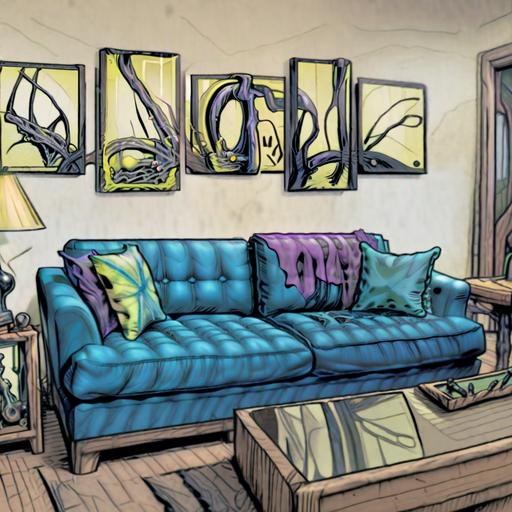} & 
    \includegraphics[width=0.137\textwidth]{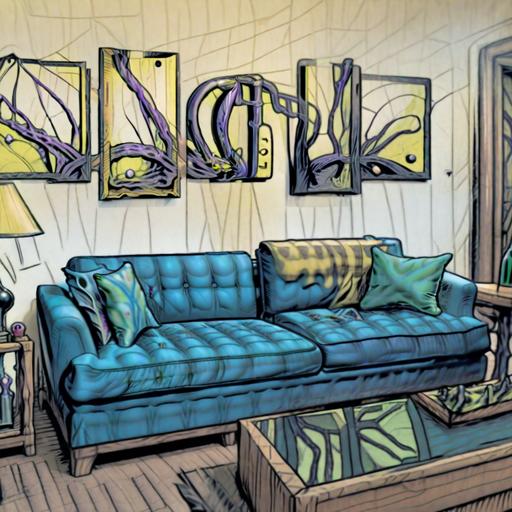} &
    \includegraphics[width=0.137\textwidth]{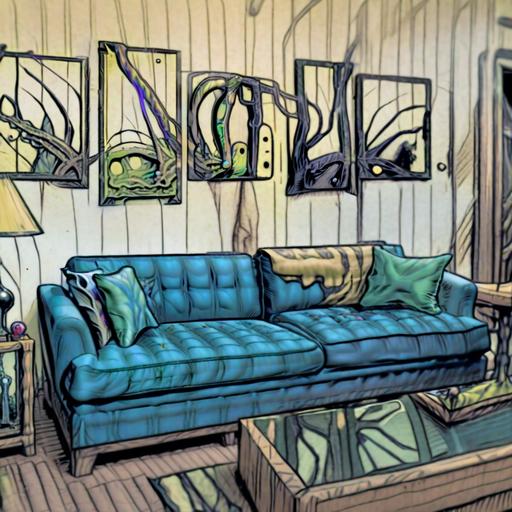} \\
    
    & \multicolumn{1}{c}{} & \multicolumn{1}{c}{1.0} & 1.1 & 1.2 & 1.3 & 1.4  \\

    \end{tabular}
    }   
    \caption{
    Qualitative comparison between the proposed inference guidance and scaling LoRA weight. 
    }
    \label{fig:guidance_comparison2}
\end{figure*}

In \cref{fig:guidance_comparison1,fig:guidance_comparison2}, we compare the proposed inference guidance with the method of scaling LoRA weights for controlling the content and style strengths. When increasing the content strength, both approaches effectively enhance the impact from the content image. However, scaling LoRA weights sometimes leads to distortions in some local details. In terms of adjusting style strength, both methods exhibit comparable performance, successfully generating high-quality stylized images.

\section{Visualization of Attention Maps}
\begin{figure*}[t]
    \centering
    \setlength{\tabcolsep}{0.85pt}
    \renewcommand{\arraystretch}{0.5}
    {
    \begin{tabular}{c @{\hspace{0.1cm}} | @{\hspace{0.1cm}} c c @{\hspace{0.1cm}} | @{\hspace{0.1cm}} c c}

         \multicolumn{1}{c}{Input}
         & \multicolumn{1}{c}{Ours}
         & \multicolumn{1}{c}{B-LoRA}
         & \multicolumn{1}{c}{Ours}
         & B-LoRA \\
        \includegraphics[width=0.16\textwidth]{images/cnt-sty/style/ink_painting.jpg} &
        \includegraphics[width=0.16\textwidth]{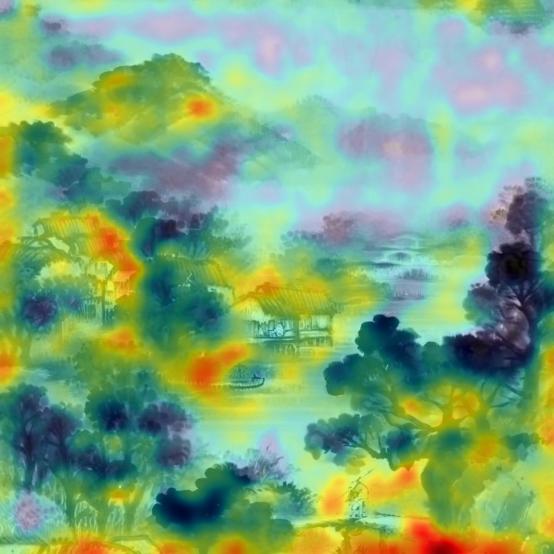} &
        \includegraphics[width=0.16\textwidth]{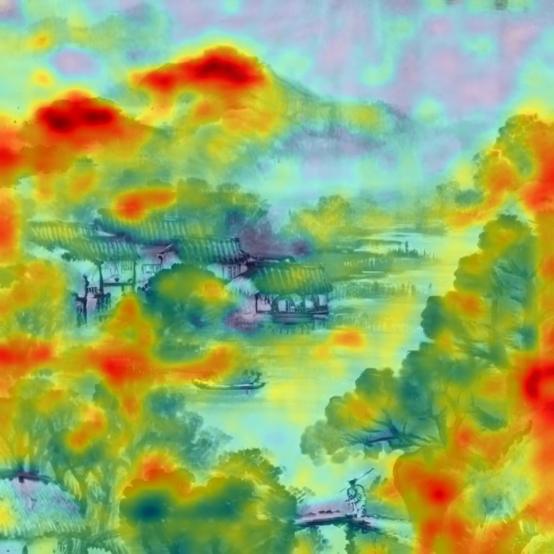} &
        \includegraphics[width=0.16\textwidth]{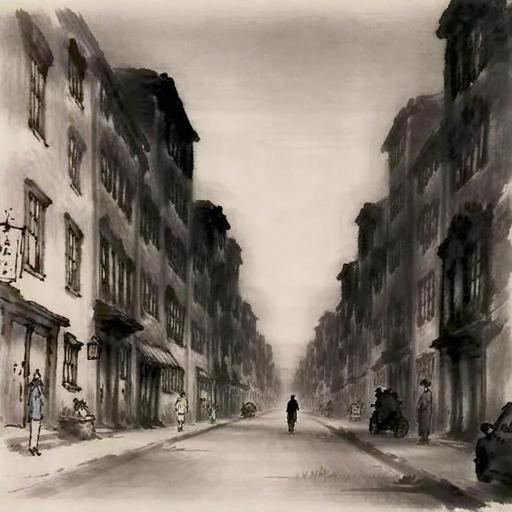} &
        \includegraphics[width=0.16\textwidth]{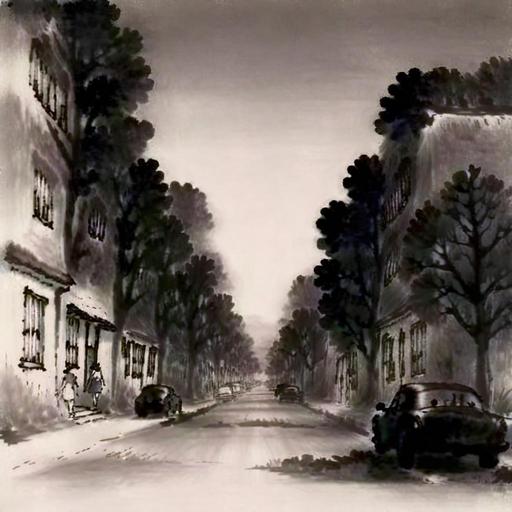} 
        \\
        \multicolumn{1}{c}{}&\multicolumn{2}{c}{Attention map visualization of ``[v]''} & \multicolumn{2}{c}{``A street in the style of [v]''} \\
        \includegraphics[width=0.16\textwidth]{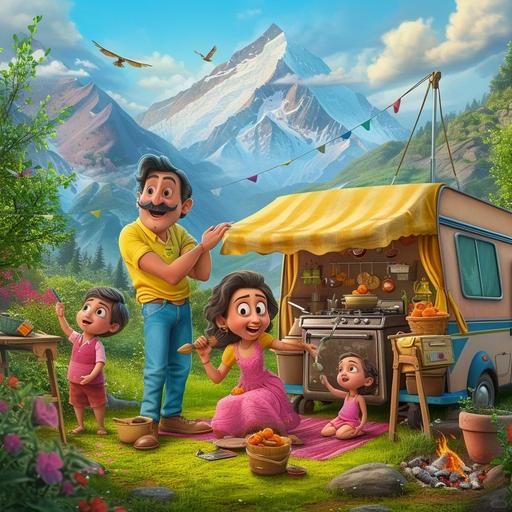} &
        \includegraphics[width=0.16\textwidth]{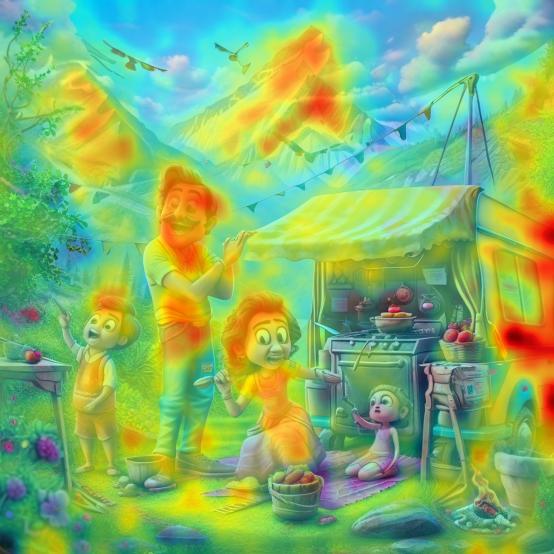} &
        \includegraphics[width=0.16\textwidth]{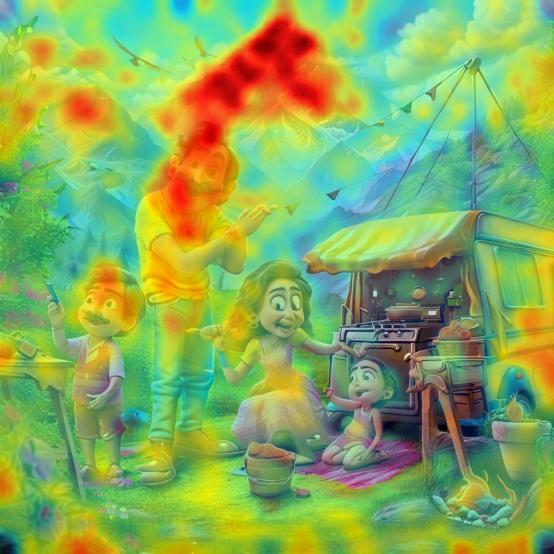} &
        \includegraphics[width=0.16\textwidth]{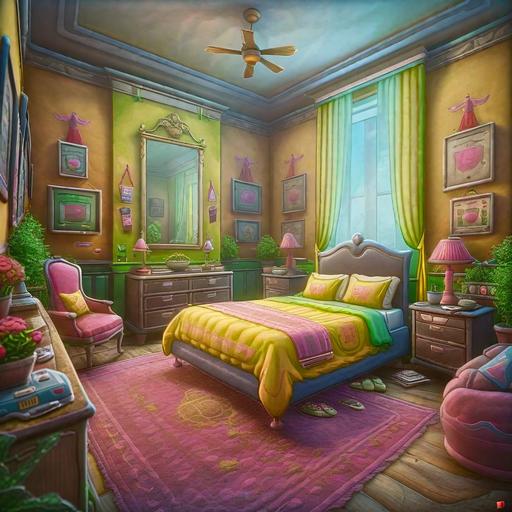} &
        \includegraphics[width=0.16\textwidth]{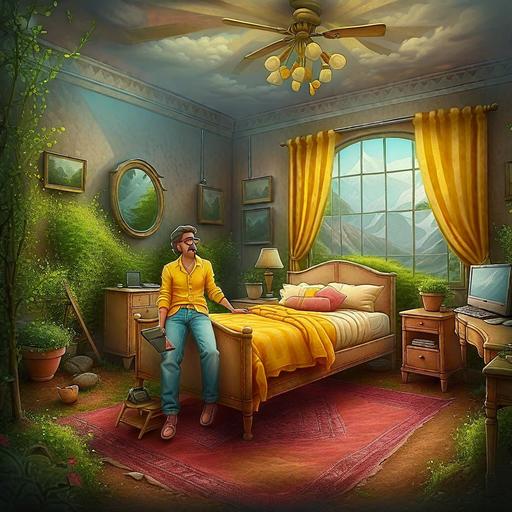} 
        \\ 
        \multicolumn{1}{c}{}&\multicolumn{2}{c}{Attention map visualization of ``[v]''} & \multicolumn{2}{c}{``A bedroom in the style of [v]''} \\         
    \end{tabular}
    }   
    \caption{
    We visualize the attention maps corresponding to the style token ``[v]'' and observe that B-LoRA tends to focus more on certain local details compared to our method. This focus contributes to the issues of content leakage and style misalignment.
    }
    \label{fig:attention_map}
\end{figure*}

In \cref{fig:attention_map}, we visualize the attention maps corresponding to the style token ``[v]'' for both our method and B-LoRA. As shown, B-LoRA tends to focus more on certain local details compared to our method. This emphasis on local details contributes to issues of content leakage and style misalignment in B-LoRA.

\section{Portrait Stylization}
\label{sec:face}
\begin{figure*}[t]
    \centering
    \setlength{\tabcolsep}{0.85pt}
    \renewcommand{\arraystretch}{0.5}
    {
    \begin{tabular}{c@{\hspace{0.1cm}} | @{\hspace{0.1cm}}c c c c c}

         \multicolumn{1}{c@{}}{Content Image}
         &&&\multicolumn{1}{c}{Style Image}&&  
         \\

        \includegraphics[width=0.16\textwidth]{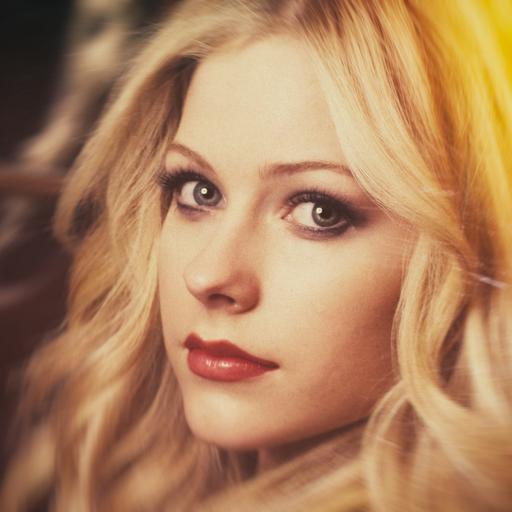} &
        \includegraphics[width=0.16\textwidth]{images/cnt-sty/style/oil_painting.jpg} &
        \includegraphics[width=0.16\textwidth]{images/cnt-sty/style/cartoon_line.jpg} &
        \includegraphics[width=0.16\textwidth]{images/cnt-sty/style/rabbit.jpg} &
        \includegraphics[width=0.16\textwidth]{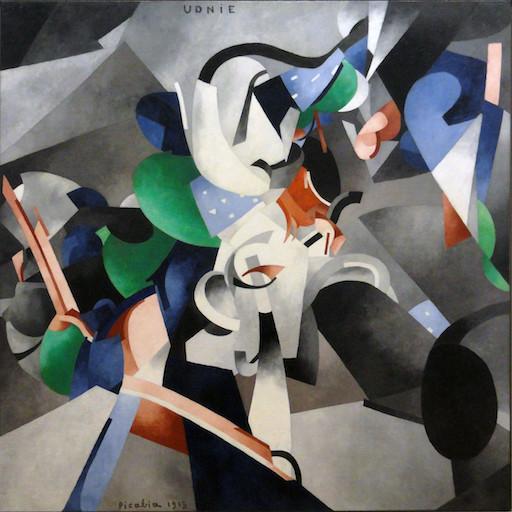} &
        \includegraphics[width=0.16\textwidth]{images/cnt-sty/style/mountain.jpg} 
        \\ 
        \noalign{\vskip 0.02cm}\hline\noalign{\vskip 0.07cm}

        \raisebox{0.55in}{Ours}&
        \includegraphics[width=0.16\textwidth]{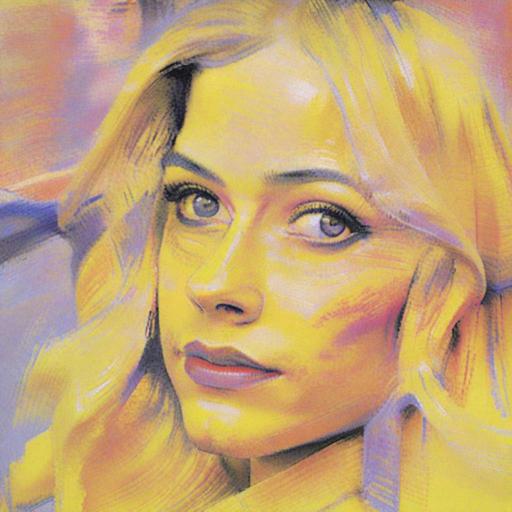} &
        \includegraphics[width=0.16\textwidth]{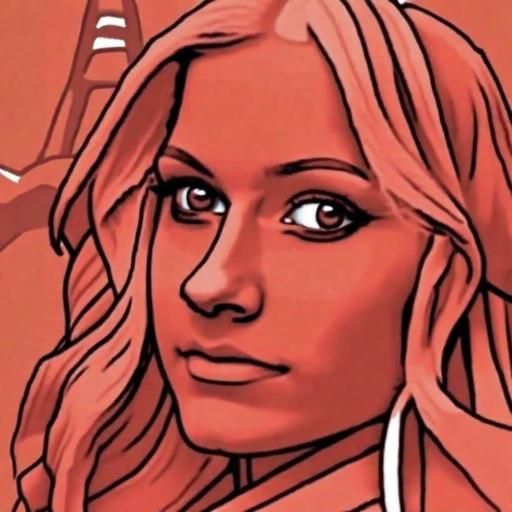} &
        \includegraphics[width=0.16\textwidth]{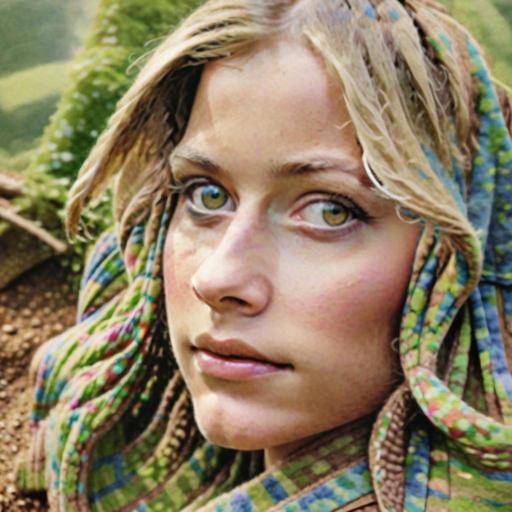} &
        \includegraphics[width=0.16\textwidth]{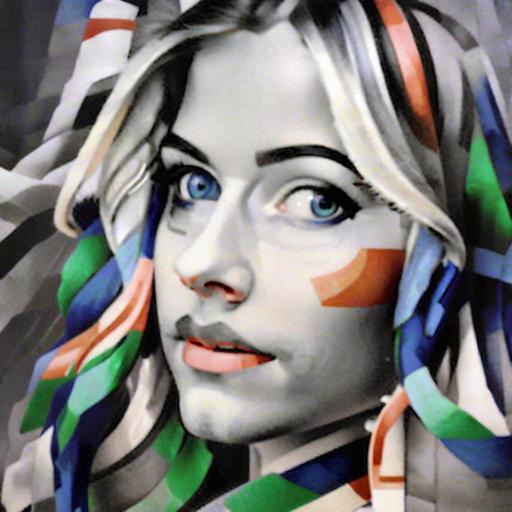} &
        \includegraphics[width=0.16\textwidth]{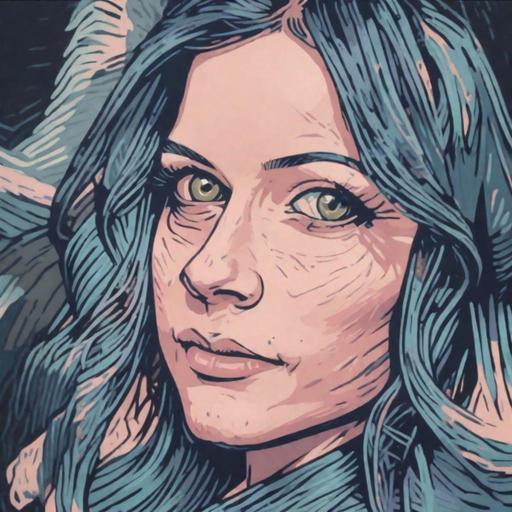} 
        \\

        \raisebox{0.55in}{B-LoRA}&
        \includegraphics[width=0.16\textwidth]{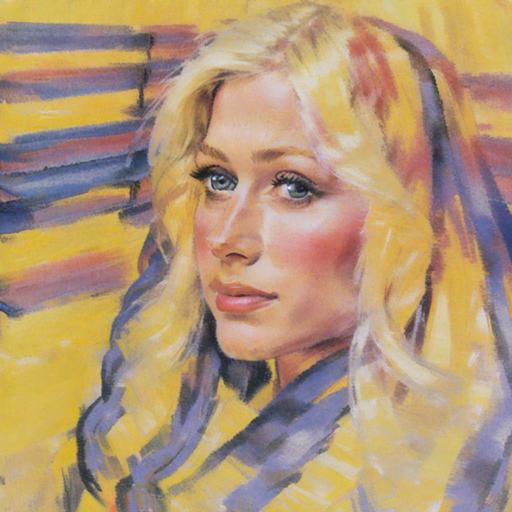} &
        \includegraphics[width=0.16\textwidth]{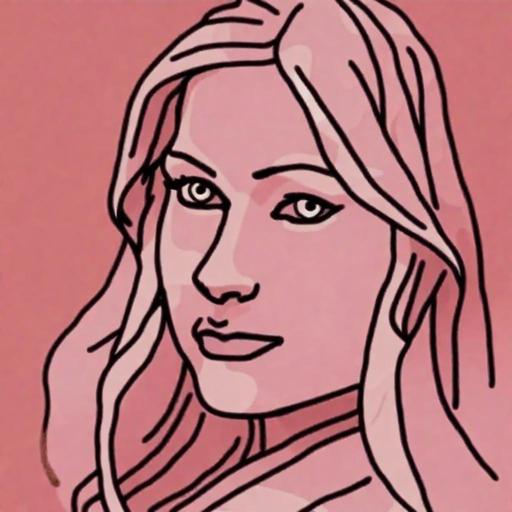} &
        \includegraphics[width=0.16\textwidth]{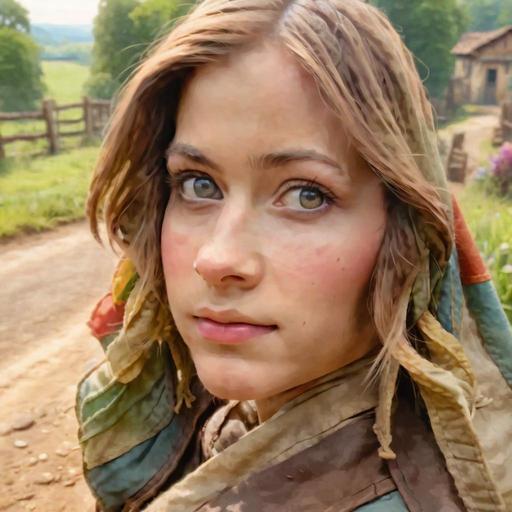} &
        \includegraphics[width=0.16\textwidth]{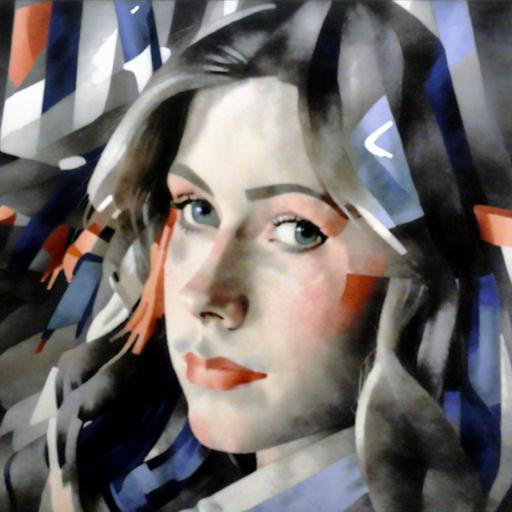} &
        \includegraphics[width=0.16\textwidth]{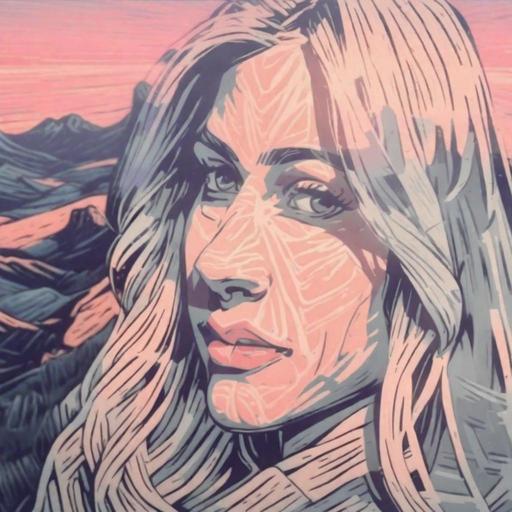} 
        \\
        
        \multicolumn{1}{c@{}}{Content Image}
         &&&\multicolumn{1}{c}{Style Image}&&  
         \\
         \includegraphics[width=0.16\textwidth]{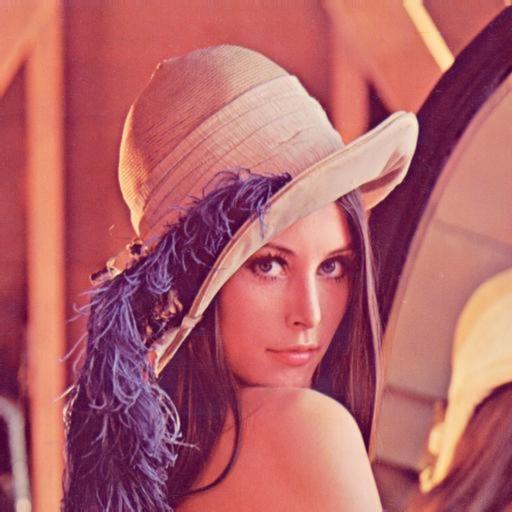} &
        \includegraphics[width=0.16\textwidth]{images/cnt-sty/style/rain_princess.jpg} &
        \includegraphics[width=0.16\textwidth]{images/cnt-sty/style/comic_deer.jpg} &
        \includegraphics[width=0.16\textwidth]{images/cnt-sty/style/orange2.jpg} &
        \includegraphics[width=0.16\textwidth]{images/cnt-sty/style/pig.jpg} &
        \includegraphics[width=0.16\textwidth]{images/cnt-sty/style/landscape.jpg} 
        \\ 
        \noalign{\vskip 0.02cm}\hline\noalign{\vskip 0.07cm}

        \raisebox{0.55in}{Ours}&
        \includegraphics[width=0.16\textwidth]{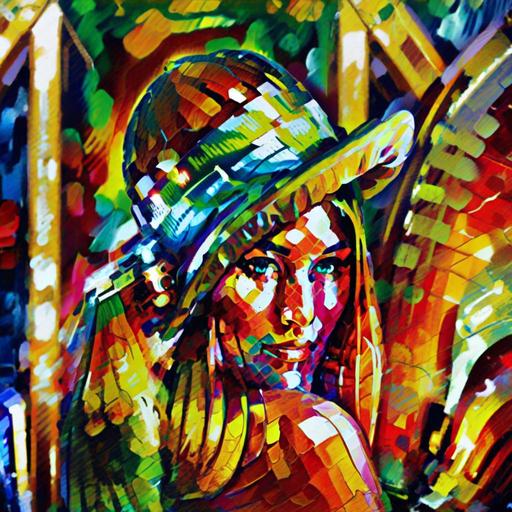} &
        \includegraphics[width=0.16\textwidth]{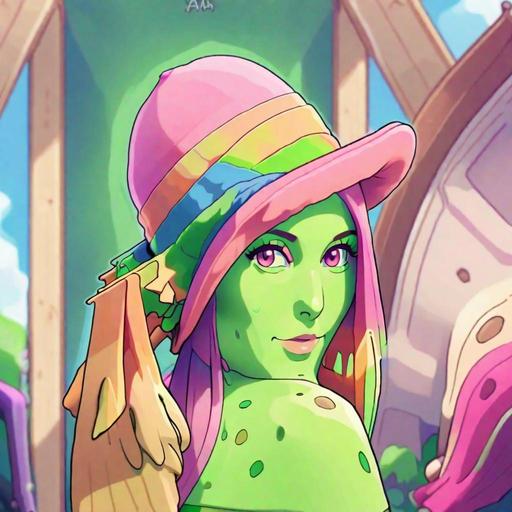} &
        \includegraphics[width=0.16\textwidth]{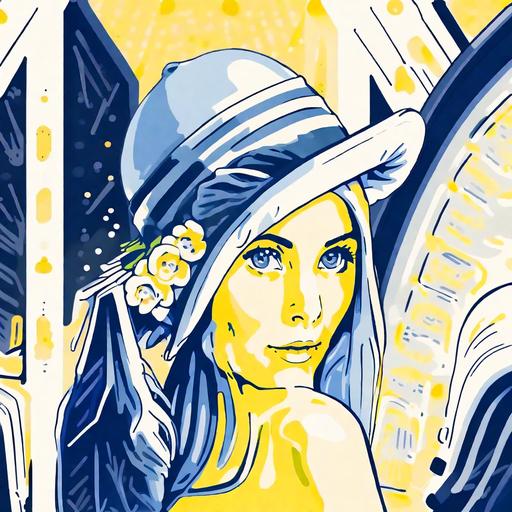} &
        \includegraphics[width=0.16\textwidth]{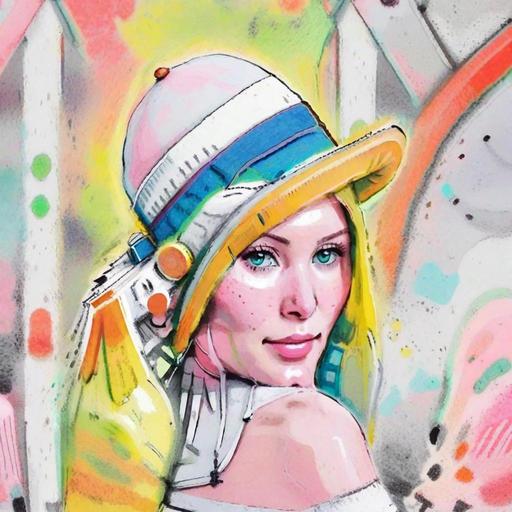} &
        \includegraphics[width=0.16\textwidth]{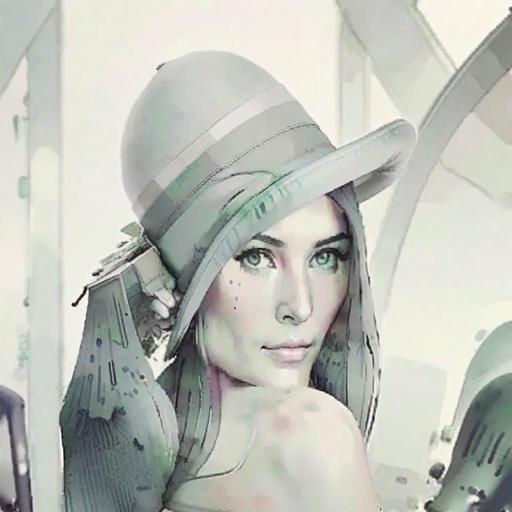} 
        \\

        \raisebox{0.55in}{B-LoRA}&
        \includegraphics[width=0.16\textwidth]{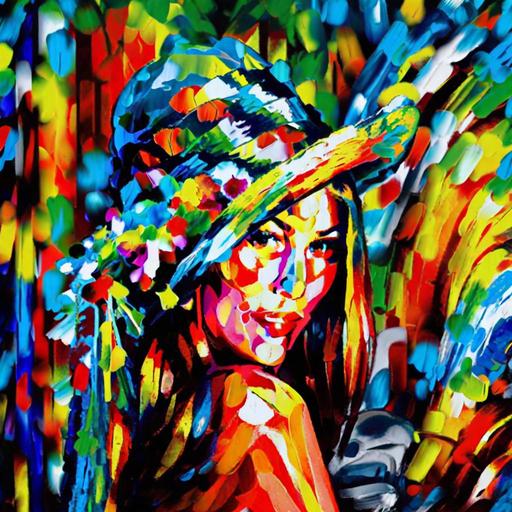} &
        \includegraphics[width=0.16\textwidth]{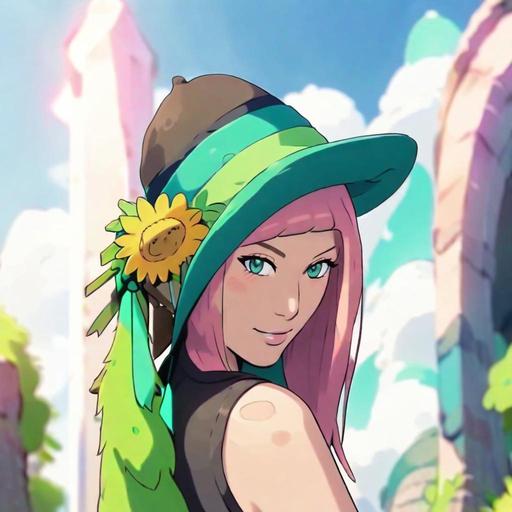} &
        \includegraphics[width=0.16\textwidth]{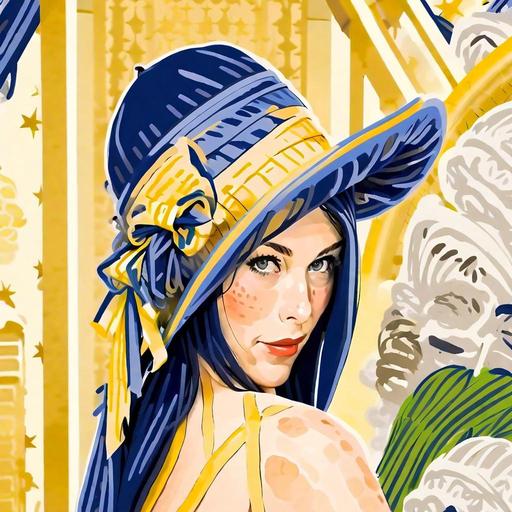}  &
        \includegraphics[width=0.16\textwidth]{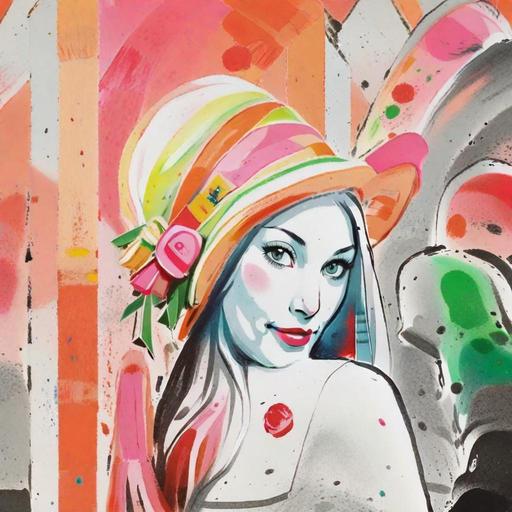}  &
        \includegraphics[width=0.16\textwidth]{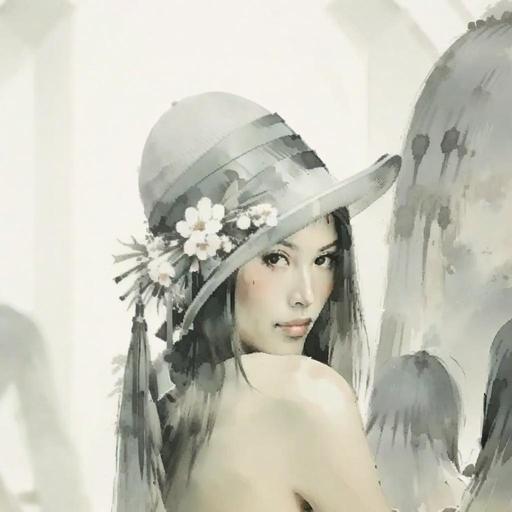} 
        \\
    \end{tabular}
    }   
    \caption{Qualitative comparison of portrait stylization between our method and B-LoRA. 
    }
    \label{fig:portrait}
\end{figure*}

In \cref{fig:portrait}, we present the portrait stylization results for both our method and B-LoRA. Our method demonstrates superior performance in preserving human identity compared to B-LoRA.

\afterpage{
\begin{figure*}[b]
    \centering
    \setlength{\tabcolsep}{0.85pt}
    \renewcommand{\arraystretch}{0.5}
    {
    \begin{tabular}{c c@{\hspace{0.1cm}} | @{\hspace{0.1cm}}c c c c}

         Content Image
         & \multicolumn{1}{c@{}}{Style Image}
         & \multicolumn{1}{c}{$\epsilon$-prediction}
         & $x_0$-prediction
         & $x_0 \rightarrow \epsilon$
         & $\epsilon \rightarrow x_0$ \\

        \includegraphics[width=0.16\textwidth]{images/cnt-sty/content/sofa.jpg} &
        \includegraphics[width=0.16\textwidth]{images/cnt-sty/style/cartoon_line.jpg} &
        \includegraphics[width=0.16\textwidth]{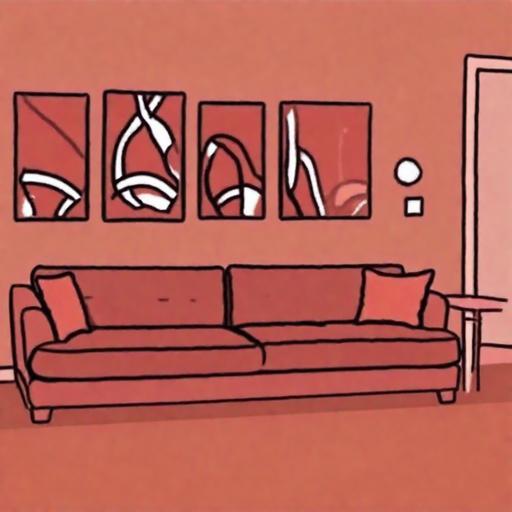} &
        \includegraphics[width=0.16\textwidth]{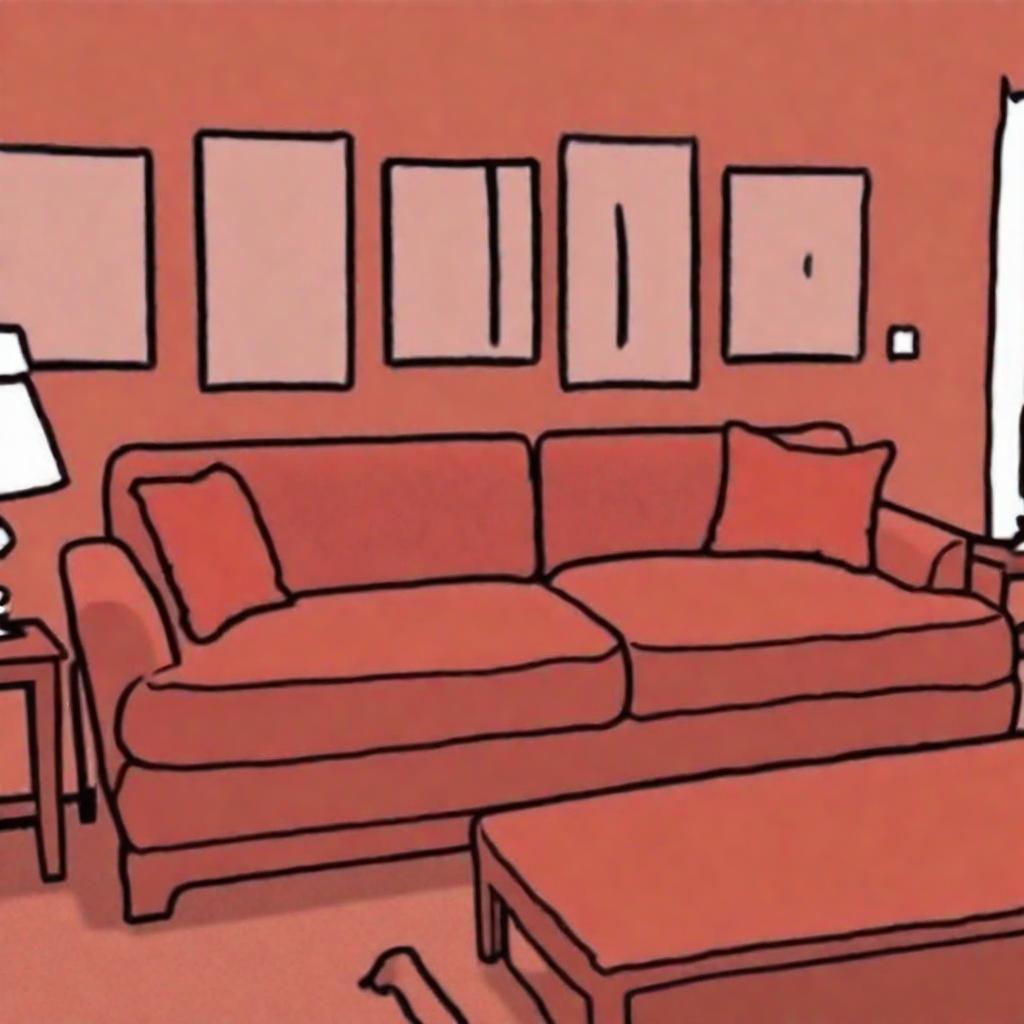} &
        \includegraphics[width=0.16\textwidth]{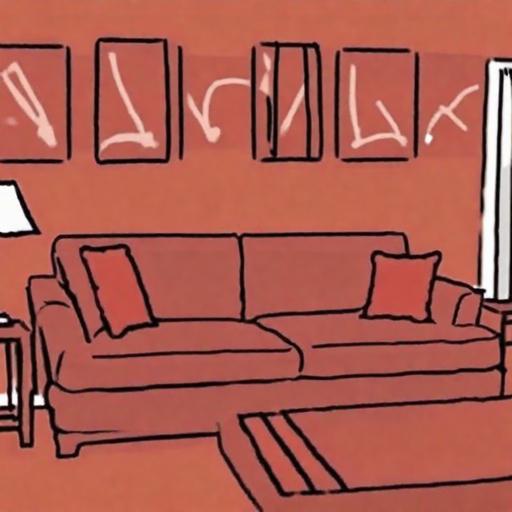} &
        \includegraphics[width=0.16\textwidth]{images/appendix/ours/sofa-stylized-cartoon_line.jpg} 
        \\ 

        \includegraphics[width=0.16\textwidth]{images/cnt-sty/content/teddy_bear.jpg} &
        \includegraphics[width=0.16\textwidth]{images/cnt-sty/style/eyes.jpg} &
        \includegraphics[width=0.16\textwidth]{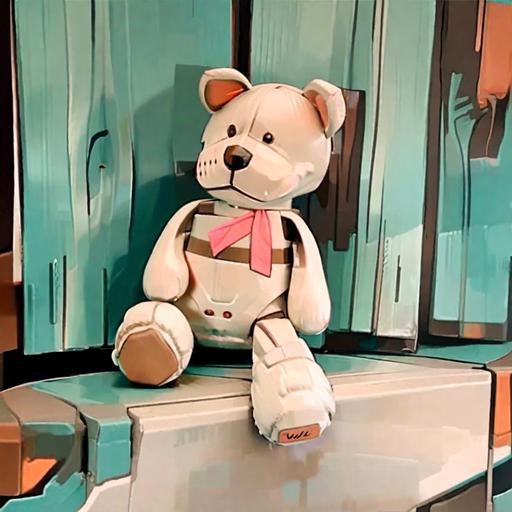} &
        \includegraphics[width=0.16\textwidth]{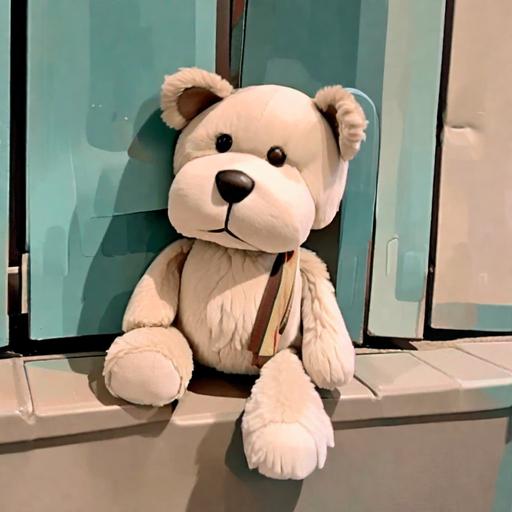} &
        \includegraphics[width=0.16\textwidth]{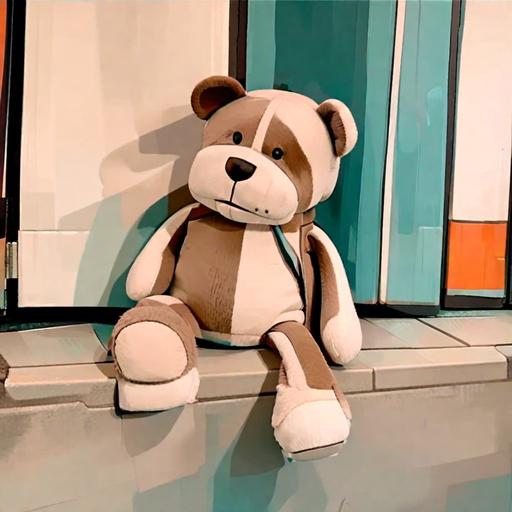} &
        \includegraphics[width=0.16\textwidth]{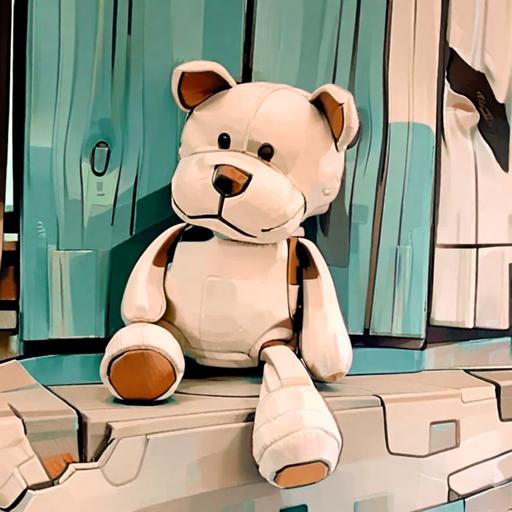} 
        \\
         
    \end{tabular}
    }   
    \caption{
    Qualitative comparison of four different loss schemes used for training the content LoRA: 1) using $\epsilon$-prediction only, 2) using $x_0$-prediction only, 3) transitioning from $x_0$-prediction to $\epsilon$-prediction ($x_0 \rightarrow \epsilon$), and 4) transitioning from $\epsilon$-prediction to $x_0$-prediction ($\epsilon \rightarrow x_0$).
    }
    \label{fig:epsilon_x0}
\end{figure*}
}
\clearpage



\end{document}